%% file: main.tex
\documentclass{article}
\usepackage{iclr2027_conference,times}
\iclrfinalcopy
\usepackage[dvipsnames]{xcolor}
\usepackage{hyperref}

\usepackage{multicol}
\usepackage{graphicx}
\usepackage{amssymb}
\usepackage{amsmath}
\usepackage{bm}
\usepackage{soul}
\usepackage{afterpage}
\usepackage{booktabs}
\usepackage{algorithm}
\usepackage[noend]{algpseudocode}
\usepackage{caption}
\usepackage{subcaption}
\usepackage{xltabular}
\usepackage{array,makecell}
\usepackage{fancyvrb}
\usepackage{adjustbox}
\usepackage{amsthm}
\usepackage{mathtools}
\newtheoremstyle{compact}
  {3pt}                
  {0pt}                
  {\itshape}           
  {}                   
  {\bfseries}          
  {.}                  
  {.5em}               
  {}                   
\theoremstyle{compact}

\usepackage{multirow}
\usepackage{newtxtext}
\usepackage{amssymb}
\usepackage{pifont}
\usepackage{listings}
\usepackage{enumitem}
\usepackage{mathrsfs}

\usepackage[toc,page]{appendix}
\usepackage{titletoc}

\usepackage{tikz}
\usetikzlibrary{patterns}
\usepackage[normalem]{ulem}

\input{macros}

\input{math_commands}

\usepackage{tcolorbox}
\newcommand{\PlanCode}[1]{\ttfamily\small{ #1}}  
\definecolor{DeepGreen}{RGB}{0, 150, 0}
\usepackage{wrapfig}
\usepackage[option]{floatflt}

\newcommand\blfootnote[1]{%
  \begingroup
  \renewcommand\thefootnote{}\footnote{#1}%
  \addtocounter{footnote}{-1}%
  \endgroup
}

\begin{document}

\title{\sys{}: Generative Predicate Invention for Task-level Robot Planning}


\author{\hspace{-18pt}\textbf{Ziyi Yang}\textsuperscript{1}, 
\textbf{Benned Hedegaard}\textsuperscript{1},
\textbf{Ahmed Jaafar}\textsuperscript{1*},
\textbf{Yichen Wei}\textsuperscript{1*},
\textbf{Skye Thompson}\textsuperscript{1},
\textbf{Shreyas Raman}\textsuperscript{\textbf{1}},\and
\textbf{Haotian Fu}\textsuperscript{\textbf{1}},
\textbf{Stefanie Tellex}\textsuperscript{\textbf{1}},
\textbf{George Konidaris}\textsuperscript{\textbf{1}},
\textbf{David Paulius}\textsuperscript{\textbf{1}},
\textbf{Naman Shah}\textsuperscript{\textbf{1}}\textsuperscript{\textbf{2}}\textsuperscript{\textdagger}
\and
\,\textsuperscript{\textbf{1}}Brown University
\quad
\textsuperscript{\textbf{2}}Allen Institute for AI
}

\maketitle

\begin{abstract}
\input{sections/abstract}
\blfootnote{* Equal contribution.}
\end{abstract}

\section{Introduction}
\input{sections/introduction} 

\section{Formal Framework}
\label{sec:problem}
\input{sections/problem_setting_new}

\vspace{-5pt}
\section{Method}
\label{sec:skillwrapper}
\vspace{-5pt}
\input{sections/skillwrapper}
\section{Experiments}
\input{sections/experiments}

\section{Related Work}
\input{sections/related_work}

\section{Conclusion}
\label{sec:conclusion}
\input{sections/conclusion}

\clearpage

\bibliographystyle{iclr2027_conference}
\bibliography{iclr2027_conference}

\clearpage
\onecolumn
\appendix
\begin{appendices}
\startcontents[appendices]
\section*{Contents}
\printcontents[appendices]{}{1}{}

\newpage
\input{sections/appendix}
\stopcontents[appendices]
\end{appendices}

\end{document}

%% file: macros.tex
\newcommand{\sys}{{\sc{SkillWrapper}}}
\newtheorem{definition}{Definition}
\newtheorem{theorem}{Theorem}
\newtheorem{lemma}{Lemma}
\newtheorem{proposition}{Proposition}

\newcommand{\cmark}{\checkmark}

\newcommand{\Booleans}{\{0,1\}} 

\newcommand{\Sequence}[1]{[{#1}]} 
\newcommand{\Tuple}[1]{\langle{#1}\rangle} 
\newcommand{\Set}[1]{\{{#1}\}} 

\newcommand{\PowerSet}[1]{2^{#1}} 
\newcommand{\Def}{\coloneqq} 

\newcommand{\Initial}[1]{{#1}_0} 
\newcommand{\Goal}[1]{{#1}_g} 
\newcommand{\Ground}[1]{\underline{#1}} 

\newcommand{\AbstractSuperscript}{^{\raisebox{0.4ex}{$\scriptscriptstyle\uparrow$}}}
\newcommand{\Abstract}[1]{{#1}\AbstractSuperscript} 

\newcommand{\state}{s}      
\newcommand{\States}{\cS}   

\newcommand{\TransitionF}{T} 

\newcommand{\initialState}{\Initial{\state}} 
\newcommand{\DefInitialState}{\initialState\in\States}
\newcommand{\GoalStates}{\Goal{\States}} 
\newcommand{\DefGoalStates}{\GoalStates\subseteq\States}

\newcommand{\EstInitiationSet}[1][\skill]{\Phi_{#1}}
\newcommand{\EstTerminationSet}[1][\skill]{\Xi_{#1}}

\newcommand{\Types}{\mathsf{T}}    
\newcommand{\TypeF}{\mathsf{t}}   

\newcommand{\parameter}[1][]{\theta_{#1}} 
\newcommand{\Parameters}[1][]{\bm{\theta}_{#1}} 

\newcommand{\DefParameters}[1][]{\Parameters[#1]\Def\Tuple{\parameter[#1]^1, \dots, \parameter[#1]^k}}

\newcommand{\object}{o}      
\newcommand{\Objects}{\cO}   
\newcommand{\ObjectTuple}{\bm{\object}}

\newcommand{\option}{\omega} 
\newcommand{\Options}{\Omega} 

\newcommand{\skill}{\option} 
\newcommand{\Skills}{\Options}

\newcommand{\DefDeterministicSkillTransitionF}{\TransitionF:\States\times\Skills\to\States}

\newcommand{\InitiationSet}[1][\option]{\cI_{#1}} 

\newcommand{\Policy}[1][]{\pi_{#1}} 

\newcommand{\TerminationCondition}[1][\option]{\beta_{#1}} 

\newcommand{\TerminationSet}[1][\option]{\TerminationCondition[#1]}

\newcommand{\PreSuccessSet}[1][\option]{\InitiationSet[#1]}
\newcommand{\SuccessSet}[1][\option]{\TerminationSet[#1]}

\newcommand{\DefSkill}[1][\skill]{{#1}\Def\Tuple{\Parameters[#1], \PreSuccessSet[#1], \Policy[#1], \SuccessSet[#1]}} 

\newcommand{\skillInstance}{\Ground{\skill}} 
\newcommand{\DefSkillInstance}{\skillInstance\Def\skill(\bm{\object})}

\newcommand{\SkillPlan}{\Ground{\bm{\skill}}} 
\newcommand{\DefSkillPlan}[1][k]{\SkillPlan=\Sequence{\skillInstance_1,\dots,\skillInstance_{#1}}}

\newcommand{\predicate}{p}   
\newcommand{\Predicates}{\mathcal{P}}  
\newcommand{\groundAtom}{\Ground{\predicate}}   
\newcommand{\GroundAtoms}{\Ground{\Predicates}} 

\newcommand{\abstractState}{\state^{\scriptscriptstyle\uparrow}}      
\newcommand{\abstractNextState}{{\state'}^{\raisebox{0.0ex}{$\scriptscriptstyle\uparrow$}}}
\newcommand{\AbstractStates}{\Abstract{\States}}    
\newcommand{\abstractAction}{\Ground{a}}     
\newcommand{\AbstractActions}{\Ground{\cA}}  

\newcommand{\AbstractPlan}{\bm{\abstractAction}}
\newcommand{\DefAbstractPlan}{\AbstractPlan\Def\Sequence{\abstractAction_1, \dots,\abstractAction_n}}

\newcommand{\DefAbstractStates}{\AbstractStates\Def\PowerSet{\GroundAtoms}} 

\newcommand{\AbstractionF}{\alpha} 
\newcommand{\DefAbstractionF}{\AbstractionF:\States\to\AbstractStates}

\newcommand{\GroundingF}{\Gamma} 
\newcommand{\DefGroundingF}{\GroundingF:\AbstractStates\to\PowerSet{\States}}

\newcommand{\LiftF}{\Lambda}

\newcommand{\predicateClassifier}{C_{\predicate}} 

\newcommand{\groundAtomClassifier}{C_{\groundAtom}} 
\newcommand{\DefGroundAtomClassifier}{\groundAtomClassifier:\States\to\Booleans}

\newcommand{\semanticLabel}[1][\predicate]{\ell_{#1}}

\newcommand{\DefPredicate}{\predicate\Def\Tuple{\semanticLabel[\predicate], \Parameters[\predicate], \predicateClassifier}}

\newcommand{\operator}{a}  
\newcommand{\Operators}{\mathcal{A}} 

\newcommand{\PRE}[1][\operator]{\textsc{Pre}_{#1}} 
\newcommand{\DefPRE}[1][\operator]{\PRE[#1]\subseteq\Predicates}

\newcommand{\EFF}[1][\operator]{\textsc{Eff}_{#1}} 
\newcommand{\AddEFF}[1][\operator]{\EFF[#1]^+} 
\newcommand{\DeleteEFF}[1][\operator]{\EFF[#1]^-} 

\newcommand{\DefAddEFF}[1][\operator]{\AddEFF[#1]\subseteq\Predicates}
\newcommand{\DefDeleteEFF}[1][\operator]{\DeleteEFF[#1]\subseteq\Predicates}

\newcommand{\DefOperator}[1][\operator]{{#1}\Def\Tuple{\skill_{#1}, \Parameters[#1], \PRE[#1], \AddEFF[#1], \DeleteEFF[#1]}}

\newcommand{\GroundPRE}[1][\operator]{\Ground{\textsc{Pre}}_{#1}} 
\newcommand{\DefGroundPRE}[1][\operator]{\GroundPRE[#1]\subseteq\GroundAtoms}

\newcommand{\GroundEFF}[1][\operator]{\Ground{\textsc{Eff}}_{#1}}

\newcommand{\GroundAddEFF}[1][\operator]{\GroundEFF[#1]^+}
\newcommand{\DefGroundAddEFF}[1][\operator]{\GroundAddEFF[#1]\subseteq\GroundAtoms}

\newcommand{\GroundDeleteEFF}[1][\operator]{\GroundEFF[#1]^-}
\newcommand{\DefGroundDeleteEFF}[1][\operator]{\GroundDeleteEFF[#1]\subseteq\GroundAtoms}

\newcommand{\DefAbstractTransitionRule}{\abstractState_t \gets (\abstractState_{t-1} \setminus \GroundDeleteEFF[\abstractAction_t]) \cup \GroundAddEFF[\abstractAction_t]}

\newcommand{\DefDomain}{\Tuple{\States,\Types,\Skills,\TransitionF}}

\newcommand{\DefSetting}{\Tuple{\initialState, \Objects}}

\newcommand{\DefSkillPlanningProblem}{\Tuple{\Objects, \initialState, \GoalStates}}

\newcommand{\Model}{\cM} 
\newcommand{\DefAbstractModel}{\Tuple{\Predicates, \Operators}}

\newcommand{\cA}{\mathcal{A}}

\newcommand{\cD}{\mathcal{D}}

\newcommand{\cH}{\mathcal{H}}
\newcommand{\cI}{\mathcal{I}}

\newcommand{\cM}{\mathcal{M}}

\newcommand{\cO}{\mathcal{O}}
\newcommand{\cP}{\mathcal{P}}

\newcommand{\cS}{\mathcal{S}}

%% file: math_commands.tex
\usepackage{amsmath,amsfonts,bm}

\def\eqref#1{equation~\ref{#1}}

\def\1{\bm{1}}

\DeclareMathAlphabet{\mathsfit}{\encodingdefault}{\sfdefault}{m}{sl}
\SetMathAlphabet{\mathsfit}{bold}{\encodingdefault}{\sfdefault}{bx}{n}

\def\gA{{\mathcal{A}}}

\def\gM{{\mathcal{M}}}

\def\gP{{\mathcal{P}}}



%% file: sections/abstract.tex
Generalizing from individual skill executions to long-horizon tasks is a core challenge in building autonomous robots.
A promising direction is learning high-level, symbolic representations of low-level robot skills, enabling abstract reasoning independent of the low-level state space.
Recent advances in foundation models have made it possible to generate symbolic predicates that operate on raw sensory inputs---a process we call \emph{generative predicate invention}---to facilitate downstream representation learning.
However, prior work learns these abstractions using heuristic or ad-hoc procedures, leaving unclear \textit{which} formal properties they ought to satisfy, and
\textit{how} these properties can guide representation learning.
%
We address these questions by characterizing conditions under which learned representations support sound and complete task-level planning, and using them to guide the design of \sys{}, a system that autonomously learns symbolic representations of black-box skills without predefined tasks, predicates, or operators.
Our approach leverages foundation models to actively collect robot data and learn human-interpretable, plannable representations directly from RGB observations.
Our extensive empirical evaluation in simulation and on real robots shows that \sys{} learns abstract representations that enable robots to compose black-box skills to solve unseen, long-horizon tasks in the real world.

%% file: sections/introduction.tex
An autonomous robot operating in the real world must process low-level sensory and motor signals while reasoning about high-level objectives~\citep{Doncieux18,KONIDARIS20191}.
Analogous to how humans can perform complex tasks, like cooking or cleaning, without reasoning about muscle-level control, robots should have internal models of their skills that abstract away the intricacies of low-level control.
Such models must capture the necessary conditions for executing a skill (e.g., \textit{``pouring from a teapot into a mug requires holding the teapot first''}) and the consequences of doing so (e.g., \textit{``pouring from a teapot into a mug fills the mug''}).
These two properties, known as \emph{preconditions} and \emph{effects} in the classical planning literature~\citep{ghallab_nau_traverso_2016}, enable compositional reasoning to identify long-horizon plans that sequence lower-level skills to solve a task.
Typically, these task-level models must be specified manually by a domain expert. However, in real-world settings, such models are nontrivial to acquire due to complex inter-skill constraints specific to the environment and the robot's embodiment.
This calls for systems that autonomously learn task-level models of black-box skills, enabling robots to solve long-horizon tasks using off-the-shelf planners.

Previous approaches that learn task-level models for robot planning have taken one of two approaches. Bottom-up methods~\citep{konidaris2009skill, james2022autonomous, shah2024reals} have provable theoretical properties---i.e., can guarantee that the resulting representations are sound and complete---but either start from restricted, structured states (e.g., configuration space~\citep{shah2024reals}) or can learn abstractions from pixel-level inputs but do not exploit foundation models (FMs) and, therefore, may struggle to scale~\citep{Konidaris2018FromST, james2022autonomous}. Alternatively, methods driven by FMs exploit their capabilities to generate semantically-meaningful predicates from pixel-level observations---a process we refer to as \emph{generative predicate invention}. 
%
%
%
On the other hand, while recent work in this line has explored how FMs can be used for predicate invention by generating Python code~\citep{liang2025visualpredicator} or
filling a pool of candidate predicates during hill-climbing search~\citep{athalye2026pix2pred}, predicate invention is still largely driven by heuristic or ad-hoc procedures. This leaves core questions unanswered:
\textit{when} should new predicates be invented, \textit{what} formal properties should the learned predicates and abstract model satisfy, and \textit{how} can these properties guide representation learning?


Our answer is twofold. First, we characterize formal conditions for abstract models to support \textit{sound} and \textit{complete} task-level planning, and use them to derive principled criteria for generative predicate invention.
%
Building on these criteria, we introduce \sys{}, a method designed to guide representation learning toward these properties.
\sys{} uses FMs to interactively collect skill execution data, propose predicates and define their semantics, and classify predicate truth values on low-level observations.
Using these data and predicates, \sys{} learns human-interpretable symbolic representations of black-box skills that enable task-level planning.
%
To the best of our knowledge, \sys{} is the first generative predicate-invention method to learn symbolic representations from scratch without predefined predicates, operators, or tasks for data collection, and the first to use formally characterized model properties to guide predicate invention.


Our core contributions are the following: \emph{1)} \sys{}, an autonomous system that leverages FMs to learn human-interpretable symbolic representations of black-box skills without predefined predicates, operators, or tasks for data collection; \emph{2)} a theoretical characterization of conditions under which abstract models support \emph{sound} and \emph{complete} task-level planning, from which we derive principled criteria for predicate invention in \sys{}; and \emph{3)} an extensive empirical evaluation of the system in simulation and on real robots.

\begin{figure*}[t]
    \centering
        \includegraphics[width=\linewidth]{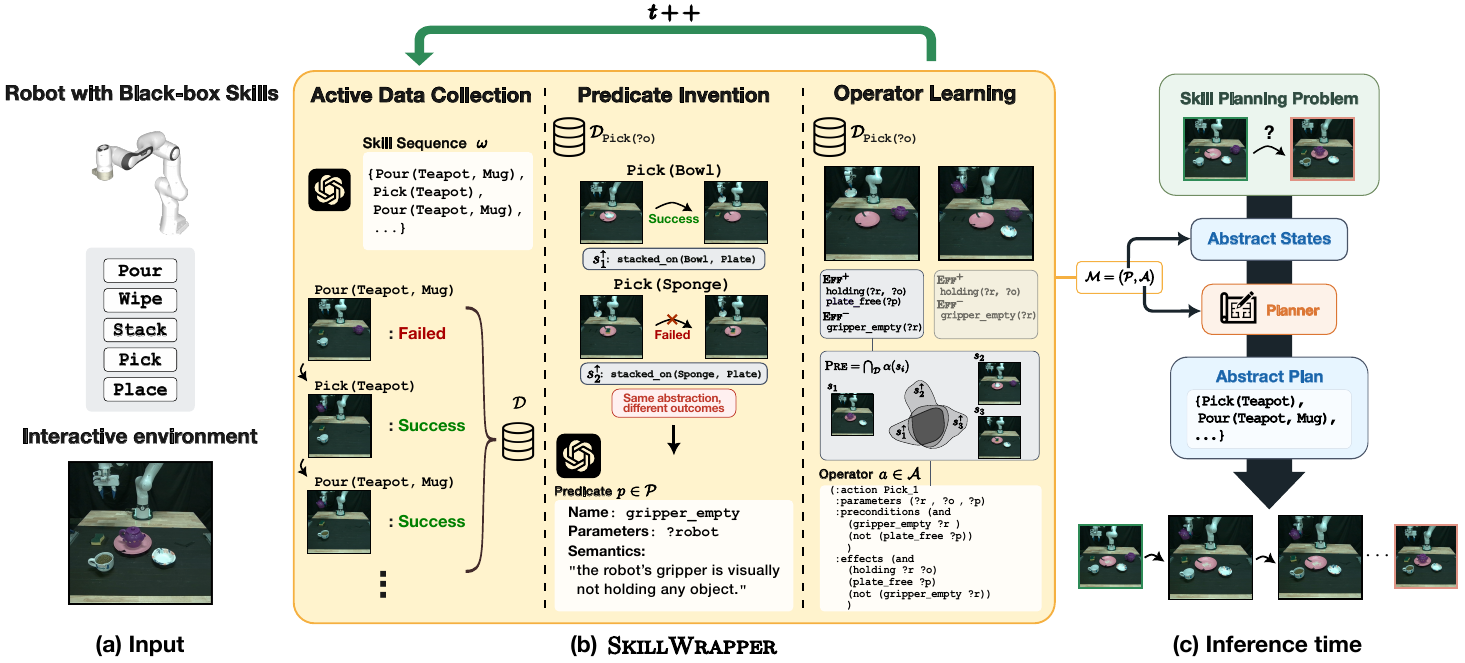}
    \caption{\textbf{Overview of\;{\scshape\bfseries SkillWrapper}.} For a robot equipped with black-box skills, \sys{} learns a task-level model $\mathcal{M}$ that is compatible with off-the-shelf classical AI planners.
    Given a planning problem specified by low-level initial and goal states, the system computes the corresponding abstract states, then calls a planner to find a solution plan to reach the goal.
    }
    \label{fig:overview}
    \vspace{-15pt}
\end{figure*}

%% file: sections/problem_setting_new.tex
We consider a problem setting in which a robot is equipped
with a set of predefined, ``black-box'' skills.
Because the robot initially lacks a transition model for these skills, it is nontrivial to compose them to solve long-horizon problems.
However, if the robot were to acquire a symbolic model of its skills, it could use classical planning to compose them, enabling it to solve unseen tasks.
For the reader's convenience, we provide a glossary of notation
in Appendix~\ref{app:notation-glossary}.


\textbf{Environment}\quad
A \emph{domain} is a tuple $\DefDomain$, where $\States$ is the continuous state space, $\Types$ is a finite set of object types, and $\Skills$ is a library of object-centric skills.
The transition function $\DefDeterministicSkillTransitionF$ characterizes the domain dynamics but is unknown to the robot.
We assume deterministic transitions, as is common in related work~\citep{silver2023predinvent,han2024interpret, liang2025visualpredicator,athalye2026pix2pred}; see Appendix~\ref{appendix:comparison_table} for a full comparison with related problem settings.
Within a domain, a \emph{scene} is defined by $\DefSetting$, consisting of an initial state $\DefInitialState$ and a set of typed objects $\Objects$. We denote the type(s) of an object $\object\in\Objects$ as $\TypeF(\object)\subseteq\Types$.

\textbf{Black-box object-centric skills}\quad
We model each skill \(\skill\in\Skills\) as a tuple
 $\DefSkill$,
where skill parameters $\DefParameters[\skill]$ specify type constraints $\TypeF(\parameter[\skill]^i)\in\Types$ on object arguments;
$\PreSuccessSet:\Objects^k\rightarrow 2^\States$ is the \textit{pre-success set}, containing states from which the skill succeeds under the low-level policy $\Policy$; and $\beta_\omega:\mathcal{O}^k\rightarrow 2^\States$ is the \textit{success set}, containing states corresponding to successful execution.
%
A skill instance $\DefSkillInstance$ is defined for objects $\bm{\object} = \Tuple{\object_1,\dots,\object_k}\in\Objects^k$ iff $\TypeF(\parameter[\skill]^i) \in \TypeF(\object_i)$ for $i\in\{1,\dots,k\}$.
%
%
%
%
An execution of \(\skillInstance\) from state \(\state \in \States\) produces a transition \(\langle \state,\skillInstance,\state'\rangle\), where \(\state'=\TransitionF(\state,\skillInstance)\). 
We consider skill executions initiated outside the corresponding success set, excluding attempts whose objective is already satisfied.
An execution succeeds if and only if it starts from the pre-success set (\(\state\in\PreSuccessSet[\skillInstance])\), in which case it terminates in the success set (\(\state'\in\SuccessSet[\skillInstance])\).
We allow a type-compatible skill instance to be attempted outside \(\PreSuccessSet[\skillInstance]\), in which case the execution fails. We assume that the robot can observe whether an execution succeeds or fails, but does not know \(\PreSuccessSet[\skillInstance]\) or \(\SuccessSet[\skillInstance]\) a priori, so the skills are considered black boxes.

%


\textbf{Symbolic abstractions and groundings}\quad
To enable reasoning over a discrete space of possible abstract states, a set of predicates is denoted $\predicate \in \Predicates$, each defined by
$\DefPredicate$, where $\semanticLabel$ is a semantic label, and
parameters $\Parameters[\predicate]$ define type constraints on object arguments of the classifier ${\predicateClassifier}$.
Grounding a predicate $\predicate$ with type-compatible objects induces a grounded predicate $\groundAtom$ and its classifier ${\DefGroundAtomClassifier}$, which tests whether the grounded predicate holds in a given state.
Given a set of predicates $\Predicates$ and a set of objects $\Objects$,
let $\GroundAtoms$ denote the set of all possible grounded predicates, and define the \emph{abstract state space} as $\DefAbstractStates$, the symbolic counterpart to the state space $\States$.
%
%
The \emph{abstraction function} $\DefAbstractionF$ maps each low-level state to the set of grounded predicates true in that state (i.e., the corresponding abstract state),
while the \emph{grounding function} $\DefGroundingF$ maps an abstract state $\abstractState\in\AbstractStates$ to the low-level states in which its grounded predicates hold.
%


An operator $\operator\in\Operators$ is defined by $\DefOperator$, denoting a skill $\skill_\operator$, typed parameters $\Parameters[\operator]$, preconditions $\DefPRE$, add effects $\DefAddEFF$, and delete effects $\DefDeleteEFF$.
Given operators $\Operators$ and objects $\Objects$, the \emph{abstract action space} $\AbstractActions$ is the set of all valid operator groundings $\abstractAction = \operator(\ObjectTuple)$, defined for objects $\bm{\object} = \Tuple{\object_1,\dots,\object_m}\in\Objects^m$ iff $\TypeF(\parameter[\operator]^i) \in \TypeF(\object_i)$ for $i\in\{1,\dots,m\}$; each abstract action $\abstractAction$ defines 
ground preconditions $\DefGroundPRE[\abstractAction]$, ground add effects $\DefGroundAddEFF[\abstractAction]$, and ground delete effects $\DefGroundDeleteEFF[\abstractAction]$.
We refer to the tuple $\cM \Def \DefAbstractModel$ as an \emph{abstract transition model}, and an \emph{abstract plan} found using $\cM$ is a sequence of abstract actions: $\DefAbstractPlan$.

\textbf{Planning problems}\quad
A \emph{skill planning problem} is defined by $\DefSkillPlanningProblem$, which is a set of objects $\Objects$, an initial state $\DefInitialState$, and goal states $\DefGoalStates$.
%
%
%
A solution to a skill planning problem is any \emph{skill plan} $\DefSkillPlan[k]$
such that
for $i \in \{1, \dots,k\}$,
$\state_i = \TransitionF(\state_{i-1}, \skillInstance_i), \state_{i-1} \in \PreSuccessSet[\skillInstance_i]$, and $\state_k \in \GoalStates$.


%

%

\textbf{Problem statement}\quad
Given a domain $\DefDomain$ containing scenes $\Set{\DefSetting}_{i=1}^N$, we define the problem as learning a vocabulary and an abstract transition model $\Model\Def\DefAbstractModel$ for the skills $\Skills$ defined over $\Types$, such that when given a skill planning problem $\DefSkillPlanningProblem$ in a novel scene, a complete symbolic planner using $\Model$ will find a valid abstract plan $\AbstractPlan$, whenever a feasible skill plan exists.


%% file: sections/skillwrapper.tex
%

                
                
                    

We introduce~\sys, a system that autonomously learns representations for black-box skills.
The algorithm iterates through three steps:
\emph{1)} active exploration for data collection,
\emph{2)} predicate invention based on contrastive examples, and
\emph{3)} operator learning using the accumulated data and predicates.
%
%
As \sys{} continues to collect data, invent and incorporate new predicates, and update its planning model, it progressively improves the fidelity of these learned representations.
Full pseudocode is in Appendix~\ref{append:pseudocodes}.
%
%
%
%
        

        
            
        
%

\subsection{Active Data Collection}
\label{sec:data-collection}
To collect data for learning, our method selects a sequence of skill instances $\DefSkillPlan$
for the robot to execute.
After executing each $\skillInstance_i$, the robot records a \emph{transition} $\Tuple{\state_{i-1}, \skillInstance_i, \state_i}$ capturing both the \emph{executability} of the skill instance (i.e., whether $\state_{i-1}\in\InitiationSet[\skillInstance_i]$) and its \emph{outcome} (i.e., the resulting state $\state_i$).
%
%
These transitions are accumulated into a dataset ${\cD = \{\Tuple{\state,\skillInstance,\state'}\}}$ used for predicate invention and operator learning.
We define $\cD_\skill$ as the set of transitions in $\cD$ involving a skill $\skill\in\Skills$.
To propose a skill sequence, \sys{} first prompts an FM with the robot skills $\Skills$ and the current predicates $\mathcal{P}$, which may be empty, to generate a candidate pool.
Each candidate sequence is scored along two axes, \emph{balance} ($b$) and \emph{coverage} ($c$).
\sys{} retains the Pareto-optimal subset containing all sequences $i$ for which no other sequence $j$ has $b_j \geq b_i$ and $c_j \geq c_i$.
The final skill sequence is sampled uniformly from this subset.


\textbf{{Balance} ($b$)} encourages candidate skill sequences to contain a desired mixture of successful and failed executions. To estimate this mixture, \sys{} simulates each candidate sequence using the current abstract model, maintaining a predicted abstract state as the sequence progresses. At each step $t$, a skill instance $\skillInstance_t$ is predicted to succeed if at least one associated ground operator $\abstractAction_t \in \AbstractActions_{\skillInstance_t}$ is applicable, i.e., its preconditions satisfy $\abstractState_{t-1} \models \GroundPRE[\abstractAction_t]$. If multiple operators are applicable, we select the most specific one, defined as the operator with the largest number of preconditions, so that conditional effects are applied only when their corresponding conditions are satisfied. The abstract state is then updated according to the selected operator, ${\DefAbstractTransitionRule}$; if no operator is applicable, the skill is predicted to fail and the abstract state remains unchanged. Let $\hat r_\text{suc}$ denote the fraction of skill executions predicted to succeed over the simulated sequence. Given a target success ratio $r_\text{suc}^*$, the balance score is $b=-|\hat r_\text{suc}-r_\text{suc}^*|$. A balanced target (e.g., $r_\text{suc}^*=0.5$) encourages both successful and failed executions, providing contrastive examples for predicate invention, while a higher target (e.g., $r_\text{suc}^*=0.9$) favors successful executions and provides more data for learning skill effects.

\textbf{Coverage ($c$)} encourages candidate sequences to cover pairs of consecutively executed skills more uniformly, helping \sys{} expose interdependencies between skills' preconditions and effects. We maintain a skill-bigram count matrix $Q$, where each entry $Q[i,j]$ records how often the consecutive pair $(\skillInstance_i,\skillInstance_j)$ has appeared in the collected dataset $\cD$. For a candidate sequence $\SkillPlan$, let $Q'$ denote the hypothetical count matrix after additionally counting the consecutive skill pairs in $\SkillPlan$. We score the candidate by how much it increases the entropy of this empirical bigram distribution, $c(\SkillPlan)=H(Q')-H(Q)$, where $H(Q)=-\sum_{i,j}p_{i,j}\log p_{i,j}$ and $p_{i,j}=Q[i,j]/\sum_{k,\ell}Q[k,\ell]$ is the normalized frequency of each skill pair. Thus, sequences containing relatively underexplored skill pairs tend to receive higher coverage scores.
%

\subsection{Predicate Invention}
\label{sec:predicate_invention}

Unlike prior work that requires an initial set of predicates (e.g., goal predicates)~\citep{silver2023predinvent,liang2025visualpredicator,liang2025exopredicatorlearningabstractmodels,athalye2026pix2pred} or partial operators (e.g., with pre-specified effects)~\citep{liang2025visualpredicator,liang2025exopredicatorlearningabstractmodels},
\sys{} begins without predicates or operators ($\Predicates = \Operators = \emptyset$), instead bootstrapping based on black-box skill executions.


Given grounded predicates $\Ground{\Predicates}$ and an object tuple $\ObjectTuple \in \Objects^k$, we define a \emph{lifting function} $\LiftF_{\ObjectTuple}$ that replaces the objects in $\ObjectTuple$ with typed variables, allowing predicates from different skill instances to be compared in a common lifted form (details in App.~\ref{appendix:operator_learning}).
We also define the \emph{abstract change} between states $\state,\state' \in \States$ as $\Delta(\state, \state') = \Tuple{\AbstractionF(\state') \setminus \AbstractionF(\state), \AbstractionF(\state) \setminus \AbstractionF(\state')}$. 

\textbf{Conditions for predicate invention.}\,
\sys{} invents new predicates only when the task-level model cannot explain newly observed transitions.
To formalize when the current vocabulary is insufficient, for each skill instance $\skillInstance$, we define the set of states from which the model predicts $\skillInstance$ would succeed 
and in which it predicts that $\skillInstance$ has succeeded: 
%
%
\begin{equation*}
\textstyle
\EstInitiationSet[\skillInstance] \coloneqq
\bigcup_{\abstractAction\in \AbstractActions_{\skillInstance}}
\GroundingF{(\GroundPRE[\abstractAction])},
\quad
\EstTerminationSet[\skillInstance] \coloneqq
\bigcup_{\abstractAction\in \AbstractActions_{\skillInstance}}
\GroundingF{((\GroundPRE[\abstractAction]
\setminus \GroundDeleteEFF[\abstractAction])
\cup \GroundAddEFF[\abstractAction])},
\end{equation*}
where $\AbstractActions_{\skillInstance}$ denotes the abstract actions for $\skillInstance$.
    
The first condition arises when the learned model incorrectly predicts the pre-success set $\PreSuccessSet[\skill]$ of a skill.
Concretely, this occurs when two transitions involving instances of the same skill, one successful and one failed, are both predicted to succeed by the current model.
Letting $\skillInstance_i = \skill(\ObjectTuple_i)$ and $\skillInstance_j = \skill(\ObjectTuple_j)$:
%
\begin{equation}
\label{eq:precond-invention}
\begin{split}
    \exists \, \langle \state_i, \skillInstance_i, \state'_i \rangle, \langle \state_j, \skillInstance_j, \state'_j \rangle \in \cD_\skill \text{ s.t. }  \state_i \in \PreSuccessSet[\Ground{\skill}_i] \text{ and } \state_j \notin \PreSuccessSet[\Ground{\skill}_j],
    \text{but }  \state_i \in \EstInitiationSet[\Ground{\skill}_i], \state_j \in \EstInitiationSet[\Ground{\skill}_j].
    %
%
\end{split}
\end{equation}
The second condition detects when the predicate vocabulary cannot distinguish between the effects of a successful and a failed execution of the same skill.
We use the lifting function $\LiftF_{\ObjectTuple}$ to account for the fact that the two transitions may involve different object arguments (e.g., $\ObjectTuple_i$ and $\ObjectTuple_j$):
\begin{equation}
\label{eq:effect-invention}
\begin{aligned}
&\exists \Tuple{\state_i, \skill(\ObjectTuple_i), \state_i'}, \Tuple{\state_j, \skill(\ObjectTuple_j), \state_j'} \in \cD_\skill \text{ s.t. } \state_i' \in \SuccessSet[\skill(\ObjectTuple_i)] \text{ and } \state_j' \notin \SuccessSet[\skill(\ObjectTuple_j)], \text{ yet } \\
& \LiftF_{\ObjectTuple_i}(\Delta(\state_i, \state_i')) = \LiftF_{\ObjectTuple_j}(\Delta(\state_j, \state_j')).
\end{aligned}
\end{equation}

\begin{figure*}[t!]
    \centering
    \includegraphics[width=0.98\linewidth]{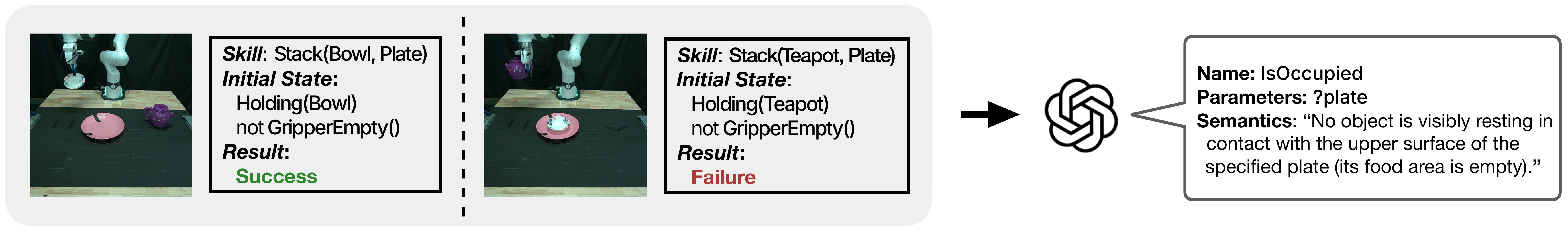}
    \caption{
        \textbf{Example of Predicate Invention.} The initial states of the two transitions both satisfy the preconditions of certain operators of the {\small \texttt{Stack}} skill, while one transition is successful, but the other is not. In this case, condition~\ref{eq:precond-invention} is triggered, and the FM is prompted to generate a new predicate.
    }
    \label{fig:predicate_invention_example}
    \vspace{-18pt}
\end{figure*}


\textbf{Predicate implementation.}\,
We employ FMs to generate and evaluate predicates:
\emph{1) Predicate generation}: Given a contrastive pair of transitions as identified by Conditions~\ref{eq:precond-invention} or \ref{eq:effect-invention}, \sys{} prompts the FM to generate a candidate predicate $\hat \predicate$ to distinguish the pair defined by the semantic label $\semanticLabel$ (see Fig.~\ref{fig:predicate_invention_example}).
\emph{2) Predicate evaluation}: Given a grounded predicate $\Ground{\predicate}$ and an image-based low-level state $\state\in\States$, the FM is queried for the truth value of $\Ground{\predicate}$ in $\state$ by its semantic label $\semanticLabel$.
    


\textbf{Predicate selection.}\,
{
    We introduce a scoring function to estimate the utility of a candidate predicate $\hat{p}$ by learning hypothetical operators $\hat{\Operators}$ under the vocabulary $\hat{\Predicates} = \Predicates \cup \{ \hat \predicate \}$.
    %
    %
    %
    For each successful transition $\Tuple{\state_i, \skillInstance_i, \state_i'} \in \cD$, there must exist at least one ground operator $\hat{\Ground{\operator}} \in \Ground{\hat{\Operators}}$ with $\state_i \models \GroundPRE[\hat{\Ground{\operator}}]$;
    for each failed transition $\Tuple{\state_j, \skillInstance_j, \state_j'} \in \cD$, no such operator may exist.
    Effect scoring follows the same principle based on the abstract change $\Delta(s,s')$.
    After evaluating over $\cD$, a candidate predicate is kept if its score exceeds an empirical threshold.
    Thus, whether a generated predicate is added to the vocabulary is determined using all observed transitions, while the contrastive pair provides visual cues to the FM for proposing a candidate.
    Further details can be found in Algorithm~\ref{appendix:predicate_invention} (App.~\ref{append:pseudocodes}).
}

\textbf{Empirical noise handling.}\,
FM-generated predicates and their evaluated truth values can be imperfect due to real-world noise. \sys{} handles such noise empirically in two places: during predicate selection, each candidate is scored over the full set of observed transitions, making the selection robust to a small fraction of misclassifications and filtering occasional low-quality predicates; during operator learning, small effect clusters are pruned as noisy outliers, following standard practice in symbolic model learning. Together, these mechanisms reduce the impact of occasional FM errors without assuming oracle predicate generation or classification. Our experiments demonstrate that \sys{} remains effective under real-world noise, while the component-wise evaluation in Appendix~\ref{appendix:vlm_study} provides additional evidence of FM reliability.


\subsection{Operator Learning}
\label{subsec:operatorlearning}

A single skill may produce distinct abstract state changes depending on the context of execution.
For example, {\small \texttt{GoTo(desk)}} may bring objects on a desk into reach, if the desk is not empty.
%
%
To represent these \emph{conditional effects}, we extend the \emph{associative model learning} paradigm~\citep{arora2018review} to our setting.
\sys{} clusters observed transitions based on their lifted abstract changes $\LiftF_{\ObjectTuple}(\Delta(s,s'))$, enabling it to learn a single operator across distinct instantiations $\skill(\ObjectTuple)$ of a skill $\skill$ (cf.~Eq.~\ref{eq:effect-invention}).
%
For each cluster $\bm{d} \subseteq \cD_\skill$, we obtain its preconditions by intersecting the lifted initial abstract states of all transitions in the cluster:
$\PRE[] = \bigcap_{i:\,\tau_i \in \bm{d}} \LiftF_{\ObjectTuple_i}(\abstractState_i)$,
where $\tau_i = \Tuple{\abstractState_i,\skill(\ObjectTuple_i),\abstractNextState_i}$.
We retain only predicates involving at least one of the skill arguments in $\ObjectTuple_i$.
%

In realistic domains, objects might belong to multiple overlapping categories (e.g., a {\small $\mathtt{Mug}$} is {\small $\mathtt{fillable}$}, while a {\small $\mathtt{Bottle}$} is both {\small $\mathtt{fillable}$} and {\small $\mathtt{openable}$}).
To solve this, {\sys} conservatively assigns operator parameters with the lowest level of the type hierarchy consistent with the data, preventing over-generalization.
%
As a result, the learned predicate set remains compact, informative, and aligned with the most recent transition data.
See Appendix~\ref{appendix:operator_learning} for a step-by-step example.

\subsection{Theoretical Analysis}
\label{sec:theoretical-analysis}

We first characterize correctness, soundness, and completeness of an abstract transition model, then establish empirical consistency and asymptotic convergence properties of \sys{}. Full proofs of the theoretical results are provided in Appendix~\ref{app:proofs}.

\textbf{Model properties.}\,
We consider three properties of an abstract model \(\Model\). A model is \textit{correct} if its predicted pre-success $\EstInitiationSet$ and success sets $\EstTerminationSet$ exactly match those of every skill; \textit{soundness} requires every abstract plan to correctly predict the outcome of its corresponding skill execution, while \textit{completeness} requires every feasible skill plan to have a corresponding abstract plan. We provide the formal definitions in
Appendix~\ref{appendix:theory}.

\sys{} can learn conditional effects by representing distinct effects of the same skill with separate operators. When the preconditions of such operators overlap, however, multiple operators may be simultaneously applicable while predicting different outcomes, making a universal soundness guarantee inherently difficult without additional disambiguation. We therefore define soundness over effect-unambiguous plans, while still allowing \sys{} to learn and use conditional effect operators in practice.
\begin{proposition}
Any correct model learned by \sys{} is sound and complete.
\end{proposition}
\noindent \textit{Proof sketch.}  
Correctness ensures that the model exactly captures where each skill succeeds and its successful outcomes.
By construction, \sys{} clusters successful transitions by effect and derives each operator's preconditions from the same cluster, preserving their correspondence; effect-unambiguity then ensures an applicable operator predicts the corresponding abstract effect.
Applying this correspondence inductively over a skill sequence establishes soundness and completeness. \qed

Next, we analyze how \sys{} approaches this property through its predicate invention conditions.
Lemma~\ref{lemma:consistency} shows that, after all identified contrastive transition pairs have been resolved, the estimated and true success sets agree on the observed terminal states. Together with operator learning and the precondition invention condition, this yields Theorem~\ref{thm:soundness}, which establishes empirical consistency of both the pre-success and success sets on \(\cD\). This result forms the basis for the asymptotic convergence property in Theorem~\ref{thm:prob-completeness}.

\begin{lemma}
\label{lemma:consistency}
Let $\cM$ be the model learned by \sys{} from dataset $\cD$
after all contrastive transition pairs satisfying the
predicate-invention conditions have been resolved.
The estimated success set agrees with the true success set
on every observed terminal state: for each transition
$\Tuple{\state, \skillInstance, \state'} \in \cD,
    \state' \in \EstTerminationSet[\skillInstance]
    \iff
    \state' \in \SuccessSet[\skillInstance]$.

\end{lemma}

\noindent \textit{Proof Sketch.}
Suppose not. Then some observed transition \(\tau = \Tuple{\state,\skillInstance,\state'}\) has inconsistent terminal membership. If \(\state' \in \EstTerminationSet[\skillInstance]\setminus\SuccessSet[\skillInstance]\), then \(\tau\) is failed but \(\state'\) satisfies the conditions predicted by a learned operator. Thus, \(\tau\) has the same lifted abstract effect as a successful transition supporting that operator, triggering Condition~\ref{eq:effect-invention}, a contradiction. Conversely, if \(\state' \in \SuccessSet[\skillInstance]\setminus\EstTerminationSet[\skillInstance]\), then \(\tau\) is successful under our definition. Operator learning therefore constructs an operator covering \(\tau\), implying \(\state'\in\EstTerminationSet[\skillInstance]\), again a contradiction. \qed
\begin{theorem}[Empirical Consistency of {\sys}]
\label{thm:soundness}
Let $\cM$ be the model learned by \sys{} from dataset $\cD$
after all identified contrastive transition pairs have been resolved.
The estimated pre-success and success sets agree with their
true counterparts on every observed transition
$\Tuple{\state, \skillInstance, \state'} \in \cD$: $\state \in \EstInitiationSet[\skillInstance] \iff \state \in \PreSuccessSet[\skillInstance], \, \state' \in \EstTerminationSet[\skillInstance] \iff \state' \in \SuccessSet[\skillInstance]$.
\end{theorem}

\textit{Proof Sketch.}\quad
For each observed transition $\Tuple{\state,\skillInstance,\state'} \in \cD$,
if $\state \in \PreSuccessSet[\skillInstance]$, the transition is successful
and belongs to a cluster used for operator learning.
By construction, the operator learned from this cluster is applicable at
$\state$, so $\state \in \EstInitiationSet[\skillInstance]$.
Conversely, if
$\state \in \EstInitiationSet[\skillInstance]\setminus
\PreSuccessSet[\skillInstance]$,
this failed transition and a successful transition from which the applicable
operator was learned would trigger
Condition~\ref{eq:precond-invention}, contradicting the resolution of all
contrastive pairs.
Thus, the pre-success memberships agree.
The success-set equivalence follows directly from
Lemma~\ref{lemma:consistency}. \qed
To account for temporal dependence in collected data,
let $\tau_{r,t}$ denote transition $t$ in episode $r$ and
$\cD_{1:r-1}$ the preceding collection history.
Conditional on this history, we model each episode's transitions
as a possibly nonstationary $\beta$-mixing process, with dependence
between observations separated by $k$ steps bounded by
$\beta(k)$~\citep{yu1994rates}.
For a candidate model $\cM$, define the transition-level error
and its average conditional error across $K$ episodes as
\[
\begin{aligned}
\ell_{\cM}(\tau)
&\Def \mathbf{1}\!\left\{
\state\in\PreSuccessSet[\skillInstance]\ominus\EstInitiationSet[\skillInstance]
\;\lor\;
\state'\in\SuccessSet[\skillInstance]\ominus\EstTerminationSet[\skillInstance]
\right\},\\
d_{\mathrm{compl},K}(\cM,\cM^*)
&\Def \frac{1}{K}\sum_{r=1}^{K}
\mathbb{E}\!\left[
\frac{1}{n}\sum_{t=1}^{n}\ell_{\cM}(\tau_{r,t})
\,\middle|\,\cD_{1:r-1}
\right].
\end{aligned}
\]
where $\ominus$ denotes symmetric difference.
We first establish the corresponding single-episode bound in
Lemma~\ref{lem:episode-generalization} (Appendix~\ref{app:proofs}),
and use it below to characterize convergence across episodes.

\begin{theorem}[Asymptotic Convergence of {\sys}]
\label{thm:prob-completeness}
Let $\cH$ be a fixed finite model hypothesis class, and let
$\widehat{\cM}_K\in\cH$ satisfy Theorem~1 on all transitions
from $K$ episodes of length $n$.
For any $\epsilon\in(0,1)$ and integer $1\le k\le n$, let
$m=\lfloor n/k\rfloor$.
With probability at least
$1-|\cH|(e^{-Km\epsilon}+K(m-1)\beta(k))$,
the learned model $\widehat{\cM}_K$ satisfies
$d_{\mathrm{compl},K}(\widehat{\cM}_K,\cM^*)\le\epsilon$.
Moreover, this error converges to zero in probability
as the number of episodes increases.
\end{theorem}

\textit{Proof Sketch.}\quad
Conditioning on previously collected data, we apply the
single-episode bound of
Lemma~\ref{lem:episode-generalization} within each episode.
Iterated conditional expectation combines the evidence across
episodes, bounding the probability that a fixed model remains
empirically consistent while its average conditional error
exceeds $\epsilon$ by
$e^{-Km\epsilon}+K(m-1)\beta(k)$.
A union bound over $\cH$, together with Theorem~1, yields
the stated guarantee.
Choosing $k=n$ eliminates the mixing penalty and makes the
failure bound vanish as $K\to\infty$, establishing convergence. \qed

It is important to note that Lemma 1 and Theorem 1 are empirical consistency results and do not require globally correct FM predictions; they establish consistency with the observed transition labels. Theorem 2, together with its single-episode result in Lemma~\ref{lem:episode-generalization}, additionally assumes correct FM-provided predicate labels. While real-world labels are inevitably noisy, our experiments show that the theoretically derived conditions remain effective in practice.

%% file: sections/experiments.tex
%
We conduct simulated and real-world experiments to evaluate:
1) the \emph{effectiveness} of the planning abstractions learned by \sys{} on unseen problems,
2) how well these abstractions \emph{generalize} across environments, and
3) the \emph{data efficiency} of learning from skill executions.
All experiments operate on image observations, and use GPT-5~\citep{openai_gpt5} for predicate generation and evaluation.
The most closely related generative predicate invention methods~\citep{liang2025visualpredicator,athalye2026pix2pred,liang2025exopredicatorlearningabstractmodels} operate in a ``completing partial operators'' setting, assuming additional supervision, such as predefined predicates, partial operators, or training tasks, and are therefore not directly comparable to our ``learning from scratch'' setting (details discussed in Appendix~\ref{appendix:comparison_table}).
We therefore use applicable baselines and controlled ablations, with \emph{FM Invent} serving as the closest adaptation of Pix2Pred~\citep{athalye2026pix2pred} to our setting.
Our evaluation spans six scenes across a simulated domain and two physical robot platforms.
Quantitative results are averaged over five independent runs for simulated experiments, and three for real-robot experiments.
In each run, \sys{} independently collects data, invents predicates, and learns the abstract model.
%
In addition to the full-system results discussed in this section, we provide a comprehensive evaluation of the FM components of \sys{} in
Appendix~\ref{appendix:vlm_study}, including predicate invention comparisons, predicate classification accuracy, unseen objects,
and varied lighting conditions or viewing angles.


\begin{figure*}[t]
    \centering
    \begin{subfigure}[b]{0.31\textwidth}
        \includegraphics[width=\textwidth]{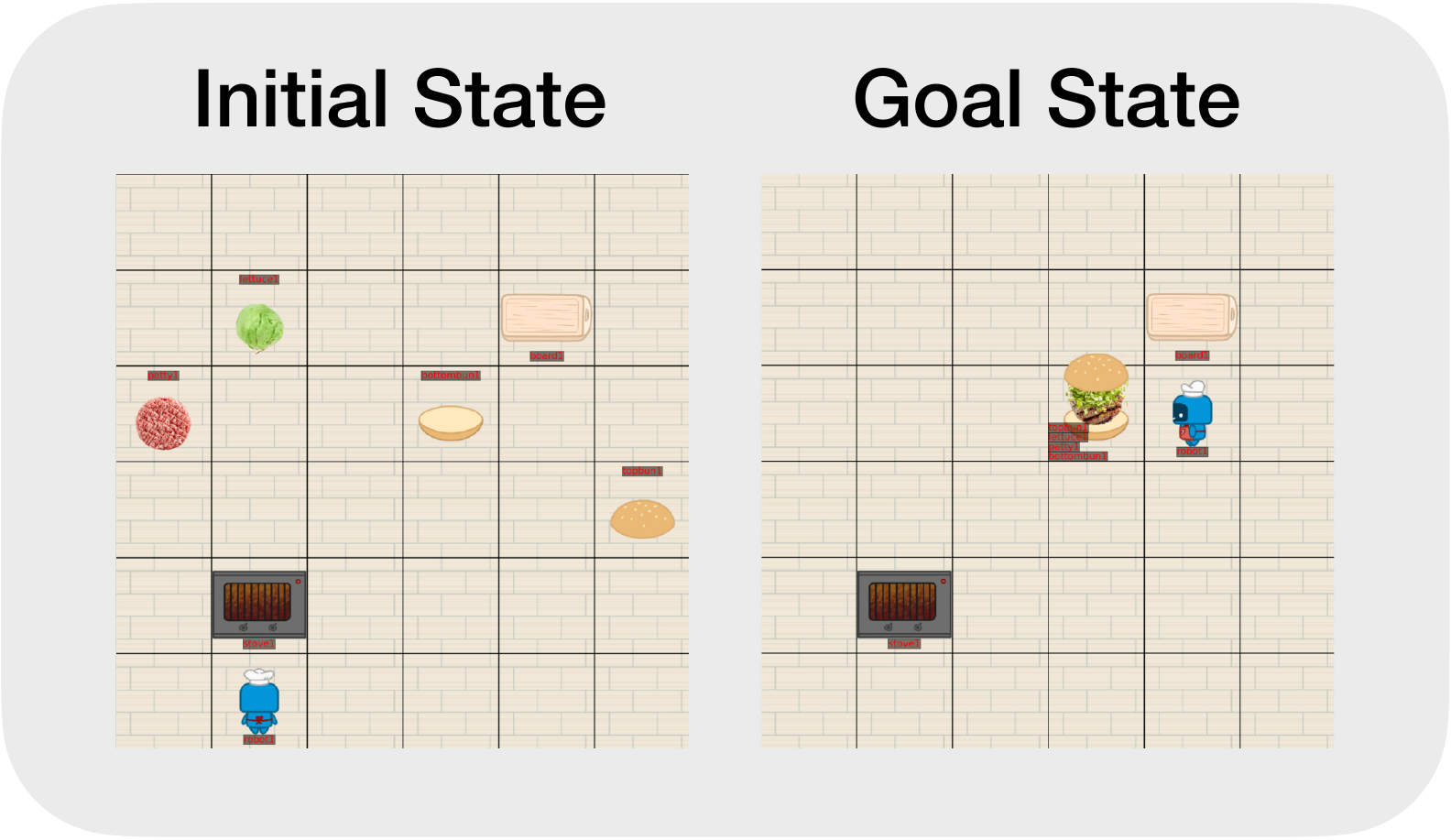}
        \caption{Robotouille}
    \end{subfigure}\hfill
    \begin{subfigure}[b]{0.31\textwidth}
        \includegraphics[width=\textwidth]{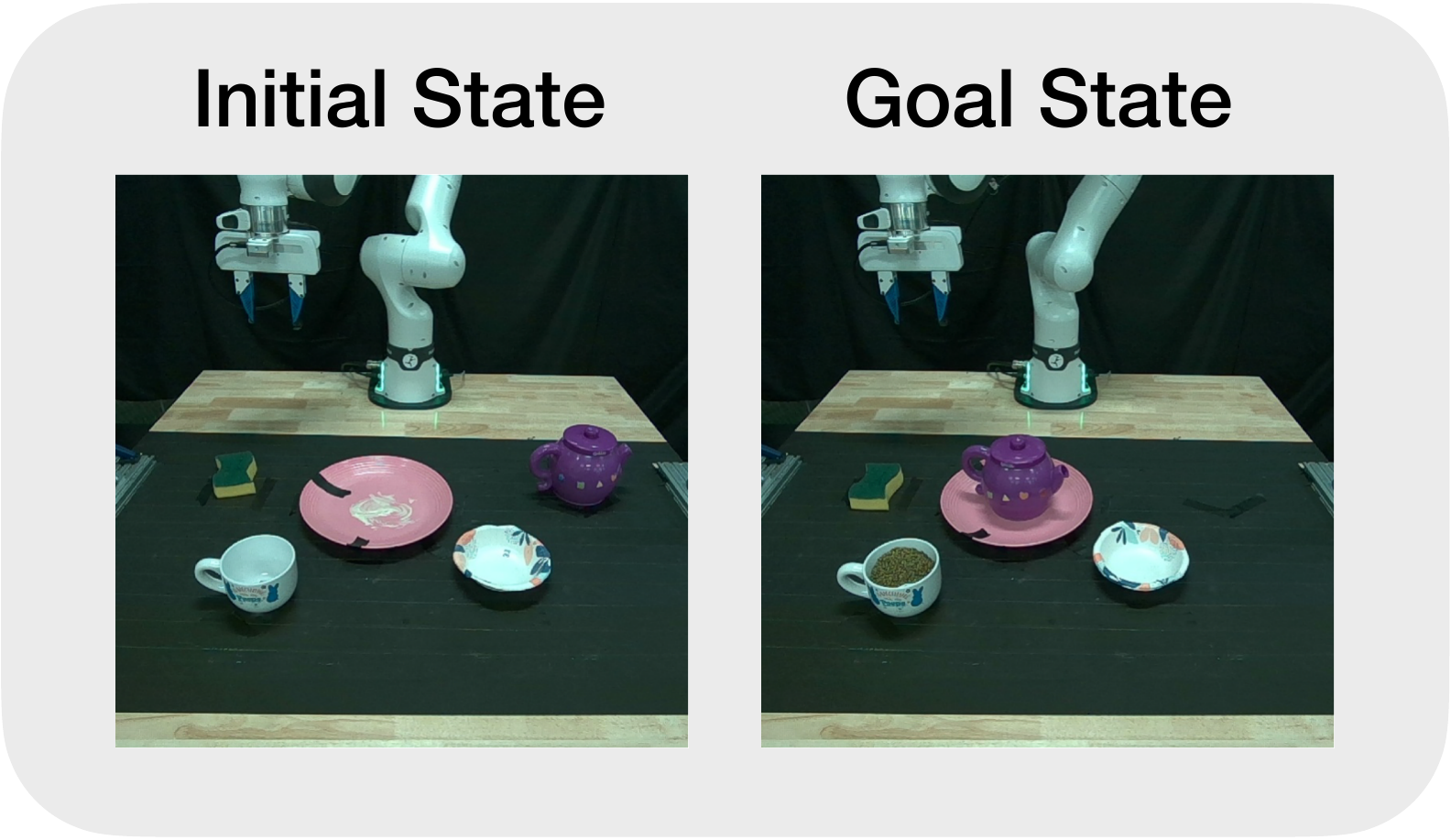}
        \caption{Panda}
    \end{subfigure}\hfill
    \begin{subfigure}[b]{0.31\textwidth}
        \includegraphics[width=\textwidth]{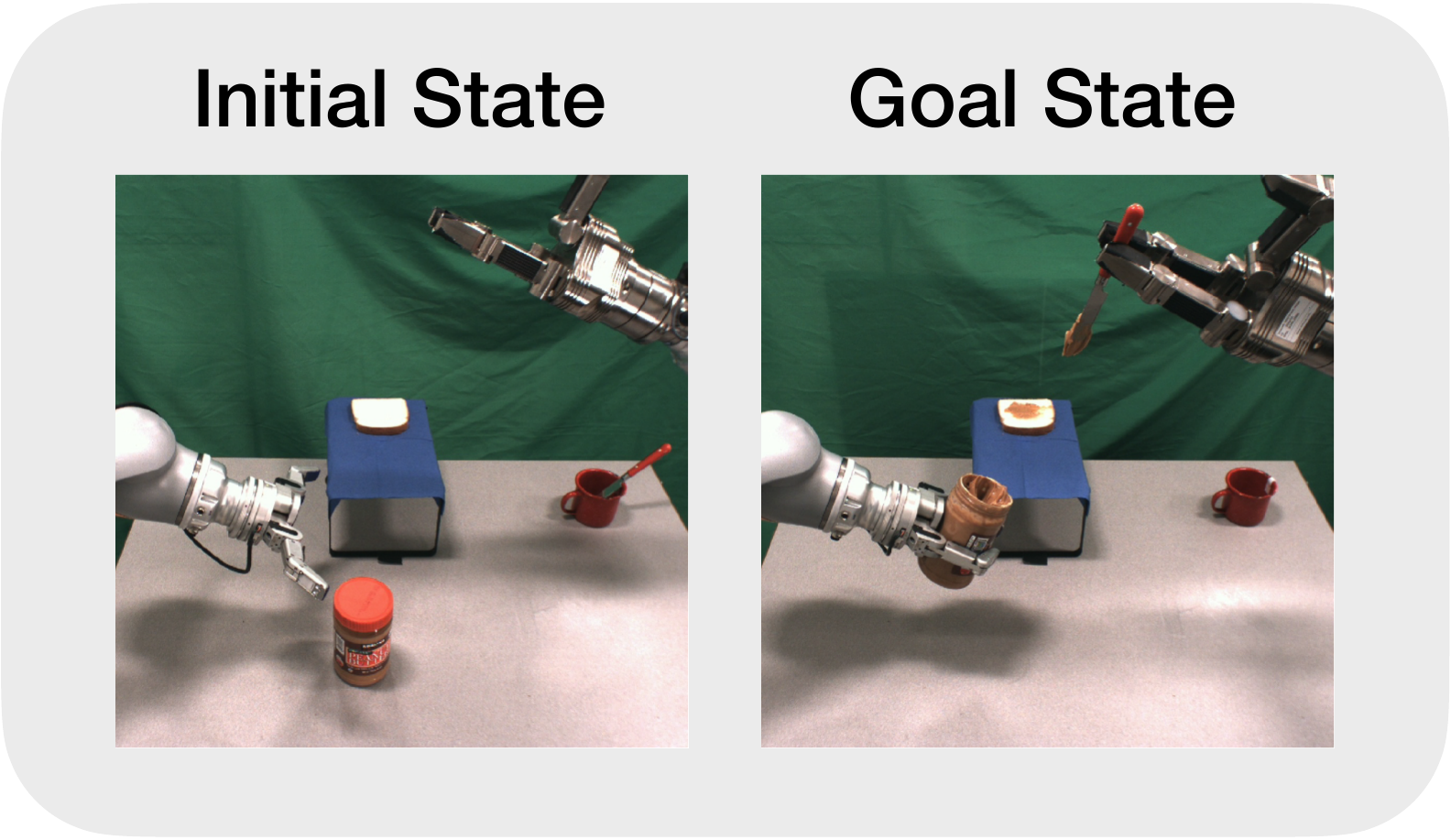}
        \caption{Bimanual Kuka}
    \end{subfigure}
    \vspace{-3pt}
    \caption{Skill Planning Problem in Different Domains.}
    \vspace{-15pt}
    \label{fig:agents}
\end{figure*}


\subsection{Simulated Experiments}

We conduct experiments in Robotouille~\citep{gonzalez-pumariega2025robotouille}, a simulated kitchen environment with six objects and a robot with five skills.
%
%
We design and categorize 50 skill planning problems for this domain: 20 easy problems, whose optimal solutions have at most 7 steps; 20 hard problems, whose optimal solutions have at most 15 steps; and 10 impossible tasks that cannot be realized in the environment, which a sound planning model must identify.
We compare \sys{} against the following baselines:

\begin{itemize}[leftmargin=*, nosep]
\vspace{-5pt}
    \item \textbf{Expert Operators} are manually defined by a human expert as PDDL predicates and operators.
    \item \textbf{System Predicates} uses the simulator's built-in predicates, which are designed to unambiguously define any simulated state, and knows the ground-truth abstract state; cannot invent new predicates.
    \item \textbf{ViLa} \citep{hu2023lookleapunveilingpower} iteratively prompts an FM for the next action, given the current image observation and the robot's action history, until the goal state is reached.
    \item \textbf{Random Exploration} ablates the data collection strategy of \sys{} by randomly sampling skills and valid object arguments. Predicate invention and operator learning are unchanged.
    \item \textbf{FM Invent} ablates the predicate invention method of \sys{} by directly prompting a foundation model (FM) with all image states to invent predicates in batches. The active data collection and operator learning procedures are the same as \sys{}.
    \item \textbf{No Heuristic} is the same as \sys{}, except that skill sequences are selected randomly from the foundation model's output without applying the balance and coverage heuristics (Sec.~\ref{sec:data-collection}).
\end{itemize}

\vspace{-5pt}

We evaluate each method on the problem sets and report the average results in Table~\ref{table:sim-results}, where \emph{Solved \%} is the percentage of problems that were successfully solved, or the proportion of impossible tasks that were correctly identified by returning an empty plan,
and \emph{planning budget} (PB) is defined as the number of plans attempted before solving a problem (adopted from~\cite{liang2025visualpredicator}).
Impossible problems probe soundness, since a \textit{sound} model cannot admit spurious abstract plans for goals that are infeasible in the environment.
We find that \sys{} outperforms all baselines without access to privileged knowledge, and even surpasses the performance of the System Predicates baseline on hard problems. 
Due to space constraints, we defer case studies, failure modes, and additional analysis to Appendix~\ref{appendix:case_study}.



\begin{table*}[b]
\vspace{-10pt}
    \centering
    \begin{minipage}[t]{0.49\textwidth}
        \centering
        \begin{adjustbox}{width=\linewidth}
        \begin{tabular}{lcc|cc|c}
        \toprule
        Method & \multicolumn{2}{c|}{Easy} & \multicolumn{2}{c|}{Hard} & \multirow{2}{*}{Impossible} \\
        & \multicolumn{1}{c}{Solved \% $\uparrow$} & \multicolumn{1}{c|}{PB $\downarrow$} 
        & \multicolumn{1}{c}{Solved \% $\uparrow$} & \multicolumn{1}{c|}{PB $\downarrow$} & \\
        \midrule
        Expert Ops. & \textbf{81.0} $\pm$ \textbf{3.7} & \textbf{1.9} $\pm$ \textbf{0.4} & \textbf{58.1} $\pm$ \textbf{3.9} & \textbf{4.2} $\pm$ \textbf{0.4} & \textbf{100} $\pm$ \textbf{0}\\
        Sys Preds. & \textbf{79.0} $\pm$ \textbf{3.7} & 2.6 $\pm$ 0.2& 22.0 $\pm$ 12.9 & 7.8 $\pm$ 1.3 & 42.0 $\pm$ 7.5 \\
        \midrule
        ViLa & 46.0 $\pm$ 16.2 & - & 13.9 $\pm$ 11.6 & - & 20.0 $\pm$ 10.9\\
        Random & 4.0 $\pm$ 2.0 & 9.6 $\pm$ 0.2 & 0 $\pm$ 0 & 10.0 $\pm$ 0 & 100 $\pm$ 0\\
        FM Invent & 23.3 $\pm$ 16.5 & 7.7 $\pm$ 1.7& 1.7 $\pm$ 2.4 & 9.7 $\pm$ 0.2 & 63.3 $\pm$ 26.2\\
        No Heuristic & \textbf{76.0} $\pm$ \textbf{4.9} & \textbf{2.5} $\pm$ \textbf{0.9} & 24.0 $\pm$ 19.6 & 7.8 $\pm$ 1.8 & 80 $\pm$ 20.9 \\
        \textbf{Ours} & \textbf{74.0} $\pm$ \textbf{3.7}& \textbf{2.7} $\pm$ \textbf{0.4}& \textbf{40.0} $\pm$ \textbf{3.2}& \textbf{6.3} $\pm$ \textbf{0.4} & \textbf{100} $\pm$ \textbf{0} \\
        \bottomrule
        \end{tabular}
        \end{adjustbox}
        \caption{Results in Robotouille Environment}
        \label{table:sim-results}
    \end{minipage}
    \hfill 
    \begin{minipage}[t]{0.49\textwidth}
        \centering
        \begin{adjustbox}{width=\linewidth}
        \begin{tabular}{lcc|cc|c}
        \toprule
        \multirow{2}{*}{Method} & \multicolumn{2}{c|}{In-domain} & \multicolumn{2}{c|}{Generalization} & \multirow{2}{*}{Impossible} \\
        & Solved \% $\uparrow$ & PB $\downarrow$ & Solved \% $\uparrow$ & PB $\downarrow$ & \\
        \midrule
        Expert Ops.  & 66.7 $\pm$ 9.4 & 3.3 $\pm$ 0.9 & 53.3 $\pm$ 9.4 & 5.3 $\pm$ 0.9& 46.7 $\pm$ 0\\
        \midrule
        ViLa  & 46.7 $\pm$ 9.4 & - & 6.7 $\pm$ 9.4& - & 6.7 $\pm$ 9.4\\
        Random  & 0 $\pm$ 0 & 10.0 $\pm$ 0 & 0 $\pm$ 0 & 10.0 $\pm$ 0 & 60 $\pm$ 16.3\\
        FM Invent & 26.7 $\pm$ 18.9 & 7.3 $\pm$ 1.9  & 13.3 $\pm$ 9.4 & 8.7 $\pm$ 0.9 & 86.7 $\pm$ 9.4\\
        No Heuristic & 53.3 $\pm$ 9.4  & 4.7 $\pm$ 0.9 & 40.0 $\pm$ 16.3 & 6.0 $\pm$ 1.6& 53.3 $\pm$ 9.4 \\
        \textbf{Ours} & \textbf{76.7} $\pm$ \textbf{9.4} & \textbf{2.7}  $\pm$ \textbf{0.9} & \textbf{60.0} $\pm$ \textbf{0} & \textbf{4.0} $\pm$ \textbf{0} & \textbf{66.7} $\pm$ \textbf{9.4} \\
        \bottomrule
        \end{tabular}
        \end{adjustbox}
        \caption{Results of Generalization Experiment}
        \label{table:franka}
    \end{minipage}
    \vspace{-5pt}
\end{table*}

\subsection{Robot Experiments}
To evaluate the applicability of \sys{} for real-world settings, we conduct two sets of experiments on two robotic platforms: a tabletop Franka Emika Panda and a bimanual manipulator with two Kuka iiwa arms.
These experiments investigate compositional generalization and data efficiency, respectively.
%
See our supplementary materials for video demonstrations of both experiments, and
Appendix~\ref{app:robot-implementation} for further details about our experimental setup.

\textbf{Generalization}\quad In this setting, a tabletop Panda manipulator has a skill set $\Omega$ consisting of five skills, and the object set $\mathcal{O}$ contains five objects (see Fig.~\ref{fig:overview}(a)).
To evaluate the compositional generalization ability of the learned planning abstractions, we design three smaller training environments, each containing a subset of $\mathcal{O}$, and thus only a subset of $\Omega$ is executable. Each training environment contains fewer than 10 possible abstract states under the expert-defined model.
After running \sys{} for one iteration in each environment, we port the learned model to a test environment containing all objects in $\mathcal{O}$, inducing 34 possible abstract states.
We prepare an evaluation set consisting of five problems in the training environments, five problems in the test environment, and five \textit{impossible} problems.
We report the results of this experiment in Table~\ref{table:franka}.

\begin{figure}[t]
    \centering
    \includegraphics[width=1.0\linewidth]{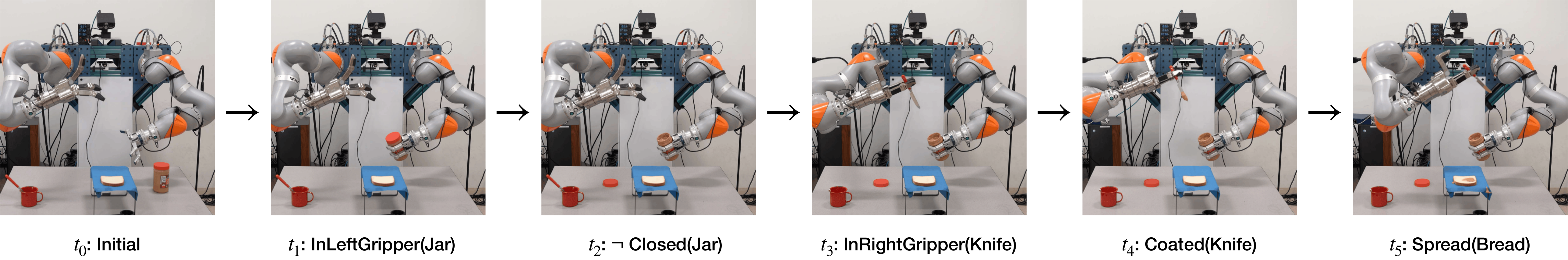}
    \caption{Example predicate truth values during a sequence of bimanual robot skill executions.}
    \vspace{-10pt}
    \label{fig:dorfl_sequence}
\end{figure}

\begin{wrapfigure}[13]{r}{0.5\textwidth}
    \centering
    \vspace{-15pt}
    \begin{minipage}[c]{0.5\textwidth}
    \includegraphics[width=\linewidth]{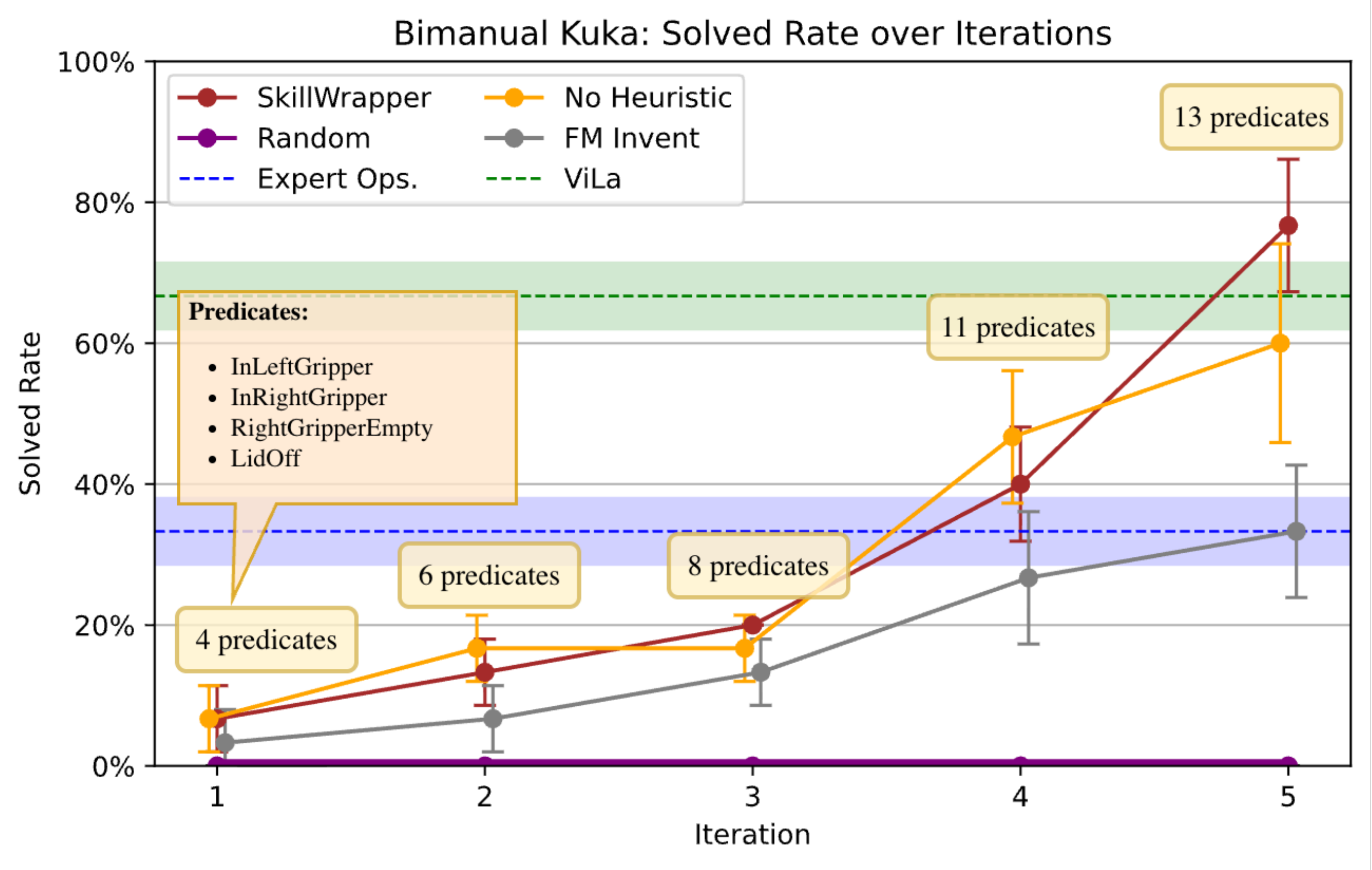}
    \vspace{-15pt}
    \caption{Bimanual manipulation results ($n=3$).
    }
    \label{fig:dorlf_result}
    \end{minipage}
\end{wrapfigure}

\textbf{Data efficiency}\quad
In this setting, a bimanual manipulator robot is equipped with six skills in an environment containing three objects.
Notably, this environment contains multiple dead ends, hindering data collection.
This design is intended to test the data efficiency of \sys{} over multiple learning iterations.
Figure~\ref{fig:dorfl_sequence} depicts an example sequence of predicate truth value changes induced by a skill sequence.
As shown in Figure~\ref{fig:dorlf_result}, \sys{} improves in performance as it collects additional data and invents new predicates, surpassing all baselines within five iterations.

\textbf{Discussion}\quad
\sys{} performs competitively with the Expert Operators while requiring only a handful of exploratory interactions, and generalizes well, suggesting that the invented predicates and learned model capture meaningful abstractions rather than overfitting.
The gap between \sys{} and ViLa demonstrates the benefit of an explicit symbolic model over naive FM prompting;
the gap against Random Exploration and No Heuristic shows the importance of guided data collection;
and the gap against FM Invent shows that our predicate invention algorithm, not the FM alone, enables efficient model learning. Together, these results indicate that \sys{} achieves a favorable combination of data efficiency and model fidelity.
Further analysis is in Appendix ~\ref{appendix:case_study}.

Qualitatively, we observe that predicate invention is primarily driven by preconditions (Eq.~\ref{eq:precond-invention}), because any individual skill effect is typically necessary to represent only when another skill's precondition depends on it.
Effect-driven predicate invention (Eq.~\ref{eq:effect-invention}) is the necessary secondary mechanism for terminal skills without downstream dependents; without it, tasks like spreading peanut butter (upon which, in our setting, no other skill depends) could not be modeled.

These results show that \sys{} is effective on real robots: planning models learned in restricted domains generalize to larger problems, and can be progressively improved even in settings with irreversible actions.
By outperforming all non-privileged baselines, and performing competitively against hand-specified or privileged baselines, \sys{} demonstrates the importance of predicate invention and active learning in enabling symbolic representations to scale to embodied tasks.


%% file: sections/related_work.tex
%
\sys{} draws ideas from different fields of research, such as model learning, abstraction learning, and task and motion planning (TAMP). 
Several methods have used FMs as high-level planners~\citep{saycan_corl,krishan2023sayplan,driess2023palme}.
Recent work has also explored how representations can be learned directly from pixels~\citep{athalye2026pix2pred}.
Lastly, multiple approaches~\citep{han2024interpret, liang2025visualpredicator} have used foundation models to learn symbolic representations of robot skills, but require extensive feedback or prior knowledge from human experts. 
%
%
TAMP has long been used to solve complex robot tasks~\citep{dantam2018incremental,shah2020anytime,garrett2021integrated}. However, these approaches require symbolic models of robot skills for task planning. Various approaches have been developed to learn such symbolic models compatible with TAMP solvers from high-dimensional inputs~\citep{Konidaris2018FromST,silver2023predinvent,shah2024reals}.
Additionally, the abstract representations learned through these methods are often not directly human-interpretable. 
\sys{} also connects to other domains in robotics, and the full related work can be found in Appendix~\ref{appendix:full_related_work}.

%% file: sections/conclusion.tex


We characterize formal conditions under which abstract models support sound and complete task-level planning, and use them to derive principled criteria for generative predicate invention in \sys{}. Our method autonomously learns human-interpretable symbolic representations of black-box skills and produces operators directly usable by classical planners, without any predefined predicates, operators, or tasks for data collection. 
As foundation models improve in commonsense reasoning and visual grounding, \sys{} can naturally benefit from these advances, while its formally derived criteria continue to direct predicate invention toward the correct abstractions.
Empirical results in simulation and on real robots show that \sys{} enables efficient long-horizon planning without hand-engineered abstractions, offering a promising path toward scalable symbolic model learning.

%% file: sections/appendix.tex
%


\section{Algorithms}
\label{append:pseudocodes}
\input{sections/appendix/algorithms}

\section{Properties of Learned Models}\label{appendix:theory}
\input{sections/generative_predicate_invention}
\section{Proofs}
\input{sections/appendix/proofs}

\section{Additional Details on the Hypothesis Class and Sample Complexity}
\label{appendix:hypothesis_class}
\input{sections/appendix/upper_bound}

\section{Comparisons of Symbolic Model Learning Methods}
\label{appendix:comparison_table}
\input{sections/appendix/comparison_table}

\section{Operator Learning Details}
\label{appendix:operator_learning}
\input{sections/appendix/operator_learning}

\clearpage

\section{Case Studies, Learned Operators, and Example Tasks}
\label{appendix:case_study}
\input{sections/appendix/learned_operators}

\section{VLM Reliability Study}
\label{appendix:vlm_study}
\input{sections/appendix/vlm_study}

\section{Implementation Details}
\input{sections/appendix/implementation_details}

\section{Related Works}
\label{appendix:full_related_work}
\input{sections/appendix/full_related_work}

\section{Glossary of Notation}
\label{app:notation-glossary}
\input{sections/appendix/glossary}
%
%

%% file: sections/appendix/algorithms.tex
\begin{algorithm}[H]
    \small
    \caption{{\sys}}
    \label{algo:main}
        \begin{algorithmic}[1]
            \State \textbf{Input:}
            Set of skills $\Skills$,
            number of iterations $m \in \mathbb{N}_1$
            
            \State \textbf{Output:}
            Abstract transition model $\Model = \DefAbstractModel$
            
            \State $\mathcal{D}, \Predicates, \Operators \gets \emptyset$
            \For{$i \in \{ 1, \dots, m \}$}
                \State $\SkillPlan \gets \Call{ProposeSkillSeq}{\Skills, \cD, \Predicates, \Operators}$ \Comment{Section~\ref{sec:data-collection}}
                \State $\cD \gets \cD \bigcup \Call{ExecuteSkills}{\bm{\Ground{\skill}}}$
                \State $\Predicates \gets \Call{InventPredicates}{\Skills, \mathcal{D}, \Predicates, \Operators}$ \Comment{Section~\ref{sec:predicate_invention}}
                \State $\Operators \gets \Call{LearnOperators}{\mathcal{D},\Predicates}$ \Comment{Section~\ref{subsec:operatorlearning}}
                
            \EndFor
            \State \Return $\Model  = \DefAbstractModel$
        \end{algorithmic}
    \end{algorithm}
\subsection{Active Data Collection}~\\
\vspace{-10pt}
\label{app:skillseq-heuristics}

\vspace{-10pt}
\begin{algorithm}[H]
\begin{small}
\caption{Propose Skill Sequences}
\label{alg:skillseqproposal}\begin{algorithmic}[1]
    \State \textbf{Input:} Skill set $\Skills$, dataset $\mathcal{D}$, predicate set $\mathcal{P}$, abstract actions $\mathcal{A}$, batch size $n$
    
    \State \textbf{Output:} Proposed skill sequence $\SkillPlan$

    \State $\mathrm{seq\_batch} \gets \Call{GenerateSkillSequences}{\Omega, n}$
    \State $\mathrm{Scores} \gets \{\}$

    \For{$\SkillPlan$ \textbf{in} $\mathrm{seq\_batch}$}
    
        \State $b \gets \Call{Balance}{\SkillPlan}$
    
        \State $c\gets \Call{Coverage}{\SkillPlan}$

        \State $\mathrm{Scores}[\SkillPlan] \gets (b, c)$
    \EndFor
    \State \Return $\Call{ParetoOptimal}{\mathrm{Scores}}$
\end{algorithmic}
\end{small}

\end{algorithm}

\begin{algorithm}[H]
\begin{small}
\caption{Coverage}
\label{alg:coverage}
\begin{algorithmic}[1]
 \State \textbf{Input:} Dataset $\mathcal{D}$, proposed skill sequence $\SkillPlan$
 \State \textbf{Output:} Coverage score $c$
    
    \State $\mathcal{Q} \gets \text{zero matrix of size } |\Omega| \times |\Omega|$ \Comment{Construct a matrix of skill-pair counts}

    \For{$\langle s_i, \underline{\omega}_i, s'_i \rangle,\langle s_{i+1}, \underline{\omega}_{i+1}, s'_{i+1} \rangle$ in $\mathcal{D}$} \Comment{Iterate over all consecutive pairs of transitions}

        \State $\mathcal{Q}[\underline{\omega}_{i}, \underline{\omega}_{i+1}] \gets \mathcal{Q}[\underline{\omega}_{i}, \underline{\omega}_{i+1}] + 1$
        
    \EndFor

    \State $\mathcal{Q^{'}} \gets \mathcal{Q}$ 
    \Comment{New skill-pair count initialized}

    \For{$\langle \underline{\omega}_{i}, \underline{\omega}_{i+1} \rangle$ in $\SkillPlan$}

    \State $\mathcal{Q^{'}}[\underline{\omega}_{i}, \underline{\omega}_{i+1}] \gets \mathcal{Q^{'}}[\underline{\omega}_{i}, \underline{\omega}_{i+1}] + 1$
    \EndFor

    \State $c \gets \Call{Entropy}{\mathcal{Q}'} - \Call{Entropy}{\mathcal{Q}}$
    \State \Return $c$

\end{algorithmic}
\end{small}
\end{algorithm}

\begin{algorithm}[H]
\small
\caption{Balance}
\label{alg:chainability}
\begin{algorithmic}[1]
    \State \textbf{Input:} Operator set $\mathcal{A}$, proposed skill sequence $\SkillPlan$, target ratio $r^*$
    \State \textbf{Output:} Balance score $b$

    \State $\mathrm{exec\_count} \gets 0$
    \Comment{Total number of executable skills}
    \State $\mathrm{sequence\_length} \gets \Call{Length}{\SkillPlan}$


    \State $\mathrm{seq} \gets \{\AbstractionF(s_0)\}$
    \Comment{Store the trace of after-execution state}
    \For{$\Ground{\skill}_i$ in $\SkillPlan$}
    \State{$\AbstractPlan \gets \{\}$}
        \For{$\Ground{\operator} \in \Ground{\Operators}_{\Ground{\skill}_i}$}
        \If{$\AbstractionF(\mathrm{seq}[-1]) \models \GroundPRE[\Ground{\operator}]$}
            \Comment{Successful execution predicted by the current model}
            \State{$\AbstractPlan \gets \AbstractPlan \bigcup \{\Ground{\operator} \}$}
    
                \EndIf

            \EndFor
        \State{$\hat{\Ground{\operator}} = \arg\max_{\Ground{\operator}\in \AbstractPlan}
|\PRE[\Ground{\operator}]|$}
        
        \State $\abstractState_{new} \gets \Call{ApplyOperator}{\hat{\Ground{\operator}}}$
        \Comment{Calculate the abstract state after execution}
        \State $\mathrm{seq} \gets \mathrm{seq} \bigcup \{\abstractState_{new}\}$
        \Comment{Append current low-level state to the trace}
        \State $\mathrm{exec\_count \gets exec\_count + 1}$
        
    \EndFor
    \State $\hat{r} \gets  \mathrm{exec\_count} / \mathrm{sequence\_length}$
    \State $b \gets - |\hat{r} - r^*| $

    \State \Return $b$

\end{algorithmic}
\end{algorithm}


\subsection{Predicate Invention}~\\
\vspace{-10pt}

\begin{algorithm}[H]
\begin{small}
\caption{Invent Predicates}
\label{algo:predicate}
\begin{algorithmic}[1]
    \State \textbf{Input:} Skill set $\Skills$, transition dataset $\mathcal{D}(\skill) = \{\langle\state, \Ground{\skill}, \state'\rangle\}_{\omega}$, existing predicate set $\Predicates$, abstract actions $\Operators$.
    \State \textbf{Output:} Predicate set $\Predicates$
    
    \For{$\skill \in \Skills$}
        \While{precondition inconsistency exists} \Comment{\eqref{eq:precond-invention}}
            \State $\predicate \gets \Call{NewPredicate}{}$
            \State $\Predicates \gets \Predicates \bigcup \{\predicate\}$ \textbf{if} \Call{ScorePrecond}{$\predicate, \Predicates, \skill, \cD$}
        \EndWhile
        
    \While{effect inconsistency exists} \Comment{\eqref{eq:effect-invention}}
            \State $\predicate \gets \Call{NewPredicate}{}$
            \State $\Predicates \gets \Predicates \bigcup \{\predicate\}$ \textbf{if} \Call{ScoreEff}{$\predicate, \Predicates, \skill, \cD$}
        \EndWhile
    \EndFor

    \State \Return $\mathcal{P}$
\end{algorithmic}
\end{small}
\end{algorithm}

\begin{algorithm}[H]
\caption{Scoring Functions for Invented Predicates}
\begin{small}
\begin{algorithmic}[1]
\State \textbf{Input:} Candidate predicate $\hat{\predicate}$, existing predicate set $\Predicates$, skill $\skill$, and transition dataset $\cD$.
\State \textbf{Parameters:} Threshold $h$

\Statex
\State \textbf{ScorePrecond:}
\State $\hat{\Predicates} \gets \Predicates \bigcup \{ \hat{\predicate} \}$
\State $\hat{\Operators} \gets \Call{LearnOperators}{\mathcal{D}, \hat{\Predicates}}$
\Comment{Hypothetical operators after including $\hat{\predicate}$}
\State $\mathrm{total} \gets 0$
\State $\mathrm{valid} \gets 0$
    \For{$\langle\state, \Ground{\skill},\state' \rangle\ \in \mathcal{D}$}
    
    \If{$\exists\, \hat{\Ground{\operator}} \models \hat{\Ground{\Operators}_{\Ground{\skill}}} \text{ , s.t. } \AbstractionF{(\state)} \in \GroundPRE[\hat{\Ground{\operator}}] \text{ , and } \state \in \PreSuccessSet[\Ground{\skill}]$}
        \State $\mathrm{valid \gets valid} + 1$
    \EndIf

    \If{$\not\exists\, \hat{\Ground{\operator}} \models \hat{\Ground{\Operators}_{\Ground{\skill}}} \text{ , s.t. } \AbstractionF{(\state)} \in \GroundPRE[\hat{\Ground{\operator}}] \text{ , and } \state \notin \PreSuccessSet[\Ground{\skill}]$}
        \State $\mathrm{valid \gets valid} + 1$
    \EndIf
    
    \State $\mathrm{total \gets total} + 1$
    \EndFor

\State \Return $\mathrm{valid}/\mathrm{total} > h$

\Statex
\State \textbf{ScoreEff:}
\State $\hat{\Predicates} \gets \Predicates \bigcup \{\predicate\}$
\State $\hat{\Operators} \gets \Call{LearnOperators}{\mathcal{D}, \hat{\Predicates}}$
\State $\mathrm{total} \gets 0$
\State $\mathrm{valid} \gets 0$
    \For{$\langle\state, \Ground{\skill},\state'\rangle\ \in \mathcal{D}$}
    
    \If{$\exists\, \hat{\operator} \in \hat{\Operators} \text{ , s.t. } \AbstractionF_{\hat{\Predicates}}(\mathrm{s'})\setminus\AbstractionF_{\hat{\Predicates}}(\mathrm{s}) = \GroundEFF[\hat{\Ground{\operator}}] \text{ , and } \mathrm{s} \in \SuccessSet[\underline{\omega}]$}
        \State $\mathrm{valid \gets valid} + 1$
    \EndIf
    
    \If{$\not\exists\, \hat{\operator} \in \hat{\Operators} \text{ , s.t. } \AbstractionF_{\hat{\Predicates}}(\mathrm{s'})\setminus\AbstractionF_{\hat{\Predicates}}(\mathrm{s}) = \GroundEFF[\hat{\Ground{\operator}}] \text{ , and } \mathrm{s} \notin \SuccessSet[\underline{\omega}]$}
        \State $\mathrm{valid \gets valid} + 1$
    \EndIf
    
    \State $\mathrm{total \gets total} + 1$
    \EndFor

\State \Return $\mathrm{valid}/\mathrm{total} > h$

\end{algorithmic}
\end{small}
\label{appendix:predicate_invention}
\end{algorithm}

\subsection{Operator Learning}~\\
\vspace{-10pt}

\begin{algorithm}[H]
\begin{small}
\caption{Learn Operators}
\label{algo:cluster-operators}
\begin{algorithmic}[1]
    \State \textbf{Input:} Dataset $\mathcal{D}(\skill) = \{\langle \state, \Ground{\skill},\state'\rangle\}_{\omega}$, predicate set $\Predicates$
    \State \textbf{Output:} Abstract actions $\Operators$
    \State $\mathrm{eff\_dict} \gets \mathrm{defaultdict}()$ \Comment{Store clustered effects}

    \For{each successful transition $\tau = \langle\state, \Ground{\skill},\state'\rangle\ \in \mathcal{D}$}
        \State $\EFF[] \gets \LiftF(\Delta(s,s'))
        $
        \State $\mathrm{eff\_dict}[\EFF[]] \gets \mathrm{eff\_dict}[\EFF[]] \bigcup \{\langle\state, \Ground{\skill},\state'\rangle\}$
    \EndFor

    \State $\mathcal{A} \gets \emptyset$
    \For{$\EFF[] \in \mathrm{eff\_dict}$}
        \State $\PRE[] \gets \Pi_{\langle\state, \Ground{\skill},\state'\rangle\ \in \mathrm{eff\_dict}[\EFF[]]} \AbstractionF(\state)$
        \State $\operator \gets \langle \PRE[], \EFF[] \rangle$
        \State $\Operators \gets \Operators \bigcup \{ \operator \}$
    \EndFor

    \State \Return $\mathcal{A}$
\end{algorithmic}
\end{small}
\end{algorithm}


%% file: sections/generative_predicate_invention.tex
Relational predicates are the basic units of the abstract representation of the low-level state space.
In this section, we characterize the conditions of the learned representations of a finite set of skills using relational predicates to support high-level planning, in the context of model learning.
From here, chaining the skills is enabled by applying predicates from the representation of each low-level skill to others for clustering (described in Section~\ref{subsec:operatorlearning}).

Effective skill planning requires an accurate abstract model and grounding functions. 
All forms of abstractions are typically lossy, i.e., while learning an abstract transition model, certain low-level environment details may not be captured. 
Conversely, the learned model must accurately retain the information needed to produce plans to guarantee soundness and completeness.
In this section, we characterize the conditions under which an abstract model facilitates correct, sound, and complete planning, for the purpose of constraining how we can \textit{construct} such a model.

To begin formalizing the relationship between a skill $\skill \in  \Skills$ and some abstract model $\Model$, we define two sets, $\EstInitiationSet[\skill]$ and $\EstTerminationSet[\skill]$, representing the states in which the model predicts that the successful skill execution may either be initiated (when $s \in \EstInitiationSet[\skill]$) or terminated (when $s \in \EstTerminationSet[\skill]$), respectively.

\begin{definition}
\label{def:alpha_zeta}
Given a skill instance $\skillInstance \in \Ground{\Skills}$ and abstract model $\Model$, we define $\EstInitiationSet[\skillInstance]$ and $\EstTerminationSet[\skillInstance]$ as follows:
\begin{align}
    \EstInitiationSet[\skillInstance] &\coloneqq \bigcup_{\abstractAction \in \AbstractActions_{\skillInstance}} 
    \GroundingF{(\GroundPRE[\abstractAction])} \\
    \EstTerminationSet[\skillInstance] &\coloneqq \bigcup_{\abstractAction \in \AbstractActions_{\skillInstance}} \GroundingF{\big( (\GroundPRE[\abstractAction] \setminus \GroundDeleteEFF[\abstractAction]) \cup \GroundAddEFF[\abstractAction] \big)}
\end{align}

For a non-instantiated skill $\skill \in \Skills$, we define $\EstInitiationSet[\skill] = \cup_{\skillInstance} \, \EstInitiationSet[\skillInstance]$ and $\EstTerminationSet[\skill] = \cup_{\skillInstance} \, \EstTerminationSet[\skillInstance]$ over all possible skill instances with type-compatible objects.
\end{definition}

We define a \emph{correct} abstract model as one that perfectly captures the pre-success set and success set of all skills. 

\begin{definition}[Correct Model] Let $\gM = \left(\gP, \gA \right)$ be a model for environment $\DefDomain$ and the set of skills $\Skills$, and let $\EstInitiationSet$ and $\EstTerminationSet$ be the approximate pre-success and success sets (Def.~\ref{def:alpha_zeta}). The model $\gM$ is a \textbf{correct model} iff:
\begin{equation}
    \forall \skill \in \Skills, \state \in \States \colon \big( \state \in \PreSuccessSet \iff \state \in \EstInitiationSet[\skill] \big) \land \big( \state \in \SuccessSet \iff \state \in \EstTerminationSet[\skill] \big).
\end{equation}

\end{definition}

A correct model exactly captures the pre-success and success sets of all skills, making correctness an idealized target for representation learning. We next define the plan-level properties of \textit{soundness} and \textit{completeness} and relate them to correctness.



A symbolic model is \emph{sound} if it correctly predicts the effects of a plan: 
whenever a complete and sound planner predicts that a sequence of skills will reach some abstract state, 
executing the corresponding skills in the environment truly leads there. 
Soundness rules out spurious symbolic transitions that do not correspond to realizable outcomes. Formally, we introduce the settings as follows: Let $\Model = (\Predicates, \Operators)$ denote an abstract transition model, where 
$\Predicates$ is a finite set of predicates defining an \emph{abstract state space} $\AbstractStates$, 
and $\Operators$ is a set of \emph{abstract actions} with preconditions and effects expressed in terms of $\Predicates$. 
Each abstract state $\abstractState \in \AbstractStates$ is obtained by an \emph{abstraction function} 
$\AbstractionF: \States \to \AbstractStates$ that maps low-level states $\state \in \States$ 
to truth assignments over $\Predicates$.
Given a skill planning problem $\mathbf{p}$, a symbolic planner can produce an abstract plan $\AbstractPlan$ using the learned model $\Model$, and the skill plan $\SkillPlan$ is then obtained by mapping all operators in $\AbstractPlan$ back to their corresponding skills.

\begin{definition}[Soundness]
    \label{def:soundenss}
    For an abstract plan $\mathbf a=[a_1,\ldots,a_k]$ with corresponding skill plan $\boldsymbol\omega=[\omega_1,\ldots,\omega_k]$ for a planning problem $\mathbf{p} = \DefSkillPlanningProblem$, let $s_i=T(s_{i-1},\omega_i)$. The model $\Model$ is \emph{sound} iff, for every $i$, 
    \[s_{i-1}\in I_{\omega_i} \quad\text{and}\quad \alpha(s_i)=\bar T_M(\alpha(s_{i-1}), a_i),\]

where
$T_M(\alpha(s_{i-1}), a_i)$ is the abstract state predicted by $\Model$ after executing $a_i$ 
from $\alpha(s_{i-1})$. 
We define soundness over abstract plans whose execution does not encounter an effect-ambiguous state, in which multiple operators of the same skill are simultaneously applicable but predict distinct effects: 
$ \exists a_i,a_j\in\mathcal A_\omega: \quad s^\uparrow\models\PRE[a_i],\; s^\uparrow\models\PRE[a_j],\; \EFF[a_i]\neq\EFF[a_j]. $
\end{definition}

A symbolic model is \emph{complete} if it never omits real solutions: 
whenever the environment admits a way to solve a skill planning problem, 
the planner can find a corresponding abstract plan in the model. 
Completeness rules out gaps in symbolic coverage that would make feasible problems appear unsolvable.

\begin{definition}[Completeness]
    The model $\mathcal{M}$ is \emph{complete} iff, for any skill planning problem 
$\mathbf{p} = \DefSkillPlanningProblem$ and any skill plan $\SkillPlan$ that achieves the goal states $\States_g$ from an initial state $\state_0$, 
there exists an abstract plan $\AbstractPlan$ over $\Model$ such that
\[
\bar{T}_{\Model}(\AbstractPlan, \AbstractionF(\state_0)) \models \AbstractionF(\States_g), \quad\alpha(S_g)\coloneqq\{\alpha(s):s\in S_g\}.
\]
\end{definition}

Although methods for constructing such models exist and have been investigated in previous work in the bottom-up setting, no existing work in the generative predicate invention setting provides or even discusses these properties in their methods.
We next establish the relationship between these model properties and analyze how \sys{} approaches them through its predicate-invention and operator-learning procedures.

%% file: sections/appendix/proofs.tex
%
\label{app:proofs}

\subsection{Establishing Soundness and Completeness}
We now relate the correctness property defined in Appendix~\ref{appendix:theory} to soundness and completeness.
\sys{} can represent distinct successful conditional effects of the same skill
using separate operators. When the preconditions of such operators overlap,
multiple operators may be simultaneously applicable while predicting different
effects, in which case the abstract model does not uniquely specify the realized
transition. As defined in Appendix~\ref{appendix:theory}, we therefore
consider soundness over effect-unambiguous plans. Under this definition,
correctness is sufficient to establish both soundness and completeness.


\setcounter{proposition}{0}
\begin{proposition}[Correctness implies soundness and completeness]
\label{prop:correctness}
Any correct abstract model
$\cM$ learned by \sys{} is sound and complete.
\end{proposition}
\noindent\textit{Proof.}
Since $\cM$ is correct, for every skill $\omega$ its estimated
pre-success and success sets exactly match the true sets:
\[
    \Phi_\omega = I_\omega,
    \qquad
    \Xi_\omega = \beta_\omega.
\]

For soundness, consider any valid abstract plan
$\mathbf{a}=[a_1,\ldots,a_n]$ and its corresponding skill sequence
$\boldsymbol{\omega}=[\omega_1,\ldots,\omega_n]$, starting from
$s_0$. Since $a_i$ is applicable at $\alpha(s_{i-1})$,
$s_{i-1}\in\Gamma(\PRE[a_i])\subseteq\Phi_{\omega_i}$.
By correctness, $s_{i-1}\in I_{\omega_i}$, so each corresponding
skill execution succeeds. 
By construction, \sys{} first clusters successful transitions by their abstract effects and then computes each operator's preconditions from the initial states of the same cluster. Thus, the precondition and effect of each operator remain associated through their supporting transitions; under effect-unambiguity, an applicable operator predicts the corresponding abstract effect.
Thus, for every step,
\[
    \alpha(s_i)
    =
    \bar{T}_{\cM}(\alpha(s_{i-1}), a_i),
    \qquad
    s_i=T(s_{i-1},\omega_i).
\]
Applying this correspondence inductively over the sequence shows that
the abstract plan correctly predicts its corresponding skill
execution, establishing soundness.

For completeness, consider any feasible skill plan
$\boldsymbol{\omega}=[\omega_1,\ldots,\omega_n]$ that reaches a goal
state from $s_0$. At every step, feasibility gives
$s_{i-1}\in I_{\omega_i}=\Phi_{\omega_i}$, so by the definition of
$\Phi_{\omega_i}$ there exists an operator
$a_i\in\mathcal{A}_{\omega_i}$ whose preconditions are satisfied.
Since the execution is successful and \(\beta_{\omega_i}=\Xi_{\omega_i}\), its successful abstract effect is represented by an operator learned from the corresponding effect cluster, whose preconditions are derived from that same cluster. Choosing this corresponding operator at each step constructs an abstract plan \([a_1,\ldots,a_n]\) for the feasible skill plan.
Its final abstract state is $\alpha(s_n)\in\alpha(S_g)$, and hence the model is complete. \qed

Having established how correctness relates to soundness and completeness, we next analyze how \sys{} approaches correctness through its predicate-invention conditions and learned operators.

\subsection{Properties of \sys{}}

\setcounter{lemma}{0}
\begin{lemma}
\label{lemma:consistency_full}
Let $\cM$ be the model learned by \sys{} from dataset $\cD$
after all contrastive transition pairs satisfying the
predicate-invention conditions have been resolved.
Then, for every observed transition
$\Tuple{\state,\skillInstance,\state'}\in\cD$, the estimated
success set agrees with the true success set at $\state'$:
\[
    \state' \in \EstTerminationSet[\skillInstance]
    \iff
    \state' \in \SuccessSet[\skillInstance].
\]
\end{lemma}

\begin{proof}
Suppose, for contradiction, that the equivalence does not hold for some
observed transition
$\tau=\Tuple{\state,\skillInstance,\state'}\in\cD$.
Then exactly one of the following two cases must hold.

First, suppose
\(\state'\in \EstTerminationSet[\skillInstance]\setminus \SuccessSet[\skillInstance]\).
By the definition of \(\EstTerminationSet[\skillInstance]\), there exists an operator
\(\underline{\operator}\in\underline{\Operators}_{\skillInstance}\) such that

$$ \state' \in \GroundingF{\big((\PRE[\underline{\operator}]\setminus\GroundDeleteEFF[\underline{\operator}])\cup\GroundAddEFF[\underline{\operator}]\big)}. $$

Since \(a\) is learned from a cluster of successful transitions sharing the same lifted abstract effect, while \(\state'\notin\SuccessSet[\skillInstance]\), \(\tau\) is a failed transition whose abstract change is not distinguished from that successful effect. This yields a contrastive pair satisfying Condition~\ref{eq:effect-invention}, contradicting the assumption that all such pairs have been resolved.

Conversely, suppose
$\state'\in
\SuccessSet[\skillInstance]\setminus
\EstTerminationSet[\skillInstance]$.
Then $\tau$ is a successful transition. Operator learning places
$\tau$ in a successful effect cluster and constructs a corresponding
operator $a\in\mathcal{A}_{\skillInstance}$. By construction, the
STRIPS update of $a$ contains $\state'$, i.e.,
$\state' \in
\GroundingF{\big((\PRE[\underline{\operator}]\setminus\GroundDeleteEFF[\underline{\operator}])\cup\GroundAddEFF[\underline{\operator}]\big)}$.
Therefore
$\state'\in\EstTerminationSet[\skillInstance]$, again a contradiction.

Hence, for every
$\Tuple{\state,\skillInstance,\state'}\in\cD$,
\[
\state'\in\EstTerminationSet[\skillInstance]
\iff
\state'\in\SuccessSet[\skillInstance].
\]
\end{proof}

\setcounter{theorem}{0}
\begin{theorem}[\bf Empirical Consistency of \sys{}]
\label{thm:soundness_appendix_full}
Let $\cM$ be the model learned by \sys{} from dataset $\cD$
after all contrastive transition pairs satisfying the
predicate-invention conditions have been resolved.
Then, for every observed transition
$\Tuple{\state,\skillInstance,\state'}\in\cD$, the estimated
pre-success and success sets agree with their true counterparts:
\[
    \state\in\EstInitiationSet[\skillInstance]
    \iff
    \state\in\PreSuccessSet[\skillInstance],
    \qquad
    \state'\in\EstTerminationSet[\skillInstance]
    \iff
    \state'\in\SuccessSet[\skillInstance].
\]
\end{theorem}

\begin{proof}
We first show that the estimated pre-success set agrees with the true
pre-success set on every observed initial state. Suppose otherwise for
some transition
$\tau_j=\Tuple{\state_j,\Ground{\skill}_j,\state'_j}\in\cD$.
There are two cases.

\noindent\textbf{(i) False positive.}
Suppose
$\state_j\notin\PreSuccessSet[\Ground{\skill}_j]$ but
$\state_j\in\EstInitiationSet[\Ground{\skill}_j]$.
By the definition of $\EstInitiationSet$, some operator associated with
$\Ground{\skill}_j$ is applicable at $\state_j$. This operator is learned
from a cluster containing at least one successful transition
$\tau_i=\Tuple{\state_i,\Ground{\skill}_i,\state'_i}$ of the same skill,
for which
$\state_i\in\PreSuccessSet[\Ground{\skill}_i]$ and
$\state_i\in\EstInitiationSet[\Ground{\skill}_i]$.
Thus, $\tau_i$ and $\tau_j$ satisfy
Condition~\ref{eq:precond-invention}, contradicting the assumption that
all such contrastive pairs have been resolved.

\noindent\textbf{(ii) False negative.}
Suppose
$\state_j\in\PreSuccessSet[\Ground{\skill}_j]$ but
$\state_j\notin\EstInitiationSet[\Ground{\skill}_j]$.
Then $\tau_j$ is a successful transition. Under a fixed dataset $\cD$,
operator learning places every successful transition into an effect
cluster and constructs an operator whose preconditions are obtained by
intersecting the abstract initial states in that cluster. Hence the
initial state of every transition in the cluster satisfies the learned
preconditions, implying
$\state_j\in\EstInitiationSet[\Ground{\skill}_j]$, a contradiction.
Further details of this operator-learning property are given in
Appendix~\ref{app:operator_learning_proof}.

Appendix~\ref{sec:pred_invent-lowlevel} provides the complete
case-by-case analysis of the possible relations between
$\EstInitiationSet$ and $\PreSuccessSet$ and shows that the above two
cases exhaust all observed pre-success inconsistencies.

Therefore, for every observed transition,
$\state\in\EstInitiationSet[\skillInstance]$ iff
$\state\in\PreSuccessSet[\skillInstance]$.
By Lemma~\ref{lemma:consistency_full}, we also have
$\state'\in\EstTerminationSet[\skillInstance]$ iff
$\state'\in\SuccessSet[\skillInstance]$ on every observed terminal
state. Together, these establish the claimed empirical consistency on
$\cD$.
\end{proof}

To account for temporal dependence in actively collected data,
we model the transitions within each episode as a possibly
nonstationary $\beta$-mixing process, with dependence between
observations separated by $k$ steps bounded by
$\beta(k)$~\citep{yu1994rates}.
Within an episode, the active data-collection strategy is fixed.

For a fixed candidate model $\cM$ and transition $\tau=\Tuple{\state,\skillInstance,\state'}$, define the error
indicator and its episode-averaged expectation as
\[
\ell_{\cM}(\tau)\Def
\mathbf{1}\!\left\{
\state\in\PreSuccessSet[\skillInstance]\ominus\EstInitiationSet[\skillInstance]
\;\lor\;
\state'\in\SuccessSet[\skillInstance]\ominus\EstTerminationSet[\skillInstance]
\right\},
\qquad
d_{\mathrm{compl}}(\cM,\cM^*)\Def
\frac{1}{n}\sum_{t=1}^{n}\mathbb{E}[\ell_{\cM}(\tau_t)],
\]
where $\ominus$ denotes symmetric difference.
By Theorem~1, the learned model has zero empirical error on every
observed transition. Since the active data-collection strategy is fixed
within each episode and updated only between episodes, we first establish
a single-episode bound in Lemma~2, and then extend the result across
multiple episodes in Theorem~2.
\begin{lemma}
\label{lem:episode-generalization}
Let $\cD_n=\{\tau_t\}_{t=1}^{n}$ be collected within one episode
under the fixed within-episode $\beta$-mixing process described above,
and let $\cH$ be a fixed finite model hypothesis class.
For any $\epsilon\in(0,1)$ and integer $1\le k\le n$, define
$m=\lfloor n/k\rfloor$.
With probability at least
\[
1-|\cH|\left(e^{-m\epsilon}+(m-1)\beta(k)\right),
\]
every model $\cM\in\cH$ satisfying the empirical consistency
conditions of Theorem~1 on $\cD_n$ satisfies
\[
d_{\mathrm{compl}}(\cM,\cM^*)\le\epsilon.
\]
\end{lemma}

\begin{proof}
For any fixed $\cM\in\cH$, define
\[
    \mathrm{Err}(\cM) \coloneqq d_{\mathrm{compl}}(\cM,\cM^*),
\]
and let $X_i$ be the binary indicator of whether an error occurs on
the $i$-th transition:
\[
    X_i
    \coloneqq
    \ell_{\cM}(\tau_i)
    =
    \mathbf{1}\!\left[
        (s_i\in \PreSuccessSet[\skillInstance_i]
        \ominus \EstInitiationSet[\skillInstance_i])
        \,\vee\,
        (s_i'\in \SuccessSet[\skillInstance_i]
        \ominus \EstTerminationSet[\skillInstance_i])
    \right].
\]
The empirical error over the $n$ observed transitions is
\[
    \widehat{\mathrm{Err}}(\cM)
    \coloneqq
    \frac{1}{n}\sum_{i=1}^n X_i.
\]
By definition,
\[
    \mathrm{Err}(\cM)
    =
    \frac{1}{n}\sum_{i=1}^n \mathbb{E}[X_i].
\]
Unlike in the stationary case, the expectations $\mathbb{E}[X_i]$
need not be identical across transition indices.

Suppose that some bad hypothesis $\cM\in\cH$ satisfies
$\mathrm{Err}(\cM)>\epsilon$. If $\widehat{\mathrm{Err}}(\cM)=0$,
then $X_i=0$ for every $i=1,\ldots,n$.

To handle temporal dependencies among consecutive transitions, fix a
time lag $k\geq 1$ and let $m=\lfloor n/k\rfloor$. Write
$p_i=\mathbb{E}[X_i]$ and partition the transition indices into the
$k$ interleaved subsequences
\[
    J_r
    =
    \{r,r+k,r+2k,\ldots\}\cap\{1,\ldots,n\},
    \qquad r=1,\ldots,k.
\]
Since these subsequences partition all $n$ indices, there exists some
$J_{r^*}$ whose average error probability is at least the overall
average:
\[
    \frac{1}{|J_{r^*}|}
    \sum_{i\in J_{r^*}} p_i
    \geq
    \frac{1}{n}\sum_{i=1}^n p_i
    =
    \mathrm{Err}(\cM).
\]
Since $|J_{r^*}|\geq m$, retain $m$ indices
$t_1<\cdots<t_m$ in $J_{r^*}$ with the largest values of $p_i$.
Then $t_{j+1}-t_j\geq k$ and
\[
    \sum_{j=1}^m p_{t_j}
    \geq
    m\,\mathrm{Err}(\cM)
    >
    m\epsilon.
\]

Let $Y_j=X_{t_j}$ for $j=1,\ldots,m$. Because zero empirical error
requires every observed transition to be correct,
\[
    \Pr\!\left(\widehat{\mathrm{Err}}(\cM)=0\right)
    \leq
    \Pr(Y_1=0,\ldots,Y_m=0).
\]
Applying the $\beta$-mixing coupling inequality to this $k$-spaced
subsequence, we compare it with independent proxy variables
$\widetilde Y_1,\ldots,\widetilde Y_m$ having the same respective
marginal distributions:
\[
    \Pr(Y_1=0,\ldots,Y_m=0)
    \leq
    \Pr(\widetilde Y_1=0,\ldots,\widetilde Y_m=0)
    +(m-1)\beta(k).
\]
The proxy variables need not be identically distributed. Since
$\Pr(\widetilde Y_j=1)=p_{t_j}$, independence gives
\[
\begin{aligned}
    \Pr(\widetilde Y_1=0,\ldots,\widetilde Y_m=0)
    &=
    \prod_{j=1}^m (1-p_{t_j}) \\
    &\leq
    \exp\!\left(-\sum_{j=1}^m p_{t_j}\right) \\
    &\leq
    e^{-m\mathrm{Err}(\cM)}
    \leq
    e^{-m\epsilon}.
\end{aligned}
\]
Therefore, for any fixed $\cM\in\cH$ with
$\mathrm{Err}(\cM)>\epsilon$,
\[
    \Pr\!\left(\widehat{\mathrm{Err}}(\cM)=0\right)
    \leq
    e^{-m\epsilon}+(m-1)\beta(k).
\]

Applying a union bound over $\cH$, we obtain
\[
\Pr\!\left[
    \exists\,\cM\in\cH:
    \mathrm{Err}(\cM)>\epsilon
    \;\wedge\;
    \widehat{\mathrm{Err}}(\cM)=0
\right]
\leq
|\cH|
\left(
    e^{-m\epsilon}+(m-1)\beta(k)
\right).
\]
Taking the complement, with probability at least
\[
    1-
    |\cH|
    \left(
        e^{-m\epsilon}+(m-1)\beta(k)
    \right),
\]
every empirically consistent model $\cM\in\cH$ satisfies
\[
    d_{\mathrm{compl}}(\cM,\cM^*)\leq\epsilon.
\]
\end{proof}

We now extend the single-episode result to multiple episodes,
allowing the active data-collection strategy to change between
episodes. Let $\tau_{r,t}$ denote transition $t$ in episode $r$,
and let $\cD_{1:r-1}$ denote the preceding collection history.
Conditional on this history, the collection strategy within episode
$r$ is fixed. Define the average conditional error across $K$
episodes as
\[
d_{\mathrm{compl},K}(\cM,\cM^*)
\Def
\frac{1}{K}\sum_{r=1}^{K}
\mathbb{E}\!\left[
\frac{1}{n}\sum_{t=1}^{n}\ell_{\cM}(\tau_{r,t})
\,\middle|\,\cD_{1:r-1}
\right].
\]
Lemma~\ref{lem:episode-generalization} gives a generalization bound within a single episode. Since the data-collection policy is updated only between episodes, we condition on the history before each episode and apply the lemma episode-wise, accumulating the error and mixing terms across episodes to obtain the following result.

\setcounter{theorem}{1}
\begin{theorem}[\bf Asymptotic Convergence of {\sys}]
\label{thm:prob-completeness-full}
Let $\cH$ be a fixed finite model hypothesis class, and let
$\widehat{\cM}_K\in\cH$ satisfy the empirical consistency conditions
of Theorem~1 on all transitions from $K$ episodes of length $n$.
For any $\epsilon\in(0,1)$ and integer $1\le k\le n$, let
$m=\lfloor n/k\rfloor$.
With probability at least
\[
1-|\cH|\left(e^{-Km\epsilon}+K(m-1)\beta(k)\right),
\]
the learned model $\widehat{\cM}_K$ satisfies
\[
d_{\mathrm{compl},K}(\widehat{\cM}_K,\cM^*)\le\epsilon.
\]
In particular, choosing $k=n$ gives $m=1$ and eliminates the
mixing term, so
$d_{\mathrm{compl},K}(\widehat{\cM}_K,\cM^*)$
converges to zero in probability as $K\to\infty$.
\end{theorem}

\begin{proof}
For a fixed model $\cM\in\cH$, define its conditional error in
episode $r$ as
\[
R_r(\cM)
\coloneqq
\mathbb{E}\!\left[
    \frac{1}{n}\sum_{t=1}^n
    \ell_{\cM}(\tau_{r,t})
    \,\middle|\,
    \cD_{1:r-1}
\right].
\]
By the definition of $d_{\mathrm{compl},K}$,
\[
d_{\mathrm{compl},K}(\cM,\cM^*)
=
\frac{1}{K}\sum_{r=1}^K R_r(\cM).
\]

Fix $\cM\in\cH$ such that
$d_{\mathrm{compl},K}(\cM,\cM^*)>\epsilon$. Then
\[
\sum_{r=1}^K R_r(\cM)>K\epsilon.
\]

We apply Lemma~\ref{lem:episode-generalization} conditionally within
each episode. Conditioning on the history $\cD_{1:r-1}$ fixes the
data-collection policy used in episode $r$, while the transitions
within that episode remain a possibly nonstationary $\beta$-mixing
process. Let $m=\lfloor n/k\rfloor$ and
\[
q\coloneqq (m-1)\beta(k).
\]
Following the construction in Lemma~\ref{lem:episode-generalization},
for episode $r$ we obtain $m$ $k$-spaced samples and independent
proxy variables with the same respective marginals. Let $B_r$ denote
the event that the coupling between the dependent samples and their
independent proxies fails. Then
\[
\Pr(B_r\mid \cD_{1:r-1})\le q.
\]
Therefore, by the union bound,
\[
\Pr\!\left(\bigcup_{r=1}^K B_r\right)
\le Kq
=
K(m-1)\beta(k).
\]

Now consider the complementary event in which none of these couplings
fails. Let $C_r$ denote the event that all $m$ independent proxy
samples selected from episode $r$ incur zero error under $\cM$.
The argument of Lemma~\ref{lem:episode-generalization} gives
\[
\Pr(C_r\mid \cD_{1:r-1})
\le
e^{-mR_r(\cM)}.
\]
This bound remains valid although the policy changes between episodes,
because it is conditioned on the complete history preceding episode
$r$.

We next combine these bounds across episodes. Let the conditioning
history include all data and proxy variables generated before the
current episode. Since $R_r(\cM)$ is determined by the history before
episode $r$, repeated conditional expectation gives
\[
\begin{aligned}
&\mathbb{E}\!\left[
    \mathbf{1}\!\left[C_1\cap\cdots\cap C_K\right]
    \exp\!\left(
        m\sum_{r=1}^K R_r(\cM)
    \right)
\right]
\\
&\qquad\le 1.
\end{aligned}
\]
Indeed, for each episode $r$,
\[
e^{mR_r(\cM)}
\Pr(C_r\mid \cD_{1:r-1})
\le
e^{mR_r(\cM)}e^{-mR_r(\cM)}
=
1,
\]
so applying this inequality successively from episode $K$ back to
episode $1$ yields the result above.

If $C_1,\ldots,C_K$ all occur and
$d_{\mathrm{compl},K}(\cM,\cM^*)>\epsilon$, then
\[
m\sum_{r=1}^K R_r(\cM)
>
Km\epsilon.
\]
Hence,
\[
\begin{aligned}
&\Pr\!\left(
    C_1\cap\cdots\cap C_K
    \,\wedge\,
    d_{\mathrm{compl},K}(\cM,\cM^*)>\epsilon
\right)
\\
&\qquad\le
e^{-Km\epsilon}.
\end{aligned}
\]

If $\cM$ has zero empirical error on all observed transitions, then it
has zero error on every selected transition in every episode.
Therefore, whenever none of the coupling events $B_r$ occurs, zero
empirical error implies $C_1\cap\cdots\cap C_K$. Combining the two
failure events above gives, for any fixed $\cM\in\cH$,
\[
\Pr\!\left[
    d_{\mathrm{compl},K}(\cM,\cM^*)>\epsilon
    \;\wedge\;
    \widehat{\mathrm{Err}}(\cM)=0
\right]
\le
e^{-Km\epsilon}
+
K(m-1)\beta(k),
\]
where $\widehat{\mathrm{Err}}(\cM)$ denotes the empirical error over
all observed transitions.

Applying a union bound over the finite hypothesis class $\cH$,
\[
\begin{aligned}
&\Pr\!\left[
    \exists\,\cM\in\cH:
    d_{\mathrm{compl},K}(\cM,\cM^*)>\epsilon
    \;\wedge\;
    \widehat{\mathrm{Err}}(\cM)=0
\right]
\\
&\qquad\le
|\cH|
\left(
    e^{-Km\epsilon}
    +
    K(m-1)\beta(k)
\right).
\end{aligned}
\]

By Theorem~\ref{thm:soundness},
$\widehat{\cM}_K$ has zero empirical error on all transitions from
the $K$ observed episodes. Thus,
\[
\Pr\!\left[
    d_{\mathrm{compl},K}
    (\widehat{\cM}_K,\cM^*)>\epsilon
\right]
\le
|\cH|
\left(
    e^{-Km\epsilon}
    +
    K(m-1)\beta(k)
\right).
\]
Taking the complement proves that, with probability at least
\[
1-
|\cH|
\left(
    e^{-Km\epsilon}
    +
    K(m-1)\beta(k)
\right),
\]
we have
\[
d_{\mathrm{compl},K}
(\widehat{\cM}_K,\cM^*)
\le\epsilon.
\]

Finally, to establish convergence in probability, choose $k=n$, so
that $m=1$. The mixing penalty then vanishes:
\[
K(m-1)\beta(k)=0.
\]
Therefore, for every fixed $\epsilon>0$,
\[
\Pr\!\left[
    d_{\mathrm{compl},K}
    (\widehat{\cM}_K,\cM^*)>\epsilon
\right]
\le
|\cH|e^{-K\epsilon}
\longrightarrow 0
\qquad\text{as }K\to\infty.
\]
Hence
$d_{\mathrm{compl},K}(\widehat{\cM}_K,\cM^*)$
converges to zero in probability.
\end{proof}

The \(\beta\)-mixing term captures temporal dependence among transitions within an episode. In \sys{}, the data-collection policy is updated only between episodes; conditioning on the preceding collection history therefore allows Lemma~\ref{lem:episode-generalization} to be applied separately within each episode, while still permitting adaptive exploration across episodes. The bound reflects the usual tradeoff between the effective sample size \(m=\lfloor n/k\rfloor\) and the mixing penalty \((m-1)\beta(k)\). In particular, choosing \(k=n\) yields \(m=1\), eliminating the within-episode mixing penalty and giving convergence as the number of collected episodes increases.

\subsection{Illustration of Predicate Invention in Low-level Space}
\label{sec:pred_invent-lowlevel}

In this section, we illustrate how {\sys} correctly identifies circumstances where new predicates need to be invented. We here discuss cases for the precondition condition (Condition~\ref{eq:precond-invention}), and the effect condition follows the same logic. 
\input{sections/appendix/venn_diagram}

\subsection{Operator Learning}
\label{app:operator_learning_proof}
\input{sections/appendix/proof_operator_learning}

%% file: sections/appendix/venn_diagram.tex
\noindent
\begin{minipage}[t!]{0.45\textwidth}
  \centering
    \begin{tikzpicture}
      \draw[thick] (0,0) rectangle (6,4);
      \node at (6.3,4) {$\mathbf{U}$};
    
        \fill[blue!30,opacity=0.5] (3,2) ellipse (1.5 and 1.2);
      \draw[thick] (3,2) ellipse (1.5 and 1.2);
      \node at (4.5,3) {$\PreSuccessSet$};
    
      \filldraw[black] (3.4,2.2) circle (2pt);
      \node[below right] at (3.4,2.2) {$s_1$};
    
      \filldraw[black] (1,0.8) circle (2pt);
      \node[below left] at (1,0.8) {$s_2$};
    \end{tikzpicture}
\end{minipage}%
\hfill
\begin{minipage}[t!]{0.55\textwidth}
  Suppose that there exist two transitions: \\
  $\langle \state_1, \Ground{\skill}_1, s_1' \rangle, \langle \state_2, \Ground{\skill}_2, \state_2' \rangle \text{ such that } \state_1 \in \PreSuccessSet, \state_2 \notin \PreSuccessSet$ initially. The learned model $\Model$ so far is consistent with the two transitions.
\end{minipage}
\\

There are 3 possible circumstances of the estimated pre-success set $\EstInitiationSet$ resulting from $\Model$ learned from $\langle \state_1, \Ground{\skill}_1, \state_1' \rangle$ and $\langle \state_2, \Ground{\skill}_2, \state_2' \rangle$. For each of them, we discuss all possible cases of other transitions with their initial state falling into different subsets of the state space (marked in red).

\textbf{(1).} The learned model has $\PreSuccessSet \subset \EstInitiationSet$. Then, for each subset that initial states of other transitions might fall in:

\noindent
\begin{minipage}[t!]{0.45\textwidth}
  \centering
    \begin{tikzpicture}
      \begin{scope}
        \fill[pattern=north east lines, pattern color=black, line width=0.5mm] 
          (3,2) ellipse (2.2 and 1.76);
      \end{scope}
    
      \fill[blue!30,opacity=0.5] (3,2) ellipse (1.5 and 1.2);
      \draw[thick] (3,2) ellipse (1.5 and 1.2);
      \node at (4.5,3) {$\PreSuccessSet$};
    
      \draw[thick] (3,2) ellipse (2.2 and 1.76);
    
      \draw[thick] (0,0) rectangle (6,4);
      \node at (6.3,4.2) {$\mathbf{U}$};
    
      \filldraw[black] (3.4,2.2) circle (2pt);
      \node[below right] at (3.4,2.2) {$s_1$};
    
      \filldraw[black] (1,0.8) circle (2pt);
      \node[below left] at (1,0.8) {$s_2$};
    
      \filldraw[red!80!black] (0.8,3) circle (2pt);
      \node[below left] at (0.8,3) {\textcolor{red!80!black}{$a$}};
    
      \filldraw[red!80!black] (1.2,2) circle (2pt);
      \node[below left] at (1.2, 2) {\textcolor{red!80!black}{$b$}};
      
      \filldraw[red!80!black] (2.5 ,1.5) circle (2pt);
      \node[below right] at (2.5, 1.5) {\textcolor{red!80!black}{$c$}};
    
    \end{tikzpicture}
\end{minipage}%
\hfill
\begin{minipage}[t!]{0.55\textwidth}
  \begin{itemize}
      \item \textbf{(a)}: $a \notin \EstInitiationSet, a \notin \PreSuccessSet$. \\$\nexists\: \langle \state, \Ground{\skill}, \state'\rangle \in\cD \text{ s.t. } \state \in \EstInitiationSet \text{ while } s \notin \PreSuccessSet$. No need to invent new predicates.
      
      \item \textbf{(b)}: $b \in \EstInitiationSet, b \notin \PreSuccessSet$.\\ $b \in \EstInitiationSet \,, s_1 \in \EstInitiationSet, \text{ while } b \notin \PreSuccessSet \text{ and } s_1 \in \PreSuccessSet$. Condition~\ref{eq:precond-invention} is satisfied, and a new predicate will be invented.
      
      \item \textbf{(c)}: $c \in \EstInitiationSet, c \in \PreSuccessSet$. \\
      $\nexists\: \langle 
      \state, \Ground{\skill}, \state'\rangle \in \cD \text{ s.t. } s \in \EstInitiationSet \text{ while } \state \notin \PreSuccessSet$. No need to invent new predicates.
      
  \end{itemize}
\end{minipage}
\\

\textbf{(2).} The learned model has $\EstInitiationSet \subset \PreSuccessSet$. Then, for each section that initial states of other transitions might fall in:

\noindent
\begin{minipage}[t!]{0.45\textwidth}
  \centering
  
    \begin{tikzpicture}
      \begin{scope}
        \fill[pattern=north east lines, pattern color=black] 
          (3,2) ellipse (0.85 and 0.7);
      \end{scope}
    
      \fill[blue!30,opacity=0.5] (3,2) ellipse (1.5 and 1.2);
      \draw[thick] (3,2) ellipse (1.5 and 1.2);
      \node at (4.5,3) {$\PreSuccessSet$};
    
      \draw[thick] (3,2) ellipse (0.85 and 0.7);
    
      \draw[thick] (0,0) rectangle (6,4);
      \node at (6.3,4.2) {$\mathbf{U}$};
    
      \filldraw[black] (3.4,2.2) circle (2pt);
      \node[below right] at (3.4,2.2) {$s_1$};
    
      \filldraw[black] (1,0.8) circle (2pt);
      \node[below left] at (1,0.8) {$s_2$};
    
      \filldraw[red!80!black] (0.8,3) circle (2pt);
      \node[below left] at (0.8,3) {\textcolor{red!80!black}{$a$}};
    
      \filldraw[red!80!black] (2,1.6) circle (2pt);
      \node[below right] at (2,1.6) {\textcolor{red!80!black}{$b$}};
    
      \filldraw[red!80!black] (2.6,2) circle (2pt);
      \node[below right] at (2.6,2) {\textcolor{red!80!black}{$c$}};
    
    \end{tikzpicture}
\end{minipage}%
\hfill
\begin{minipage}[t!]{0.55\textwidth}
  \begin{itemize}
  
      \item \textbf{(a)}: $a \notin \EstInitiationSet, a \notin \PreSuccessSet$. \\$\nexists\: \langle 
      \state, \Ground{\skill}, \state'\rangle \in \cD \text{ s.t. } s \in \EstInitiationSet \text{ while } \state \notin \PreSuccessSet$. No need to invent new predicates.
      
      \item \textbf{(b)}: $b \notin \EstInitiationSet, b \in \PreSuccessSet$.\\ This indicates the new transition is inconsistent with the learned model ($ b \models \PreSuccessSet$ but $b \not \models \EstInitiationSet$). No new predicates need to be invented, and updating $\Model$ by re-calculating the operators will ensure that every transition is supported. Formally, $\exists \, \Ground{\operator}_\skill \in  \Ground{\Operators}_\skill, \text{ such that } s \in \EstInitiationSet, \forall \langle s, \omega, s' \rangle \in \mathcal{D}_\omega$. Then it is reduced to (a).
      
      
      \item \textbf{(c)}: $c \in \EstInitiationSet, c \in \PreSuccessSet$. \\
      $\nexists\: \langle 
      \state, \Ground{\skill}, \state'\rangle \in \cD \text{ s.t. } s \in \EstInitiationSet \text{ while } \state \notin \PreSuccessSet$. No need to invent new predicates.
      
  \end{itemize}
\end{minipage}
\\

\textbf{(3).} The learned model has $\EstInitiationSet \cap \PreSuccessSet \neq \emptyset, \, \EstInitiationSet \nsubseteq \PreSuccessSet, \, \PreSuccessSet \nsubseteq \EstInitiationSet$. Then, for each section that initial states of other transitions can fall in:

\noindent
\begin{minipage}[t!]{0.45\textwidth}
  \centering
  
    \begin{tikzpicture}
      \begin{scope}
        \clip (3,0) rectangle (6,4); 
        \fill[pattern=north east lines, pattern color=black, line width=0.5mm, distance=1mm] (0,0) rectangle (6,4);
      \end{scope}
    
      \fill[blue!30,opacity=0.5] (3,2) ellipse (1.5 and 1.2);
      \draw[thick] (3,2) ellipse (1.5 and 1.2);
      \node at (4.5,3) {$\PreSuccessSet$};
    
      \draw[thick] (0,0) rectangle (6,4);
      \node at (6.3,4.2) {$\mathbf{U}$};
    
      \draw[dashed, thick] (3,0) -- (3,4);
    
      \filldraw[black] (3.4,2.2) circle (2pt);
      \node[below right] at (3.4,2.2) {$s_1$};
    
      \filldraw[black] (1,0.8) circle (2pt);
      \node[below left] at (1,0.8) {$s_2$};
    
      \filldraw[red!80!black] (0.8,3) circle (2pt);
      \node[below left] at (0.8,3) {\textcolor{red!80!black}{$a$}};
    
      \filldraw[red!80!black] (2,1.6) circle (2pt);
      \node[below right] at (2,1.6) {\textcolor{red!80!black}{$b$}};
    
      \filldraw[red!80!black] (3.5,1.5) circle (2pt);
      \node[below right] at (3.5,1.5) {\textcolor{red!80!black}{$c$}};
    
      \filldraw[red!80!black] (5.2,2) circle (2pt);
      \node[below right] at (5.2,2) {\textcolor{red!80!black}{$d$}};
    
    \end{tikzpicture}
    
\end{minipage}%
\hfill
\begin{minipage}[t!]{0.55\textwidth}
  \begin{itemize}
  
      \item \textbf{(a)}: $a \notin \EstInitiationSet, a \notin \PreSuccessSet$. \\$\nexists\: \langle 
      \state, \Ground{\skill}, \state'\rangle \in \cD \text{ s.t. } s \in \EstInitiationSet \text{ while } \state \notin \PreSuccessSet$. No need to invent new predicates.
      
      \item \textbf{(b)}: 
      $b \notin \EstInitiationSet, b \in \PreSuccessSet$. \\
      Same situation as (2).b. No new predicates need to be invented, and updating $\Model$ by re-calculating the operators will ensure that every transition is supported.
      
      
      \item \textbf{(c)}: $c \in \EstInitiationSet, c \in \PreSuccessSet$. \\
      $\nexists\: \langle 
      \state, \Ground{\skill}, \state'\rangle \in \cD \text{ s.t. } s \in \EstInitiationSet \text{ while } \state \notin \PreSuccessSet$. No need to invent new predicates.
      
      \item \textbf{(d)}: $d \in \EstInitiationSet, d \notin \PreSuccessSet$.\\ $d \in \EstInitiationSet \,, s_1 \in \EstInitiationSet, \text{ while } d \notin \PreSuccessSet \text{ and } s_1 \in \PreSuccessSet$. Condition~\ref{eq:precond-invention} is satisfied, and a new predicate will be invented.
      
  \end{itemize}
\end{minipage}
\\

So far, we have discussed all possible cases of different transitions and $\EstInitiationSet$ resulting from the current model, and the predicate invention condition is proved to handle all cases.

%% file: sections/appendix/proof_operator_learning.tex
This section provides additional details on the operator learning algorithm. Specifically, it explains: \textbf{(Q1)} why Condition \ref{eq:precond-invention} uses learned operators as a measure for predicate invention without worrying about the operator learning algorithm being wrong, and \textbf{(Q2)} why it is impossible to have a transition predicted to be failed but actually successful when learning from a fixed dataset $\cD$.

In this section, $\cD$ denotes abstract transitions instead of low-level ones in the main paper, and $\cD^{suc}$ and $\cD^{fail}$ denote the subsets of successful and failed transitions in the dataset, respectively. 
Assuming preconditions are all conjunctive:

\begin{proposition}

        Let $\Operators_\skill$ be abstract actions learned by \sys{} from a subset of transitions $\cD_\skill \subseteq \cD$. 
        For every successful transition, there exists an operator whose preconditions are satisfied by the initial state:
        Formally, 
        $$\big( \langle \abstractState_i, \Ground{\operator}, \abstractNextState_i \rangle \in \cD^{suc}_\skill \big) \implies \big(\exists\, \Ground{\operator} \in \Ground{\Operators}_\skill, \abstractState_i\models \PRE[\Ground{\operator}] \big).$$
\end{proposition}

\textit{Proof}.
    %
    Suppose each abstract action $\operator \in \Operators_\skill$ is learned from its corresponding subset of dataset $\bm{d}_\skill \subseteq \cD_\skill$ sharing the same lifted abstract effects. \sys{} calculates its preconditions as:
    \[
    \PRE = \bigcap_{\langle \abstractState_i, \Ground{\operator}, \abstractNextState_i \rangle \in \cD^{suc}_\skill}\LiftF(\abstractState_i)
    \]
    By the definition of set intersection, the initial state of any successful transition must satisfy this intersection. Therefore, there must exist some $\Ground{\operator} \in \Ground{\Operators}_\skill$ such that $\abstractState_i \models \PRE[\Ground{\operator}]$. \qed
    
\begin{proposition}
        Let $\Operators_\skill$ be abstract actions learned by \sys{} from a subset of transitions $\cD_\skill \subseteq \cD$. If none of the operators in $\Operators_\skill$ has its preconditions satisfied by the initial state of a transition, the transition must fail. Formally, $$ \big(\forall \Ground{\operator} \in \Ground{\Operators}_\skill, \state_i \not\models\PRE[\Ground{\operator}] \big) \implies \big(\langle \abstractState_i, \Ground{\operator}, \abstractNextState_i \rangle \in \cD_\skill^{fail}\big).$$
\end{proposition}

\textit{Proof}. 
%
Suppose, for each operator $\operator \in \Operators_\skill$ learned from its corresponding subset of dataset $\bm{d}_\skill \subseteq \cD_\skill$ with the same abstract effects. 
For any successful initial state, $\LiftF{(\abstractState)} \models \big(\bigcap_{\langle \abstractState_i, \Ground{\operator}, \abstractNextState_i \rangle \in \cD^{suc}_\skill} \LiftF({\abstractState}_i) \big) = \PRE$, because taking intersection only relax the precondition.  Hence, for any state $\abstractState \not \in \PRE$, it would mean that its corresponding transition is not included in $\cD^{suc}_{\skill}$. Given that any transition in $\cD_\skill$ must be either successful or failed, $\langle \abstractState_i, \Ground{\operator}, \abstractNextState_i \rangle \in \cD_\skill^{fail}.$ \qed


\textbf{(A1)} With the above properties established, we have narrowed down the only unproven case to false negatives—where $\langle \abstractState_i, \Ground{\operator}, \abstractNextState_i \rangle \in \cD_\skill^{fail}$ but $\exists \, \Ground{\operator} \in \Ground{\Operators}, \abstractState_i \models \PRE[\Ground{\operator}]$. In such a case, further refinement is infeasible with the current abstract state representations unchanged, and predicate invention must be triggered to distinguish these false negatives.


\textbf{(A2)} The scenario where a transition's initial state fails to satisfy the preconditions of any $\operator \in \Operators$ yet still executes successfully is impossible under a fixed $\cD$. Such a transition would necessarily belong to $\cD^{suc}$, a contradiction directly ruled out by Proposition 2.

%% file: sections/appendix/upper_bound.tex
Recall that our learned symbolic model has the form
\[
\mathcal{M} = (\mathcal{P}, \mathcal{A}),
\]
where $\mathcal{P}$ is a set of predicates 
and
$\mathcal{A} = \bigcup_{\omega \in \Omega} \mathcal{A}_\omega$ is a set of abstract operators, with $\mathcal{A}_\omega$ the set of operators associated with skill $\omega \in \Omega$.

Each operator $a \in \mathcal{A}_\omega$ is defined by its preconditions and add/delete effects:
\[
a \triangleq \bigl(\omega, \boldsymbol{\theta}_a, \textsc{Pre}_a,\textsc{Eff}_a^+,\textsc{Eff}_a^-\bigr),
\quad
\textsc{Pre}_a,\textsc{Eff}_a^+,\textsc{Eff}_a^- \subseteq \mathcal{P}.
\]

In our implementation, the number of VLM calls is finite, and the resulting models use a small number of predicates in environments with finitely many objects.
For the \emph{theoretical analysis}, we make this implicit resource bound explicit:

\begin{itemize}
    \item Let $P_{\max}$ denote a fixed maximum number of predicates that \sys{} is allowed to invent.
    \item Let $\mathcal{P}_{\mathrm{FM}}$ denote the finite family of
predicate classifiers expressible by the FM.
    \item Let $\mu_{\max}$ denote the maximum arity of any predicate $\sigma \in \mathcal{P}$.

\end{itemize}
Because \sys{} learns one operator per lifted effect cluster, we can derive an upper bound on the number of operators per skill $\omega \in \Omega$ based on the number of possible effect sets. For any operator $a \in \gA_\omega$ and predicate $\sigma \in \gP$, there are three possible cases: $\sigma \in \textsc{Eff}_a^+$, $\sigma \in \textsc{Eff}_a^-$, or $\sigma \notin \textsc{Eff}_a^-\cup\textsc{Eff}_a^+$.
%
Because the upper bound of possible instances of $p$ is $|\mathcal{O}|^{\mu_{\max}}$, we can express the maximum number of operators as

\[
A_{\max} = 3^{P_{\max} \cdot |\mathcal{O}|^{\mu_{\max}}}
\]

We therefore define the hypothesis class analyzed in Theorem~\ref{thm:prob-completeness} as
\begin{equation}
\label{eq:hypothesis_class_def}
\cH =
\Bigl\{
(\cP,\cA)
\;|\;
\cP\subseteq\mathcal P_{\mathrm{FM}},
\ |\cP|\le P_{\max},
\ |\cA_\omega|\le A_{\max}\, \forall\omega\in\Omega
\Bigr\}.
\end{equation}

\subsection{Upper Bound on \texorpdfstring{$|\mathcal{H}|$}{|H|}}

We now derive a practical upper bound on the size of $\mathcal{H}$ in~\eqref{eq:hypothesis_class_def}.

Let $\mathcal{P}_{\mathrm{FM}}$ denote the finite family of predicate classifiers expressible by the FM.
Since a learned model contains at most $P_{\max}$ predicates, the number of possible predicate sets $\mathcal{P} \subseteq \mathcal{P}_{\mathrm{FM}}$ is bounded by
\[
\sum_{j=0}^{P_{\max}}
\binom{|\mathcal P_{\mathrm{FM}}|}{j}
\;\le\;
\left(|\mathcal P_{\mathrm{FM}}|+1\right)^{P_{\max}}.
\]

Fix one such predicate set $\mathcal{P}$ with $|\mathcal{P}| \le P_{\max}$. For each operator $\alpha$, its symbolic definition is given by three subsets of $\mathcal{P}$:
\[
\textsc{Pre}_a,\ \textsc{Eff}_a^+,\ \textsc{Eff}_a^- \subseteq \mathcal{P}.
\]
We consider negative preconditions, so there are three possibilities for one predicate $p$: $p \in \textsc{Pre}_a$, $-p \in \textsc{Pre}_a$, and $p \notin \textsc{Pre}_a$. Since $\textsc{Eff}_a^+$ and $\textsc{Eff}_a^-$ are considered in $A_{\max}$, a single operator has at most
\begin{equation}
\label{eq:operator_count}
3^{P_{\max} \cdot |\mathcal{O}|^{\mu_{\max}}}
\end{equation}
distinct configurations of preconditions.

For a fixed skill $\omega \in \Omega$, we allow at most $A_{\max}$ operators. Treating each of the $A_{\max}$ operator ``slots'' as independently choosing one of the $3^{P_{\max} |\mathcal{O}|^{\mu_{\max}}}$ possible configurations in~\eqref{eq:operator_count}, the total number of operator-sets $\mathcal{A}_\omega$ for that skill is bounded by
\begin{equation}
\label{eq:skill_operator_sets}
\bigl(3^{P_{\max} \cdot |\mathcal{O}|^{\mu_{\max}}}\bigr)^{A_{\max}}
\;=\;
3^{P_{\max} A_{\max} |\mathcal{O}|^{\mu_{\max}} }.
\end{equation}
Across all skills $\omega \in \Omega$ and all possible predicate sets, we obtain the bound
\begin{equation}
\label{eq:h_size_bound}
|\mathcal{H}|
\;\le\;
\left(|\mathcal P_{\mathrm{FM}}|+1\right)^{P_{\max}}
\prod_{\omega \in \Omega}
3^{P_{\max} A_{\max} |\mathcal{O}|^{\mu_{\max}} }
\;=\;
\left(|\mathcal P_{\mathrm{FM}}|+1\right)^{P_{\max}}
3^{P_{\max} A_{\max} |\Omega| |\mathcal{O}|^{\mu_{\max}}}.
\end{equation}


\subsection{Sample Complexity for a Target \texorpdfstring{$(\epsilon,\delta)$}{(ε,δ)}}

Theorem~\ref{thm:prob-completeness} states that, for any $\epsilon>0$,
\begin{equation}
\label{eq:pac_bound_main_text}
\Pr[d_{\mathrm{compl},K}(\widehat{\cM}_K,\cM^*)>\epsilon]
\le
|\mathcal{H}|\left(e^{-Km\epsilon}+K(m-1)\beta(k)\right),
\end{equation}
where $m = \left\lfloor n/k \right\rfloor$ is the effective sample size, $\widehat{\mathcal{M}}_K$ is the model returned by \sys{} after observing $K$ episodes of \(n\) transitions each, and $d_{\mathrm{compl}}$ is the distance defined in the main text (probability of a ``missed feasible'' event under the transition distribution).

Substituting the bound on $|\mathcal{H}|$ from~\eqref{eq:h_size_bound} into~\eqref{eq:pac_bound_main_text} yields
\begin{equation}
\begin{aligned}
\Pr\Bigl[
d_{\mathrm{compl},K}\bigl(\widehat{\mathcal{M}}_K,\mathcal{M}^*\bigr)
> \epsilon
\Bigr]
\;\le\;&
\left(|\mathcal{P}_{\mathrm{FM}}|+1\right)^{P_{\max}}
3^{P_{\max} A_{\max} |\Omega| |\mathcal{O}|^{\mu_{\max}}}
\\
&\cdot
\left(
e^{-Km\epsilon}
+
K(m-1)\beta(k)
\right).
\end{aligned}
\end{equation}
Let $C = \left(|\mathcal{P}_{\mathrm{FM}}|+1\right)^{P_{\max}}3^{P_{\max} A_{\max} |\Omega| |\mathcal{O}|^{\mu_{\max}}}$.
To guarantee that this probability is at most $\delta \in (0,1)$, it suffices that
\[
C \left( e^{-Km\epsilon} + K(m-1)\beta(k) \right)
\;\le\;
\delta.
\]

Instead of trying to solve for $n$ simultaneously across both terms, we split the total allowable failure probability ($\delta$) in half. To guarantee the total sum is less than $\delta$, it suffices to ensure that each individual piece is less than $\delta/{2}$:
\begin{gather}
C e^{-Km\epsilon} \le \frac{\delta}{2}, \label{eq:e_term}
\\
C K(m-1)\beta(k) \le \frac{\delta}{2} \label{eq:linear_term}
\end{gather}
Solve \eqref{eq:e_term}, we get
$$m \ge \frac{1}{K\epsilon} \left( \ln C + \ln 2 + \ln \frac{1}{\delta} \right)$$
Substitute back $C$:
\begin{equation}
    m \ge \frac{1}{K\epsilon} \left( P_{\max}\ln\!\left(|\mathcal P_{\mathrm{FM}}|+1\right)
+
P_{\max}A_{\max}|\Omega||\mathcal O|^{\mu_{\max}}\ln 3 + \ln 2 + \ln \frac{1}{\delta} \right)
\end{equation}

Now we look at the second split inequality in \eqref{eq:linear_term}:
$$\beta(k) \le \frac{\delta}{2CK(m-1)}$$
In $\beta$-mixing process, the mixing coefficient $\beta(k)$ typically decays exponentially with the time-gap parameter $k$ (i.e., $\beta(k) \le c \cdot e^{-bk}$ for physical constants $c, b > 0$). Apply this standard assumption, \eqref{eq:linear_term} becomes
$$c \cdot e^{-bk} \le \frac{\delta}{2CK(m-1)}.$$
Solve the equation, we have
$$k \ge \frac{1}{b}\left(\ln C + \ln K + \ln(m-1) + \ln \frac{1}{\delta} + \ln(2c)\right)$$

Using the error-splitting technique under standard exponential mixing ($\beta(k) \le c \cdot e^{-bk}$), we found:
\begin{equation}
m = \mathcal{O}\left( \frac{\ln C + \ln(1/\delta)}{K\epsilon} \right)
\end{equation}

\begin{equation}
   k = \mathcal{O}\left( \ln C + \ln K + \ln(1/\delta) + \ln(m) \right)
\end{equation}

To simplify the algebra, let $V = \ln C + \ln(1/\delta)$. This yields
$$m = \mathcal{O}\left(\frac{V}{K\epsilon}\right) \quad \text{and} \quad k = \mathcal{O}\left(V + \ln(Km)\right)$$

Since $Km = O({V}/{\epsilon})$,
$$k = \mathcal{O}\left( V + \ln\left(\frac{V}{\epsilon}\right) \right) = \mathcal{O}\left( V + \ln(V) + \ln\left(\frac{1}{\epsilon}\right) \right).$$

Because $V$ grows faster than its own logarithm ($\ln(V)$), the term $\ln(V)$ gets absorbed by $V$ term. This leaves a clean bound for $k$:
$$k = \mathcal{O}\left( V + \ln\left(\frac{1}{\epsilon}\right) \right) = \mathcal{O}\left( \ln C + \ln\left(\frac{1}{\delta}\right) + \ln\left(\frac{1}{\epsilon}\right) \right)$$

Since \(m=\lfloor n/k\rfloor\), per-episode sample complexity satisfies $mk\leq n < (m+1)k$. Multiplying the two complexities:
$$n = \mathcal{O}\left( \left[ \frac{\ln C + \ln(1/\delta)}{K\epsilon} \right] \cdot \left[ \ln C + \ln\left(\frac{1}{\delta}\right) + \ln\left(\frac{1}{\epsilon}\right) \right] \right),$$

and the total sample complexity:
$$N = Kn = \mathcal{O}\left( \left[ \frac{\ln C + \ln(1/\delta)}{\epsilon} \right] \cdot \left[ \ln C + \ln\left(\frac{1}{\delta}\right) + \ln\left(\frac{1}{\epsilon}\right) \right] \right).$$

The final total sample complexity of $N$ 
to ensure
\[
d_{\mathrm{compl,K}}\bigl(\widehat{\mathcal{M}}_K,\mathcal{M}^*\bigr) \le \epsilon
\quad\text{with probability at least } 1-\delta
\]
is expressed as

\begin{equation}
\label{eq:final_complexity}
\begin{aligned}
N = \mathcal{O}\Bigg(
\frac{1}{\epsilon}\Big[
&\left(
P_{\max}\ln\left(|\mathcal{P}_{\mathrm{FM}}|+1\right)
+
P_{\max}A_{\max}|\Omega||\mathcal{O}|^{\mu_{\max}}
+
\ln(1/\delta)
\right)^2
\\
&+
\left(
P_{\max}\ln\left(|\mathcal{P}_{\mathrm{FM}}|+1\right)
+
P_{\max}A_{\max}|\Omega||\mathcal{O}|^{\mu_{\max}}
+
\ln(1/\delta)
\right)
\ln(1/\epsilon)
\Big]
\Bigg).
\end{aligned}
\end{equation}

In our experiments, the arity and realized numbers of predicates are small. 
Setting $P_{\max}$ to match the practical budget yields numerical values in~\eqref{eq:final_complexity} for the regimes we study. Our theoretical result therefore formalizes how increasing the capacity of the symbolic model (via larger $P_{\max}$) trades off against the number of transitions needed to achieve a desired completeness level.

\clearpage

%% file: sections/appendix/comparison_table.tex

\begin{table}[H]
    \centering
    \footnotesize 
    \caption{Comparison of symbolic model learning methods.}
    \label{tab:method-comparison}
    \begin{tabularx}{\linewidth}{Xcccccc} 
        \toprule
        \textbf{Method} & 
        \makecell{\textbf{Fully Obs. \&} \\ \textbf{Deterministic}} & 
        \makecell{\textbf{No Initial} \\ \textbf{Pred.}} & 
        \makecell{\textbf{Theoretical} \\ \textbf{Analysis}} & 
        \makecell{\textbf{Auton.} \\ \textbf{Learning}} & 
        \makecell{\textbf{Interp.} \\ \textbf{Rep.}} &
        \makecell{\textbf{Real-World} \\ \textbf{Eval.}} \\
        \midrule
        \cite{liang2025visualpredicator} & \cmark & & & & \cmark & \\ \relax
        \cite{athalye2026pix2pred} & \cmark & & & & \cmark & \cmark \\ \relax
        \cite{liang2025exopredicatorlearningabstractmodels} & \cmark & & & & \cmark & \\ \relax
        \citet{silver2021learningsymbolicoperatorstask} & \cmark & & & & & \\ \relax
        \citet{silver2023predinvent}  & \cmark & & & & & \\ \relax
        \cite{han2024interpret} & \cmark & \cmark & & & \cmark & \cmark \\ \relax
        \cite{Konidaris2018FromST} & & \cmark & \cmark & & & \cmark \\ \relax
        \cite{shah2024reals} & \cmark & \cmark & & & & \cmark \\ 
        \midrule
        \textbf{\sys{} (Ours)}  & \cmark & \cmark & \cmark & \cmark & \cmark & \cmark \\ 
        \bottomrule
    \end{tabularx}
\end{table}

\paragraph{Fully Observable and Deterministic Assumption}
Prior and concurrent work in predicate invention for model learning and symbolic model learning commonly make two key assumptions, deterministic skills and a fully observable environment; \sys{} follows this paradigm. In the future, to relax these assumptions, modeling states as distributions instead of sets and introducing probabilistic representations---such as Probabilistic PDDL (PPDDL)---presents a promising direction, if they are grounded in a concrete theoretical framework. A prominent example of work that relaxes these assumptions is provided by \citet{Konidaris2018FromST}.

\paragraph{Predicate Learning from Scratch}
The problem of predicate invention can be categorized into ``completing partial operators" and ``learning from scratch". In the former setting~\citep{liang2025visualpredicator, liang2025exopredicatorlearningabstractmodels, athalye2026pix2pred, silver2023predinvent, silver2021learningsymbolicoperatorstask}, the system is initialized with predefined goal predicates that partially specify the operators. Because these methods do not invent predicates needed to model skill effects, the predefined goal predicates must already be assigned to the appropriate locations in the partial operators. Moreover, predefined tasks are provided at the same time to enable the collection of transition data. Together, these requirements impose substantial prior knowledge. In the latter setting, the system must begin with an empty predicate set. This ``learning from scratch" approach is significantly more challenging but better reflects realistic scenarios where untrained users, rather than planning experts, interact with the system. \sys{} adopts this latter paradigm, aligning with the work of \citet{Konidaris2018FromST, shah2024reals, han2024interpret}. 

\paragraph{Theoretical Analysis}
Theoretical rigor is essential for learning a task-level model with formal guarantees. While existing research has explored predicate invention through planning heuristics~\citep{silver2023predinvent, silver2021learningsymbolicoperatorstask} or derived guarantees for non-generative cases~\citep{Konidaris2018FromST}, we contend that current generative predicate invention approaches~\citep{liang2025visualpredicator,liang2025exopredicatorlearningabstractmodels, athalye2026pix2pred} do not characterize the conditions under which the learned abstractions support sound and complete planning. This limitation primarily stems from their reliance on greedy best-first search and predefined goal predicates. In contrast, \sys{} only invents predicates when necessary to reconcile the symbolic model with observed data, and validates that a proposed predicate improves the model before keeping it. Hence, our method does not require an ad-hoc greedy predicate selection procedure, enabling a formal analysis of when these properties hold, as shown in Appendix~\ref{app:proofs}.

\paragraph{Autonomous Learning}
As demonstrated in our experiments, efficiently collecting transition data within object-centric environments with parameterized skills remains a significant challenge. Current approaches typically rely on pre-collected datasets~\citep{Konidaris2018FromST, shah2024reals}, human guidance~\citep{han2024interpret}, or the specification of explicit training tasks~\citep{liang2025visualpredicator, liang2025exopredicatorlearningabstractmodels, athalye2026pix2pred, silver2021learningsymbolicoperatorstask, silver2023predinvent}. In contrast, achieving true scalability requires a shift toward autonomous learning. \sys{} introduces a novel paradigm for this purpose, utilizing goal-agnostic skill sequences for pure exploration. By decoupling data collection from specific tasks or human guidance, \sys{} offers a more flexible and promising path forward.

\paragraph{Interpretable Representation}
In previous works where abstractions are captured by in-domain classifiers~\citep{Konidaris2018FromST} or generated via fixed templates~\citep{silver2021learningsymbolicoperatorstask, silver2023predinvent}, specifying real-world tasks was challenging because the predicates and operators were unreadable by non-experts, and often required manual inspection of the states where classifiers were triggered or templates were satisfied. In contrast, recent research leveraging LLMs or VLMs for predicate invention produces abstractions that are directly human-interpretable~\citep{liang2025visualpredicator, liang2025exopredicatorlearningabstractmodels, athalye2026pix2pred}. Within this context, \sys{} is the first to characterize the essential properties required for such systems.

\paragraph{Real-World Evaluation}
Evaluation on physical robotic platforms is essential to demonstrate a system’s practical applicability. Traditional Task and Motion Planning (TAMP) typically assumes fully observable and deterministic environments, effectively abstracting away sensory noise within its theoretical frameworks. Therefore, real-world evaluation was not necessary since the environments are perfectly controlled, assuming full access to perfect state information, such as object poses. However, recent works leveraging foundation models without real-world evaluation~\citep{liang2025visualpredicator, liang2025exopredicatorlearningabstractmodels} have raised concerns regarding the noise and uncertainty inherent in probabilistic neural networks. To address these concerns, we highlight the reliability of \sys{} in real-world scenarios by providing a detailed evaluation of its performance in inventing and classifying predicates under the complexities of physical interaction (see Appendix~\ref{appendix:vlm_study}).

%% file: sections/appendix/operator_learning.tex
\subsection{Lifting and Substitution}

Object lifting removes object identities from an abstract state while preserving their types and relational structure, allowing predicate invention and operator learning to compare transitions involving different object instances. Given an abstract state $\abstractState$ and object tuple $\ObjectTuple$, let $\abstractState|_{\ObjectTuple}$ denote the restriction of $\abstractState$ to grounded predicates containing at least one object in $\ObjectTuple$. For a substitution $\theta = [\bm{v}/\bm{o}]$, let $\theta(f)$ denote the result of replacing each object $o_i$ in a grounded predicate formula $f$ with the corresponding typed variable $v_i$.

In \sys{}, the substitution associated with $\ObjectTuple$ is constructed deterministically from the skill arguments and object types. We therefore define the lifting function as
\[
\LiftF_{\ObjectTuple}(\abstractState)
\coloneqq
\theta_{\ObjectTuple}\left(
\abstractState|_{\ObjectTuple}
\right),
\]
where $\theta_{\ObjectTuple}$ is the deterministic substitution induced by $\ObjectTuple$. Thus, two low-level states $\state_i$ and $\state_j$, associated with object tuples $\ObjectTuple_i$ and $\ObjectTuple_j$, have the same lifted abstract representation when
\[
\LiftF_{\ObjectTuple_i}\!\left(\AbstractionF(\state_i)\right)
=
\LiftF_{\ObjectTuple_j}\!\left(\AbstractionF(\state_j)\right).
\]
This allows transitions involving different object identities to be clustered and compared in a common lifted representation.

\subsection{Running Example}
We provide a running example here to illustrate the operator learning algorithm. Suppose that the environment contains the following objects with their typing: $\mathtt{Robot - [robot]}$, $\mathtt{Table - [table]}$, $\mathtt{Bottle - [object, openable]}$, $\mathtt{Apple - [object]}$, and $\mathtt{Banana - [object]}$. The environment has these predicates:

\begin{itemize}
    \item $\mathtt{(is\_on \enspace ?object \enspace ?table)}$: The object is on the table.
    \item $\mathtt{(hand\_free \enspace ?robot)}$: The robot has a free hand or gripper.
    \item $\mathtt{(is\_holding \enspace ?robot \enspace ?object)}$: The robot is currently holding the object.
\end{itemize}

\vspace{0.5em}
We have collected three successful transitions for the skill $\mathtt{PickUp(robot, object)}$:

\begin{lstlisting}[language=Lisp,mathescape=true, deletekeywords={and, not, or,  first, second}]

Transition 1:
    Skill: PickUp(Robot, Bottle)
    Initial State: (and
        (is_on Bottle Table)
        (hand_free Robot)
        (not (is_holding Robot Bottle))
        (is_on Banana Table)
        )
        
    Goal State: (and
        (not (is_on Bottle Table))
        (not (hand_free Robot))
        (is_holding Robot Bottle)
        (is_on Banana Table)
        )
        
Transition 2:
    Skill: PickUp(Robot, Apple)
    Initial State: (and
        (is_on Apple Table)
        (hand_free Robot)
        (not (is_holding Robot Apple))
        (is_on Banana Table)
        )
        
    Goal State: (and
        (not (is_on Apple Table))
        (not (hand_free Robot))
        (is_holding Robot Apple)
        (is_on Banana Table)
        )

Transition 3:
    Skill: PickUp(Robot, Bottle)
    Initial State: (and
        (hand_free Robot)
        (not (is_holding Robot Bottle))
        (is_on Banana Table)
        )
        
    Goal State: (and
        (not (hand_free Robot))
        (is_holding Robot Bottle)
        (is_on Banana Table)
        )
\end{lstlisting}

We first lift all objects with respect to the skill instances and calculate the lifted effect, so that skills operating on different objects can be grouped into one cluster.

\begin{lstlisting}[language=Lisp,mathescape=true, deletekeywords={and, not, or,  first, second}]

Transition 1:
    Eff+: < (is_holding ?robot_1 ?object_1 ) >
    Eff-: < (is_on ?object_1 ?table_1 ),  (hand_free ?robot_1) >
    
Transition 2:
    Eff+: < (is_holding ?robot_1 ?object_1 ) >
    Eff-: < (is_on ?object_1 ?table_1 ),  (hand_free ?robot_1) >

Transition 3:
    Eff+: < (is_holding ?robot_1 ?object_1 ) >
    Eff-: < (hand_free ?robot_1) >
\end{lstlisting}

Then, transition 1 and 2 fall in the same cluster because they share the same effect, whereas transition 3 forms a cluster itself. From the two clusters, we can calculate their preconditions by taking the intersections of their initial states in the lifted form. 

\begin{lstlisting}[language=Lisp,mathescape=true, deletekeywords={and, not, or,  first, second}]

Cluster 1 (Transition 1, 2):
    Pre: (and
    (not (is_holding ?robot_1 ?object_1))
    (is_on ?object_1 ?table_1)
    (hand_free ?robot_1) 
    )

Cluster 2 (Transition 3):
    Pre: (and
    (not (is_holding ?robot_1 ?object_1))
    (hand_free ?robot_1) 
    )
    
\end{lstlisting}

Their action parameters are determined by taking the union of the parameters in their precondition and effects. Finally, we learn two operators for the skill $\mathtt{PickUp(robot, object)}$:

\begin{lstlisting}[language=Lisp,mathescape=true, deletekeywords={and, not, or,  first, second}]

Operator 1 (Transition 1,2):
(:action Pick_1
:parameters ( ?robot_1 - robot ?object_1 - object ?table_1 - table)
:precondition (and 
    (not (is_holding ?robot_1 ?object_1))
    (hand_free ?robot_1 )
    (is_on ?object_1 ?table_1)  
) 
:effect (and 
    (is_holding ?robot_1 ?object_1 )
    (not (is_on ?object_1 ?table_1 ))
    (not (hand_free ?robot_1))
 ) 
)

Operator 2 (Transition 3):
(:action Pick_2
:parameters ( ?robot_1 - robot ?openable_1 - openable )
:precondition (and 
    (not (is_holding ?robot_1 ?openable_1))
    (hand_free ?robot_1 )
) 
:effect (and 
    (is_holding ?robot_1 ?openable_1 )
    (not (hand_free ?robot_1))
 ) 
)
\end{lstlisting}

We highlight that the operator learning algorithm of \sys{} is general and practical with these important details:

\begin{itemize}
    \item \sys{}’s operator learning algorithm can capture conditional effects, provided that the relevant predicates are nullary or involve at least one skill parameter, such as $\mathtt{is\_on(?object, ?table)}$, in the relevant grounded abstract transitions.
    \item \sys{}'s operator learning algorithm handles multi-type objects by using the lowest level of the type hierarchy consistent with the data, preventing over-generalization while retaining compositional structure. In the example above, we learn the parameter of $\mathtt{PickUp\_2}$ to be $\mathtt{?openable}$ instead of $\mathtt{?object}$.
    \item \sys{}'s operator learning algorithm clusters transitions after lifting them, which enables efficient symbolic model learning. In the example above, we successfully learn from cluster 1, where transitions belong to different grounded skills. 
    \item \sys{}'s operator learning algorithm discards tautological predicates---those that are always true or always false. In the example, $\mathtt{(is\_on \enspace Banana \enspace Table)}$ is true for all states, and thus is removed at the initial step.
\end{itemize}

%% file: sections/appendix/learned_operators.tex
\subsection{Result Analysis of Baseline Comparisons}
We present an analysis of the failure modes observed in \sys{} and its baselines; note that the System Predicates baseline is evaluated exclusively within the Robotouille domain. One common failure mode is misclassifications during inference time, specifically when abstracting raw visual observations into symbolic states using the learned predicates. As our primary contributions are centered on theoretical and algorithmic design, we did not incorporate advanced prompting or visual-reasoning techniques—such as ViperGPT~\citep{surís2023vipergptvisualinferencepython}—which could potentially mitigate these perception-level errors.

\textbf{Expert Operators} function as oracle knowledge and represent the theoretical upper bound for learning-based systems, as evidenced by the results in the Robotouille \citep{gonzalez-pumariega2025robotouille} domain. However, this baseline’s performance degrades significantly in real-world experiments, where it is outperformed by \sys{} and other learning-based methods. A potential explanation for this shift is that foundation models may more effectively ground and interpret predicate semantics when those abstractions are self-generated rather than manually defined by human experts. The central advantage of our approach is that it reduces reliance on domain expertise, empowering robots to autonomously derive symbolic representations that are inherently tailored to their specific embodiment, physical capabilities, and task distributions.

\textbf{System Predicates} utilizes a fixed set of eight predicates designed to cover all possible states within the domain. However, this compact representation proves insufficient for capturing essential environmental dynamics.  Specifically, the absence of a predicate $\mathtt{clear\_above}$ (or its semantic equivalent) prevents the model from representing the critical precondition of the \textit{Pick} skill: that an object must be unobstructed from the top to be successfully grasped. Because predicate invention is disabled in this baseline, the planner frequently generates unsound plans—such as attempting to retrieve an object from the bottom of a stack—resulting in a significantly lower success rate for high-complexity or impossible tasks.

\textbf{ViLa}'s failure modes include neglecting skill parameter constraints, repeating failed skills within its action history, and failing to recognize goal state has been reached. Although \sys{} still outperformed this baseline, ViLa exhibits stronger-than-expected results in the Bi-manual Kuka domain, likely due to similarities with its training distribution. A critical finding is its consistently poor performance on \textit{Impossible} tasks, which underscores the risks of unsoundness and the reliability concerns typical of purely generative planners.

\textbf{Random Exploration} is severely constrained by the inherent inefficiency of stochastic data collection. Despite this, the predicates and operators it manages to acquire are of high quality, as this baseline utilizes the same underlying invention algorithm as \sys{}. The primary failure mode stems from the low probability of randomly executing skills with complex preconditions. This hinders the invention process, because predicate invention can only happen when the skills are successfully executed. For example, in the Robotouille setting, the \textit{Cook} action requires the item to be on the stove and the robot's hand to be empty. Consequently, it yields near-zero success rates across most domains, solving only the simplest tasks where basic skills like \textit{Pick} suffice. Furthermore, its good performance on \textit{Impossible} tasks is a byproduct of model incompleteness: because it lacks the operators necessary to chain complex actions together, the planner correctly—albeit for the wrong reasons—identifies those tasks as unsolvable.

\textbf{FM Invent} tends to continuously propose new predicates regardless of whether the existing set is sufficiently expressive to model the domain. While it successfully gathers transition data for all skills by leveraging the active collection algorithm of \sys{}, the over-invented predicates impose a significant classification burden on the foundation model during inference. Consequently, the learned operators become over-specified, containing numerous redundant preconditions and effects that hinder effective planning. This explains the baseline’s poor success rate alongside its high performance on \textit{Impossible} tasks; the resulting model is so overly constrained by superfluous symbolic logic that it essentially views most valid transitions as infeasible.

\textbf{No Heuristic} exhibits a failure mode similar to System Predicates, e.g., performing well on \textit{Easy} problems but degrading significantly on \textit{Hard} tasks in Robotouille domain. Our investigation reveals that this baseline generally invents fewer predicates than \sys{}, leading to the omission of critical symbolic logic in two out of five experimental runs. This explains the high variance observed in \textit{Hard} and \textit{Impossible} scenarios and suggests that \textit{Easy} tasks can often be solved even with an incomplete predicate set—a finding that mirrors our observations of System Predicates in the Robotouille domain. Taken together with the Franka domain results, these findings validate the utility of our two engineered heuristics for skill sequence proposal. Interestingly, performance in the bi-manual Kuka experiments remained relatively stable; we attribute this to transition "dead-ends" that naturally constrain the exploration space, combined with preconditions that are simpler to model than those in other domains. Nevertheless, enhancing exploration strategies remains a promising avenue for future research. Finally, we note that while metrics such as success rate and planning budget effectively evaluate learned operators, there is a distinct need for novel metrics specifically designed to assess exploration efficiency.

{\scshape\bfseries SkillWrapper}'s failure mode is similar to the case of Random Exploration. In the Robotouille domain, for instance, a comparison between the PDDL operators learned by \sys{} and those of the Expert baseline reveals that the inclusion of just one extra predicate, $\mathtt{on\_cutting\_board(item)}$, partitions the transition data into excessively small clusters. Consequently, the transition data in these smaller clusters cannot support the model learning algorithm to effectively filter out spurious preconditions. This case study underscores the critical equilibrium required between predicate invention and data acquisition: if a system over-invents predicates relative to available transition data, the resulting operators become over-specified and fail to generalize. In \sys{}, this balance is empirically regulated via the scoring function threshold and the length of the proposed skill sequences. A remaining sub-optimality is that our current framework does not filter for semantic synonyms or antonyms, which increases the classification burden on the foundation model. While these redundancies are typically managed by the model's internal reasoning, a more sophisticated prompting system could further mitigate this issue and enhance computational efficiency.

\subsection{Learned Predicates \& Operators}
~\\

\begin{lstlisting}[language=Lisp,mathescape=true, deletekeywords={and, not, or,  first, second}]
# Learned Predicates & Operators of Burger domain

# Predicates
  name: hand_empty
  types:
  - robot
  semantic: the robot isn't holding any pickupable object.

  name: station_surface_empty
  types:
  - station
  semantic: no pickupable object is resting on the station's top surface.
  
  name: chopped
  types:
  - cuttable
  semantic: the cuttable object is visually in multiple small, separated pieces rather than a single intact piece.

  name: holding
  types:
  - robot
  semantic: the robot's gripper is visibly grasping and carrying the specified pickupable object.

  name: cooked
  types:
  - cookable
  semantic: the specified cookable exhibits a cooked appearance (browned surface or grill marks, not raw/pink).

  name: on_station
  types:
  - pickupable
  - station
  semantic: the specified pickupable is visibly resting in contact with the station's top surface.

  name: stacked_on
  types:
  - pickupable
  - pickupable
  semantic: the first pickupable is visibly in direct contact and vertically supported by the second pickupable, with no other object between them.

  name: top_surface_clear
  types:
  - pickupable
  semantic: no pickupable object is in direct vertical contact resting on the specified pickupable.

  name: on_stove_surface
  types:
  - cookable
  semantic: the specified cookable is visibly in direct contact with and supported by the stove's cooking surface (grill area), not held by any agent or resting on another object.
  
# Operators
(:action Pick_1
:parameters ( ?pickupable_p3 - pickupable  ?pickupable_p5     - pickupable  ?pickupable_p6 - pickupable  ?robot_p0 - robot  ?station_p1 -     station  ?station_p2 - station )
:precondition (and 
        (hand_empty ?robot_p0)
        (on_station ?pickupable_p6 ?station_p1)
        (top_surface_clear ?pickupable_p6)
        (not (holding ?robot_p0 ?pickupable_p3))
        (not (holding ?robot_p0 ?pickupable_p5))
        (not (holding ?robot_p0 ?pickupable_p6))
        (not (on_station ?pickupable_p6     ?station_p2))
        (not (station_surface_empty ?station_p1))
) 
:effect (and     
        (holding ?robot_p0 ?pickupable_p6) 
        (station_surface_empty ?station_p1) 
        (not (hand_empty ?robot_p0 ))
        (not (on_station ?pickupable_p6 ?station_p1))
     ) 
)

(:action Pick_2
:parameters ( ?pickupable_p6 - pickupable  ?robot_p0 - robot )
:precondition (and 
        (hand_empty ?robot_p0 )
        (top_surface_clear ?pickupable_p6 )
        (not (holding ?robot_p0 ?pickupable_p6))
) 
:effect (and 
        (holding ?robot_p0 ?pickupable_p6) 
        (not (hand_empty ?robot_p0))
 ) 
)

(:action Pick_3
:parameters ( ?pickupable_p3 - pickupable  ?pickupable_p4     - pickupable  ?pickupable_p6 - pickupable  ?robot_p0 - robot )
:precondition     (and 
        (hand_empty ?robot_p0 )
        (stacked_on ?pickupable_p4 ?pickupable_p3)
        (top_surface_clear ?pickupable_p4 )
        (not (holding ?robot_p0 ?pickupable_p3))
        (not (holding ?robot_p0 ?pickupable_p4))
        (not (holding ?robot_p0 ?pickupable_p6))
        (not (top_surface_clear ?pickupable_p3))
) 
:effect (and 
        (holding ?robot_p0 ?pickupable_p4) 
        (top_surface_clear ?pickupable_p3 ) 
        (not (hand_empty ?robot_p0 ))
        (not (stacked_on ?pickupable_p4 ?pickupable_p3))
 ) 
)

(:action Place_1
:parameters ( ?pickupable_p3 - pickupable  ?pickupable_p4     - pickupable  ?pickupable_p5 - pickupable  ?pickupable_p6 - pickupable  ?robot_p0     - robot  ?station_p2 - station )
:precondition (and 
        (holding ?robot_p0 ?pickupable_p4)
        (station_surface_empty ?station_p2 )
        (top_surface_clear ?pickupable_p4 )
        (not (hand_empty ?robot_p0 ))
        (not (holding ?robot_p0 ?pickupable_p5))
        (not (holding ?robot_p0 ?pickupable_p6))
        (not (on_station     ?pickupable_p4 ?station_p2))
        (not (stacked_on ?pickupable_p3 ?pickupable_p4))
        (not (stacked_on ?pickupable_p4 ?pickupable_p3))
) 
:effect (and 
        (hand_empty ?robot_p0 ) 
        (on_station ?pickupable_p4 ?station_p2) 
        (not (holding ?robot_p0 ?pickupable_p4))
        (not (station_surface_empty ?station_p2 ))
 ) 
)

(:action Place_2
:parameters ( ?pickupable_p5 - cookable  ?pickupable_p3     - pickupable  ?pickupable_p4 - pickupable  ?pickupable_p6 - pickupable  ?robot_p0     - robot  ?station_p1 - station )
:precondition (and 
        (holding ?robot_p0 ?pickupable_p5)
        (station_surface_empty ?station_p1)
        (top_surface_clear ?pickupable_p5)
        (not (hand_empty ?robot_p0))
        (not (holding ?robot_p0 ?pickupable_p3))
        (not (holding ?robot_p0 ?pickupable_p4))
        (not (holding ?robot_p0 ?pickupable_p6))
        (not (on_station ?pickupable_p3 ?station_p1))
        (not (on_station ?pickupable_p5 ?station_p1))
        (not (on_stove_surface ?pickupable_p5))
        (not (stacked_on ?pickupable_p5 ?pickupable_p6))
        (not (stacked_on ?pickupable_p6 ?pickupable_p5))
) 
:effect (and 
        (hand_empty ?robot_p0 ) 
        (on_station ?pickupable_p5 ?station_p1) 
        (on_stove_surface ?pickupable_p5 ) 
        (not (holding ?robot_p0 ?pickupable_p5))
        (not (station_surface_empty ?station_p1 ))
 ) 
)

(:action Stack_1
:parameters ( ?pickupable_p3 - pickupable  ?pickupable_p4     - pickupable  ?pickupable_p5 - pickupable  ?robot_p0 - robot )
:precondition (and 
        (holding ?robot_p0 ?pickupable_p3)
        (top_surface_clear ?pickupable_p3 )
        (top_surface_clear ?pickupable_p4 )
        (not (hand_empty ?robot_p0 ))
        (not (holding ?robot_p0 ?pickupable_p4))
        (not (holding ?robot_p0 ?pickupable_p5))
        (not (stacked_on ?pickupable_p3 ?pickupable_p4))
    ) 
:effect (and 
        (hand_empty ?robot_p0 ) 
        (stacked_on ?pickupable_p3 ?pickupable_p4)     
        (not (holding ?robot_p0 ?pickupable_p3))
        (not (top_surface_clear ?pickupable_p4     ))
 ) 
)

(:action Cut_1 
:parameters ( ?cuttable_p4 - cuttable  ?pickupable_p3 - pickupable      ?pickupable_p6 - pickupable  ?robot_p0 - robot  ?station_p2 - station )
:precondition (and 
        (hand_empty ?robot_p0 )
        (on_station ?cuttable_p4 ?station_p2)
        (top_surface_clear ?cuttable_p4 )
        (not (chopped ?cuttable_p4 ))
        (not (holding ?robot_p0 ?cuttable_p4))
        (not (holding ?robot_p0 ?pickupable_p6))
        (not (stacked_on ?pickupable_p3 ?cuttable_p4))
    ) 
:effect (and 
        (chopped ?cuttable_p4 ) 
 ) 
)

(:action Cook_1
:parameters ( ?cookable_p5 - cookable  ?pickupable_p3 - pickupable      ?pickupable_p4 - pickupable  ?pickupable_p6 - pickupable  ?robot_p0 - robot      ?station_p1 - station )
:precondition (and 
        (hand_empty ?robot_p0 )
        (on_station ?cookable_p5 ?station_p1)
        (on_stove_surface ?cookable_p5 )
        (top_surface_clear ?cookable_p5 )
        (not (cooked ?cookable_p5))
        (not (holding ?robot_p0 ?cookable_p5))
        (not (holding ?robot_p0 ?pickupable_p3))
        (not (holding ?robot_p0 ?pickupable_p4))
        (not (holding ?robot_p0 ?pickupable_p6))
        (not (stacked_on ?cookable_p5 ?pickupable_p6))
        (not (stacked_on ?pickupable_p6 ?cookable_p5))
) 
:effect (and 
        (cooked ?cookable_p5 ) 
 ) 
)

\end{lstlisting}

\begin{lstlisting}[language=Lisp,mathescape=true, deletekeywords={and, not, or,  first, second}]
# Learned Predicates & Operators of Franka domain

# Predicates
  name: gripper_empty
  types:
  - robot
  semantic: the robot's gripper is visually not holding any object.

  name: holding
  types:
  - robot
  - pickupable
  semantic: the robot's gripper is visually grasping the specified pickupable object.

  name: mug_full
  types:
  - container
  semantic: the mug container's visible contents reach the inner rim or above (no empty interior region is visible above the contents).

  name: plate_top_unoccupied
  types:
  - plate
  semantic: no object is visibly resting in contact with the upper surface of the specified plate (its food area is empty).

  name: stacked_on
  types:
  - pickupable
  - plate
  semantic: the specified pickupable object is visually resting in direct contact with the upper surface of the specified plate.

  name: plate_is_dirty
  types:
  - plate
  semantic: the specified plate's upper surface contains visible food traces whose appearance differs from the plate's base surface (visible residue to be wiped).

# Operators
(:action Pick_1 
:parameters ( ?pickupable_p1 - pickupable  ?robot_p2 - robot )
:precondition (and 
        (gripper_empty ?robot_p2 )
        (not (holding ?robot_p2 ?pickupable_p1))
) 
:effect (and 
        (holding ?robot_p2 ?pickupable_p1) 
        (not (gripper_empty ?robot_p2 ))
 ) 
)

(:action Pick_2 
:parameters ( ?pickupable_p0 - pickupable  ?pickupable_p1 - pickupable  ?pickupable_p3 - pickupable  ?plate_p4 - plate  ?robot_p2 - robot )
:precondition (and 
        (gripper_empty ?robot_p2 )
        (stacked_on ?pickupable_p1 ?plate_p4)
        (not (holding ?robot_p2 ?pickupable_p0))
        (not (holding ?robot_p2 ?pickupable_p1))
        (not (holding ?robot_p2 ?pickupable_p3))
        (not (plate_is_dirty ?plate_p4 ))
        (not (plate_top_unoccupied ?plate_p4 ))
) 
:effect (and 
        (holding ?robot_p2 ?pickupable_p1) 
        (plate_top_unoccupied ?plate_p4 ) 
        (not (gripper_empty ?robot_p2 ))
        (not (stacked_on ?pickupable_p1 ?plate_p4))
 ) 
)

(:action Place_1
:parameters ( ?pickupable_p1 - pickupable  ?robot_p2 - robot )
:precondition (and 
        (holding ?robot_p2 ?pickupable_p1)
        (not (gripper_empty ?robot_p2 ))
) 
:effect (and 
        (gripper_empty ?robot_p2 ) 
        (not (holding ?robot_p2 ?pickupable_p1))
 ) 
)

(:action Pour_1
:parameters ( ?container_p0 - container  ?pickupable_p0 - pickupable ?pickupable_p3 - pourable  ?robot_p2 - robot )
:precondition (and 
        (holding ?robot_p2 ?pickupable_p3)
        (not (gripper_empty ?robot_p2 ))
        (not (holding ?robot_p2 ?pickupable_p0))
        (not (mug_full ?container_p0 ))
) 
:effect (and 
        (mug_full ?container_p0 ) 
 ) 
)

(:action Stack_1
 :parameters (?pickupable_p0 - pickupable
              ?pickupable_p1 - pickupable
              ?pickupable_p3 - pickupable
              ?plate_p4 - plate
              ?robot_p2 - robot)
 :precondition (and
        (holding ?robot_p2 ?pickupable_p1)
        (plate_top_unoccupied ?plate_p4)
        (not (gripper_empty ?robot_p2))
        (not (holding ?robot_p2 ?pickupable_p0))
        (not (holding ?robot_p2 ?pickupable_p3))
        (not (plate_is_dirty ?plate_p4))
        (not (stacked_on ?pickupable_p1 ?plate_p4)))
 :effect (and
        (gripper_empty ?robot_p2)
        (stacked_on ?pickupable_p1 ?plate_p4)
        (not (holding ?robot_p2 ?pickupable_p1))
        (not (plate_top_unoccupied ?plate_p4))))
    
(:action Wipe_1
:parameters ( ?pickupable_p0 - sponge  ?plate_p4 - plate      ?robot_p2 - robot )
:precondition (and 
        (holding ?robot_p2 ?pickupable_p0)
        (plate_is_dirty ?plate_p4 )
        (not (gripper_empty ?robot_p2 ))
) 
:effect (and 
        (not (plate_is_dirty ?plate_p4 ))
 ) 
)
\end{lstlisting}

\begin{lstlisting}[language=Lisp,mathescape=true, deletekeywords={and, not, or,  first, second}]
# Learned Predicates & Operators of Bi-manual Kuka domain

# Predicates
  name: InLeftGripper
  types:
  - robot
  - openable
  semantic: the openable object is currently grasped by the robot's left gripper, with the gripper fingers in contact enclosing the object.

  name: InRightGripper
  types:
  - robot
  - utensil
  semantic: the utensil is physically enclosed and held by the robot's right gripper, with the gripper fingers in contact around the utensil.
  
  name: RightGripperEmpty
  types:
  - robot
  semantic: the robot's right gripper is visibly not holding anything. the fingers are not enclosing or contacting any part of the utensil, and the grasping space is empty.

  name: LidOff
  types:
  - openable
  semantic: the openable object's lid/cover is not attached; its opening is exposed and the interior is visible.

  name: LeftGripperEmpty
  types:
  - robot
  semantic: the robot's left gripper is visibly not holding any object. the fingers are not enclosing or contacting anything and the grasping space is clear.

  name: InContainer
  types:
  - utensil
  semantic: the utensil is located within another object's interior volume, with part of it below that object's rim and laterally enclosed by its walls.

  name: OpenableOnTable
  types:
  - openable
  semantic: the openable object's base is in contact with and supported from below by the tabletop (resting stably on the surface).

  name: UtensilOnTable
  types:
  - utensil
  semantic: the utensil is resting on and supported by the tabletop, with part of it visibly contacting the table surface (i.e., not enclosed within another object).

  name: Closed
  types:
  - openable
  semantic: the openable's lid is visibly attached and covering its opening; the interior is not visible.
  
  name: Coated
  types:
  - utensil
  semantic: a visible layer or clump of material adheres to the utensil's working end (e.g., the blade shows a smear that was absent before).

  name: SpreadOn
  types:
  - food
  semantic: the food's top surface shows a visible layer or patch of spread material (e.g., a smear) on it.

  name: HeldByRobot
  types:
  - robot
  - openable
  semantic: the openable object is currently enclosed and held by any of the robot's grippers (left or right), with the gripper fingers in contact around the object.
 
# Operators
(:action LeftArmPick_1
:parameters ( ?openable_p0 - openable  ?robot_p1 - robot )
:precondition (and 
        (Closed ?openable_p0 )
        (LeftGripperEmpty ?robot_p1 )
        (OpenableOnTable ?openable_p0 )
        (not (HeldByRobot ?robot_p1 ?openable_p0))
        (not (InLeftGripper ?robot_p1 ?openable_p0))
        (not (LidOff ?openable_p0 ))
) 
:effect (and 
        (HeldByRobot ?robot_p1 ?openable_p0)     
        (InLeftGripper ?robot_p1 ?openable_p0) 
        (not (LeftGripperEmpty ?robot_p1 ))
        (not (OpenableOnTable ?openable_p0 ))
 ) 
)

(:action RightArmPick_1
:parameters ( ?robot_p1 - robot  ?utensil_p2 - utensil )
:precondition (and 
        (InContainer ?utensil_p2 )
        (RightGripperEmpty ?robot_p1 )
        (not (Coated ?utensil_p2 ))
        (not (InRightGripper ?robot_p1 ?utensil_p2))
        (not (UtensilOnTable ?utensil_p2 ))
) 
:effect (and 
        (InRightGripper ?robot_p1 ?utensil_p2) 
        (not (InContainer ?utensil_p2 ))
        (not (RightGripperEmpty ?robot_p1 ))
 ) 
)

(:action Drop_1
:parameters ( ?robot_p1 - robot  ?utensil_p2 - utensil )
:precondition (and 
        (InRightGripper ?robot_p1 ?utensil_p2)
        (not (InContainer ?utensil_p2 ))
        (not (RightGripperEmpty ?robot_p1 ))
        (not (UtensilOnTable ?utensil_p2 ))
) 
:effect (and 
        (RightGripperEmpty ?robot_p1 ) 
        (UtensilOnTable ?utensil_p2 ) 
        (not (InRightGripper ?robot_p1 ?utensil_p2))
 ) 
)

(:action Open_1
:parameters ( ?openable_p0 - openable  ?robot_p1 - robot ?utensil_p2 - utensil )
:precondition (and 
        (Closed ?openable_p0 )
        (HeldByRobot ?robot_p1 ?openable_p0)
        (InLeftGripper ?robot_p1 ?openable_p0)
        (RightGripperEmpty ?robot_p1 )
        (not (InRightGripper ?robot_p1 ?utensil_p2))
        (not (LeftGripperEmpty ?robot_p1 ))
        (not (LidOff ?openable_p0 ))
        (not (OpenableOnTable ?openable_p0 ))
) 
:effect (and 
        (LidOff ?openable_p0 ) 
        (not (Closed ?openable_p0 ))
 ) 
)

(:action Scoop_1
 :parameters (?openable_p0 - openable
              ?robot_p1 - robot
              ?utensil_p2 - utensil)
 :precondition (and
        (HeldByRobot ?robot_p1 ?openable_p0)
        (InLeftGripper ?robot_p1 ?openable_p0)
        (InRightGripper ?robot_p1 ?utensil_p2)
        (LidOff ?openable_p0)
        (not (Closed ?openable_p0))
        (not (Coated ?utensil_p2))
        (not (InContainer ?utensil_p2))
        (not (LeftGripperEmpty ?robot_p1))
        (not (OpenableOnTable ?openable_p0))
        (not (RightGripperEmpty ?robot_p1))
        (not (UtensilOnTable ?utensil_p2)))
 :effect (and
        (Coated ?utensil_p2)))

(:action Spread_1
:parameters ( ?food_p0 - food  ?openable_p0 - openable ?robot_p1 - robot  ?utensil_p2 - utensil )
:precondition (and 
        (Coated ?utensil_p2 )
        (HeldByRobot ?robot_p1 ?openable_p0)
        (InLeftGripper ?robot_p1 ?openable_p0)
        (InRightGripper ?robot_p1 ?utensil_p2)
        (not (InContainer ?utensil_p2 ))
        (not (LeftGripperEmpty ?robot_p1 ))
        (not (RightGripperEmpty ?robot_p1 ))
        (not (SpreadOn ?food_p0 ))
        (not (UtensilOnTable ?utensil_p2 ))
) 
:effect (and 
        (SpreadOn ?food_p0 ) 
 ) 
)
\end{lstlisting}

Additionally, we present here predicates and operators written by the PDDL expert for each environment:

\begin{lstlisting}[language=Lisp,mathescape=true, deletekeywords={and, not, or,  first, second, null, open}]
# Predicates & Operators Written by PDDL Expert

## Robotouille Domain
# Predicates
  name: hand_empty
  types: 
  null
  semantic: The agent's hand is empty.

  name: is_holding
  types:
  - pickupable
  semantic: The agent is holding a `pickupable` object.

  name: obj_free
  types:
  - pickupable
  semantic: A `pickupable` object has nothing on top.

  name: station_free
  types:
  - station
  semantic: A `station` is free and has nothing on top.

  name: is_on_top
  types:
  - pickupable
  - pickupable
  semantic: The first `pickupable` object is right on top of the second `pickupable` object with nothing in between.
  
  name: is_on_station
  types:
  - pickupable
  - station
  semantic: A `pickupable` object is on top of a `station`.

  name: is_cut
  types:
  - cuttable
  semantic: The `cuttable` object is cut or sliced.

  name: is_cooked
  types:
  - cookable
  semantic: The `cookable` object is cooked and ready for consumption or further meal preparation.

# Operators
(:action Pick
:parameters (?robot - robot ?pickupable - pickupable)     
:precondition (and     
        (hand_empty)     
        (obj_free ?pickupable)     
        (not (is_holding ?pickupable))
)     
:effect (and     
        (not (hand_empty))     
        (not (obj_free ?pickupable))     
        (is_holding ?pickupable)     
 ))

(:action PickFromStack
:parameters (?robot - robot ?top - pickupable ?bot - pickupable)     
:precondition (and     
        (hand_empty)     
        (not (is_holding ?top))     
        (obj_free ?top)     
        (is_on_top ?top ?bot)
)     
:effect (and     
        (not (hand_empty))     
        (is_holding ?top)     
        (obj_free ?bot)     
        (not (obj_free ?top))     
        (not (is_on_top ?top ?bot))     
)
)

(:action PickFromStation
:parameters (?robot - robot ?top - pickupable ?bot - station)     
:precondition (and     
        (hand_empty)     
        (not (is_holding ?top))     
        (obj_free ?top)     
        (is_on_station ?top ?bot)
)     
:effect (and     
        (not (hand_empty))     
        (is_holding ?top)     
        (station_free ?bot)     
        (not (obj_free ?top))     
        (not (is_on_station ?top ?bot))     
)
)

(:action Place
:parameters (?robot - robot ?pickupable - pickupable ?station - station)    
:precondition (and     
        (is_holding ?pickupable)      
        (station_free ?station)     
        (not (obj_free ?pickupable))
)     
:effect (and      
        (not (is_holding ?pickupable))      
        (obj_free ?pickupable)     
        (hand_empty)      
        (not (station_free ?station))     
        (is_on_station ?pickupable ?station)     
)
)

(:action Stack
:parameters (?robot - robot ?top - pickupable ?bot - pickupable)     
:precondition (and     
        (is_holding ?top)     
        (obj_free ?bot)     
        (not (obj_free ?top))
)     
:effect (and      
        (not (is_holding ?top))     
        (hand_empty)     
        (obj_free ?top)     
        (not (obj_free ?bot))     
        (is_on_top ?top ?bot)     
)
)
        
(:action Cut
:parameters (?robot - robot
        ?cuttable - cuttable
        ?board - cuttingboard)
:precondition (and
        (hand_empty)
        (obj_free ?cuttable)
        (is_on_station ?cuttable ?board)
        (not (is_cut ?cuttable)))
:effect (and
        (is_cut ?cuttable)))

(:action Cook
:parameters (?robot - robot ?cookable - cookable ?cooker - cooker)     
:precondition (and     
        (hand_empty)     
        (obj_free ?cookable)     
        (is_on_station ?cookable ?cooker)     
        (not (is_cooked ?cookable))
)     
:effect (and     
        (is_cooked ?cookable)     
)
)

## Franka Domain
# Predicates
  name: hand_empty
  types:
  - robot
  semantic: The `robot` is not holding anything in its gripper.

  name: is_holding
  types:
  - robot
  - pickupable
  semantic: The `robot` is holding a `pickupable` object in its gripper.

  name: is_clean
  types:
  - plate
  semantic: The `plate` object is cleaned after wiping.

  name: is_on_top
  types:
  - pickupable
  - plate
  semantic: A `pickupable` object is on top of a `plate` object.

  name: plate_empty
  types:
  - plate
  semantic: A `plate` object has nothing on top of it.

  name: container_filled
  types:
  - container
  semantic: A `container` has something inside of it.

# Operators
(:action Pick
:parameters (?robot - robot ?pickupable - pickupable)     
:precondition (and     
        (hand_empty ?robot)     
        (not (is_holding ?robot ?pickupable))     
)     
:effect (and     
        (not (hand_empty ?robot))     
        (is_holding ?robot ?pickupable)     
)      
)

(:action PickFromTable
:parameters (?robot - robot ?pickupable - pickupable ?plate - plate)     
:precondition (and     
        (hand_empty ?robot)     
        (not (is_holding ?robot ?pickupable))     
        (is_on_top ?pickupable ?plate)     
        (not (plate_empty ?plate))     
)     
:effect (and     
        (not (hand_empty ?robot))     
        (is_holding ?robot ?pickupable)     
        (not (is_on_top ?pickupable ?plate))     
        (plate_empty ?plate)     
)      
)

(:action Place
:parameters (?robot - robot ?pickupable - pickupable)     
:precondition (and     
        (not (hand_empty ?robot))     
        (is_holding ?robot ?pickupable)     
)     
:effect (and     
        (hand_empty ?robot)     
        (not (is_holding ?robot ?pickupable))     
)      
)

(:action Pour
:parameters (?robot - robot ?pourable - pourable ?container - container)    
:precondition (and     
        (is_holding ?robot ?pourable)     
        (not (container_filled ?container))     
)     
:effect (and     
        (container_filled ?container)     
)      
)

(:action Stack
:parameters (?robot - robot ?pickupable - pickupable ?plate - plate)     
:precondition (and     
        (not (hand_empty ?robot))     
        (is_holding ?robot ?pickupable)     
        (not (is_on_top ?pickupable ?plate))     
        (plate_empty ?plate)     
        (is_clean ?plate)     
)     
:effect (and     
        (hand_empty ?robot)     
        (not (is_holding ?robot ?pickupable))     
        (is_on_top ?pickupable ?plate)     
        (not (plate_empty ?plate))     
)      
)

(:action Wipe
:parameters (?robot - robot ?sponge - sponge ?plate - plate)     
:precondition (and     
        (is_holding ?robot ?sponge)     
        (not (is_clean ?plate))     
        (plate_empty ?plate)     
)     
:effect (and     
        (is_clean ?plate)     
)      
)

## Bi-manual Kuka Domain
# Predicates
  name: lefthand_empty
  types:
  - robot
  semantic: The `robot` is not holding anything in its left hand.

  name: righthand_empty
  types:
  - robot
  semantic: The `robot` is not holding anything in its right hand.

  name: in_lefthand
  types:
  - robot
  - openable
  semantic: The `robot` is holding an `openable` object in its left hand.

  name: in_righthand
  types:
  - robot
  - utensil
  semantic: The `robot` is holding an `utensil` object in its right hand.

  name: is_opened
  types: !!python/tuple
  - openable
  semantic: An `openable` object (such as a jar) is opened.

  name: is_clean
  types:
  - utensil
  semantic: A `utensil` object (such as a knife) is clean.

  name: is_upright
  types:
  - utensil
  semantic: A `utensil` object (such as a knife) is upright and graspable.

  name: contains_ingredient
  types: !!python/tuple
  - utensil
  semantic: A `utensil` object contains a spreadable ingredient.

  name: spread_on_top
  types:
  - food
  semantic: The `food` object contains a spreadable ingredient on top of it.

# Operators
(:action LeftArmPick
:parameters (?robot - robot ?openable - openable)     
:precondition (and      
        (lefthand_empty ?robot)     
        (not (in_lefthand ?robot ?openable))     
)     
:effect (and     
        (not (lefthand_empty ?robot))     
        (in_lefthand ?robot ?openable)     
)     
)

(:action RightArmPick
:parameters (?robot - robot ?utensil - utensil)     
:precondition (and      
        (righthand_empty ?robot)     
        (is_upright ?utensil)     
        (not (in_righthand ?robot ?utensil))     
)     
:effect (and     
        (not (righthand_empty ?robot))     
        (not (is_upright ?utensil))     
        (in_righthand ?robot ?utensil)     
)     
)

(:action Drop
:parameters (?robot - robot ?utensil - utensil)     
:precondition (and      
        (not (righthand_empty ?robot))     
        (in_righthand ?robot ?utensil)     
)     
:effect (and     
        (righthand_empty ?robot)     
        (not (in_righthand ?robot ?utensil))     
)     
)

(:action Open
:parameters (?robot - robot ?openable - openable)     
:precondition (and     
        (not (is_opened ?openable))      
        (in_lefthand ?robot ?openable)     
        (not (lefthand_empty ?robot))     
        (righthand_empty ?robot)     
)     
:effect (and     
        (is_opened ?openable)     
)     
)

(:action Scoop
:parameters (?robot - robot ?knife - utensil ?jar - openable)     
:precondition (and     
        (in_lefthand ?robot ?jar)     
        (in_righthand ?robot ?knife)     
        (is_opened ?jar)     
        (is_clean ?knife)     
        (not (contains_ingredient ?knife))     
)     
:effect (and     
        (contains_ingredient ?knife)     
        (not (is_clean ?knife))     
)     
)

(:action Spread
:parameters (?robot - robot ?knife - utensil ?bread - food)     
:precondition (and     
        (in_righthand ?robot ?knife)     
        (contains_ingredient ?knife)     
        (not (spread_on_top ?bread))      
)     
:effect (and     
        (not (contains_ingredient ?knife))      
        (spread_on_top ?bread)     
)     
)

\end{lstlisting}

\subsection{Example Tasks}
\begin{figure}[htbp]
  \centering

  \begin{subfigure}[b]{0.45\linewidth}
    \centering
    
    \includegraphics[width=\textwidth]{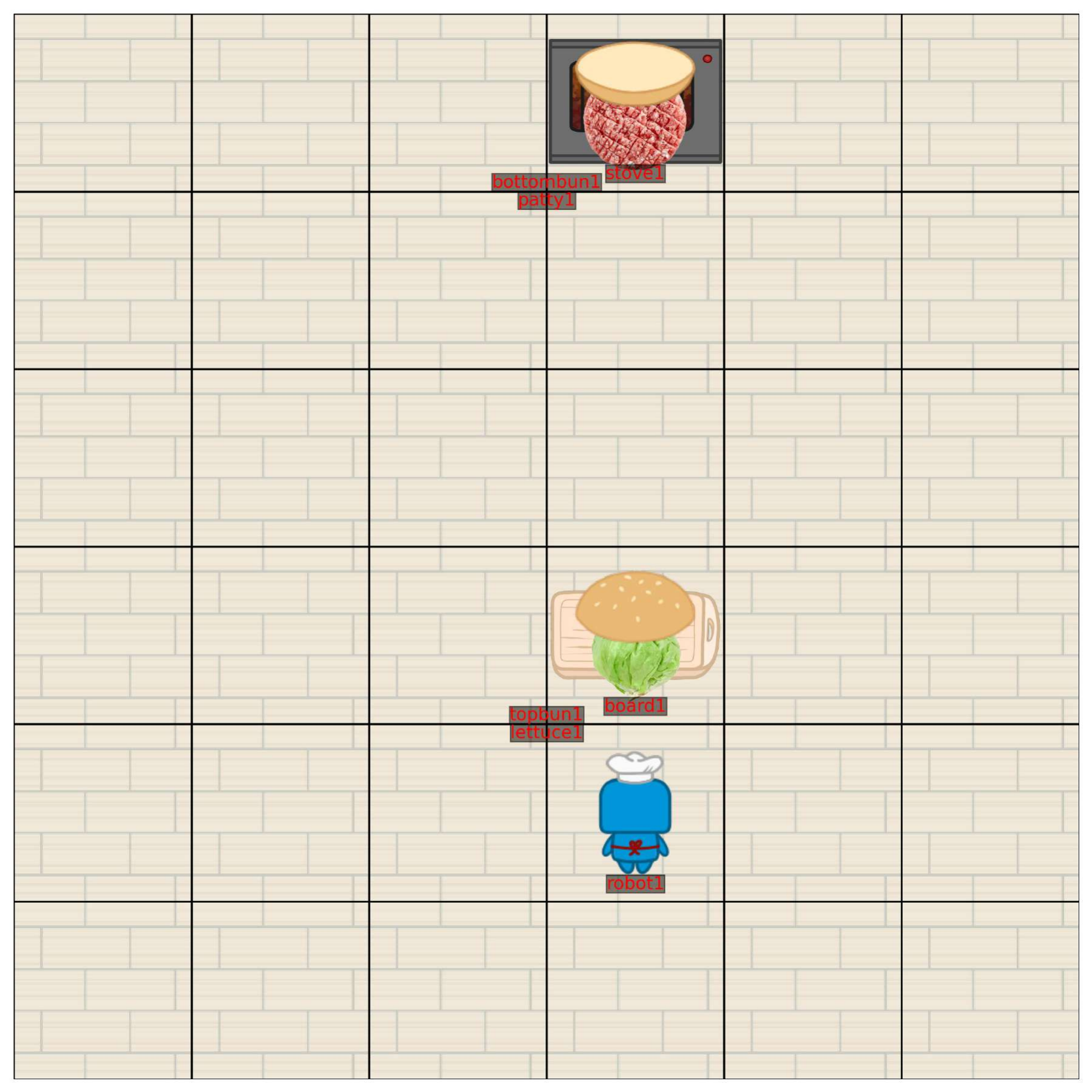}
    \caption{Initial state}
    \label{app:impossible_burger}
  \end{subfigure}
  \hfill 
  \begin{subfigure}[b]{0.45\linewidth}
    \centering
    \
    \includegraphics[width=\textwidth]{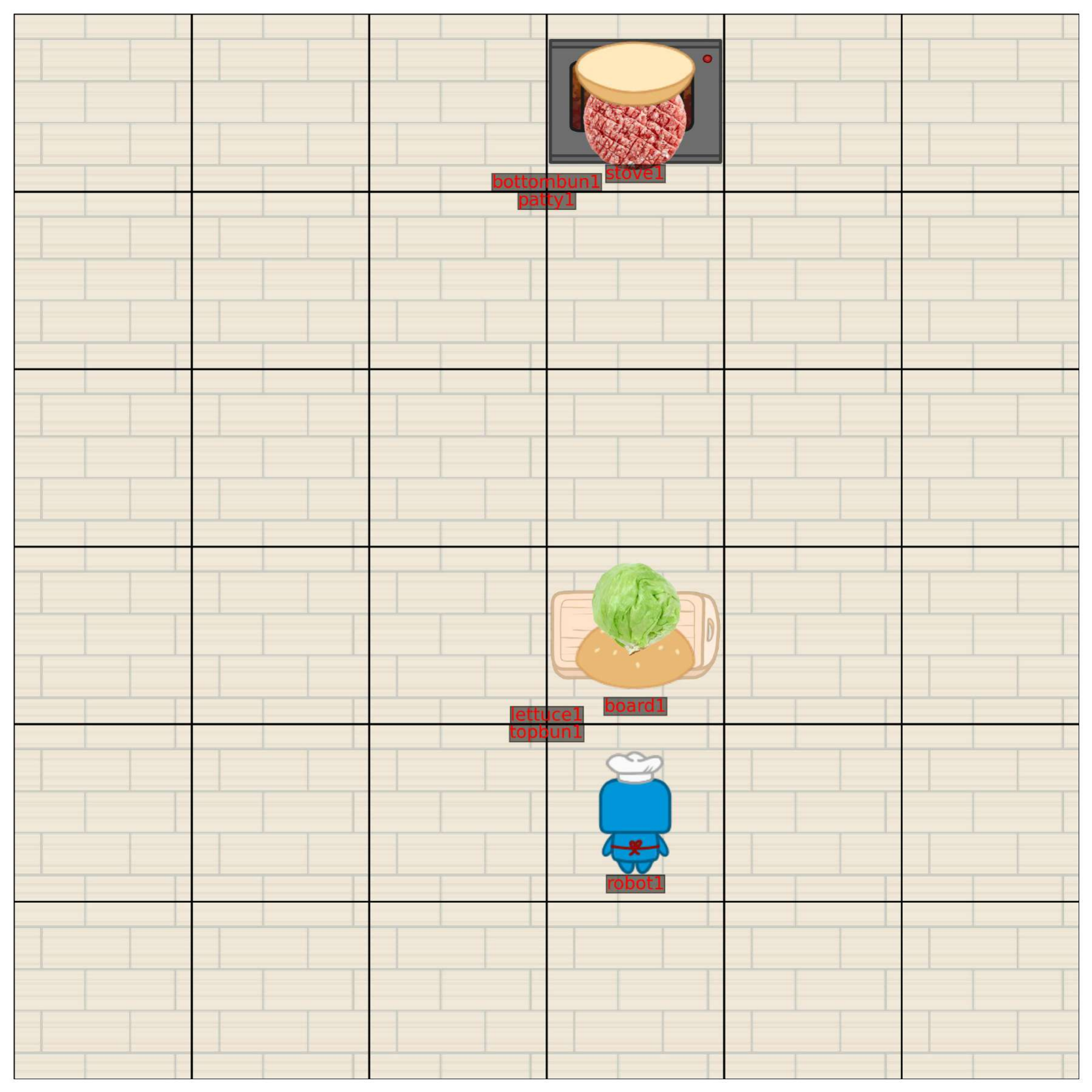}
    \caption{Goal state}
    \label{app:impossible_franka}
  \end{subfigure}

  \vspace{10pt} 

  \begin{subfigure}[b]{0.45\linewidth}
    \centering
    \includegraphics[width=\textwidth]{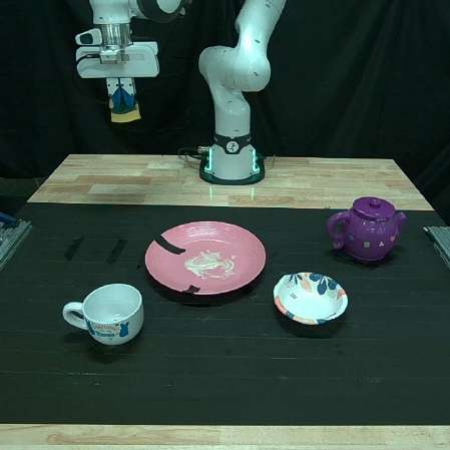}
    \caption{Initial state}
  \end{subfigure}
  \hfill 
  \begin{subfigure}[b]{0.45\linewidth}
    \centering
    \includegraphics[width=\textwidth]{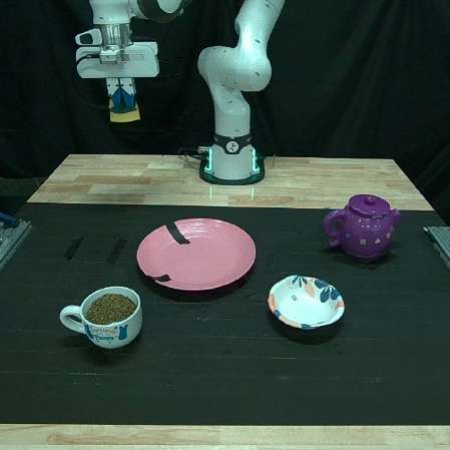}
    \caption{Goal state}
  \end{subfigure}

  \caption{Example Impossible Tasks In Robotouille.}
  \label{fig:impossible_tasks}
\end{figure}

We provide examples of the skill planning problems for evaluation. Specifically, Robotouille uses images generated by the simulator, Franka uses images taken by a camera in front of the robot, and Bimanual Kuka uses images taken by its egocentric camera. For impossible tasks, the impossibilities can be caused by not having a intermediate place for stack permutation (Fig.~\ref{app:impossible_burger}) or limiting by the current skill set (Fig.~\ref{app:impossible_franka}), e.g., cleaning up a filled mug or spreading ranch on the plate.

\begin{figure}[htbp]
  \centering

  \begin{subfigure}[b]{0.4\linewidth}
    \centering
    \includegraphics[width=\textwidth]{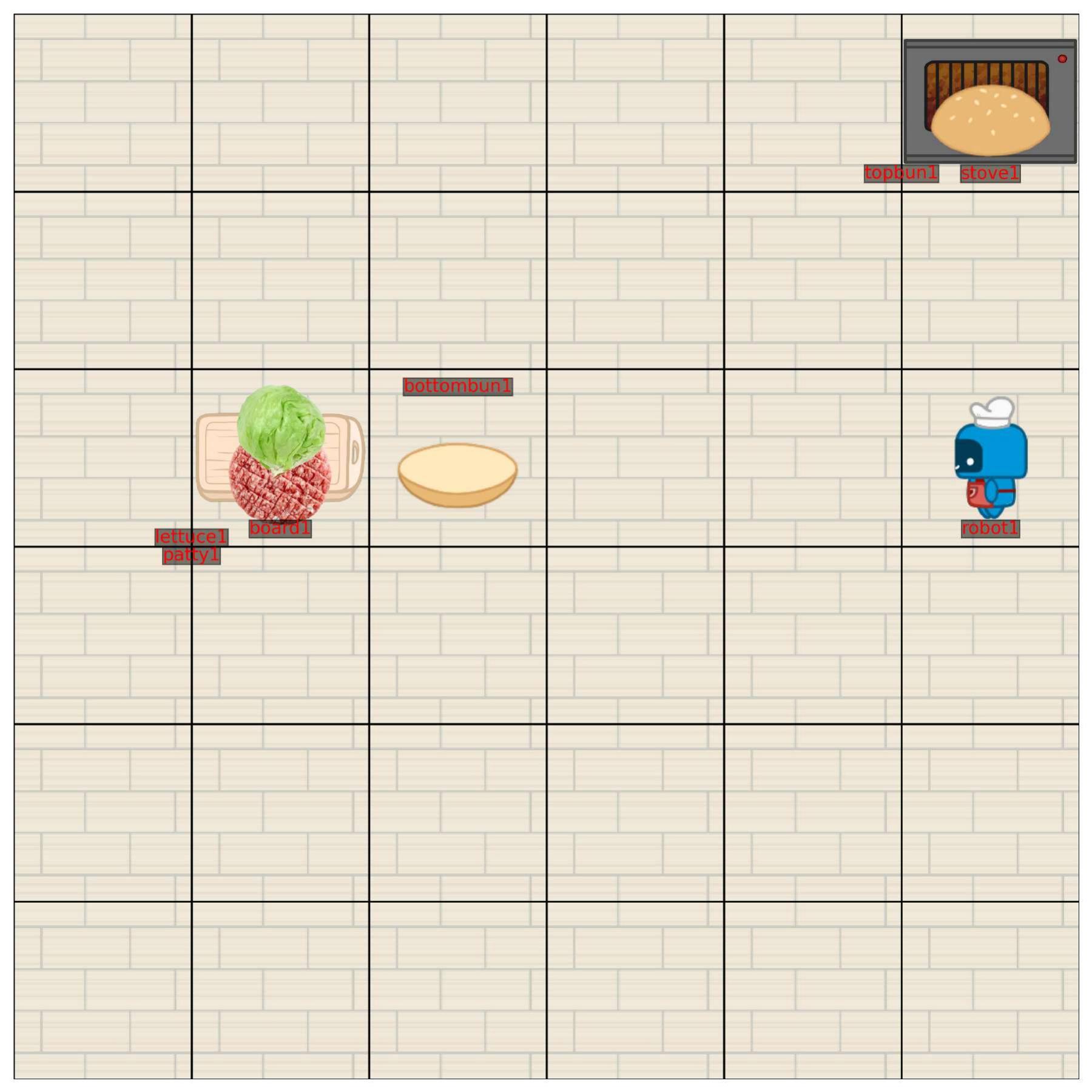}
    \caption{Initial state}
  \end{subfigure}
  \hspace{15pt}
  \begin{subfigure}[b]{0.4\linewidth}
    \centering
    \includegraphics[width=\textwidth]{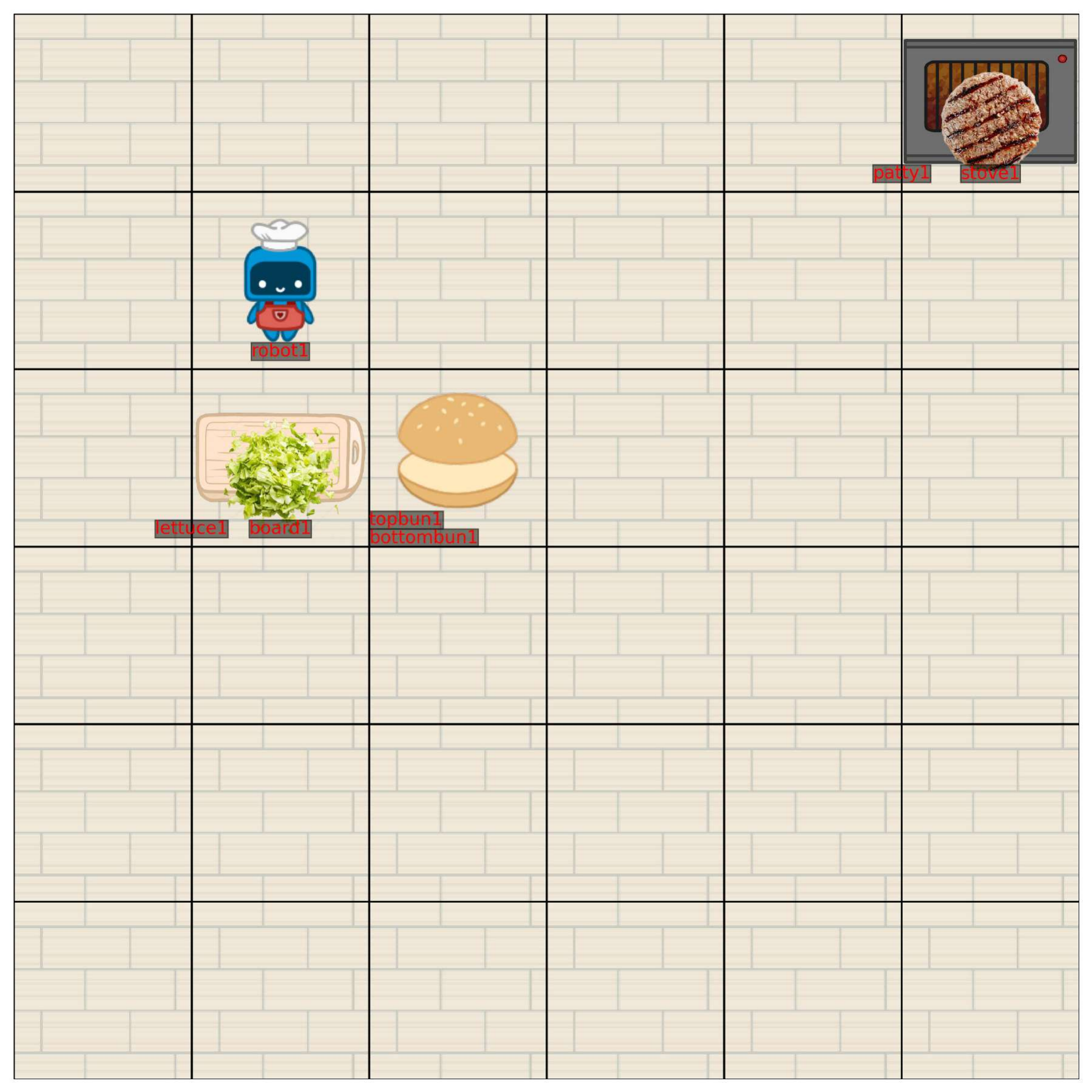}
    \caption{Goal state}
  \end{subfigure}

  \caption{Example task in Robotouille.}
\end{figure}

\begin{figure}[htbp]
  \centering
  \begin{subfigure}[b]{0.4\linewidth}
    \centering
    \includegraphics[width=\textwidth]{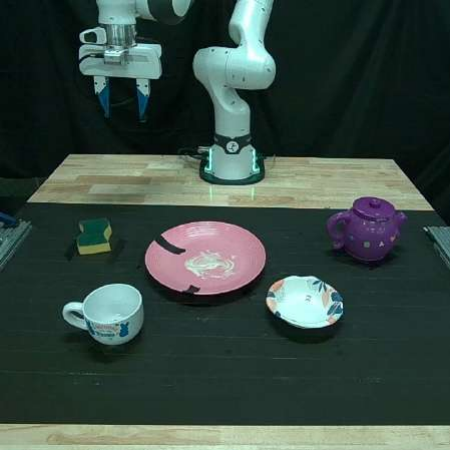}
    \caption{Initial state}
  \end{subfigure}
  \hspace{15pt}
  \begin{subfigure}[b]{0.4\linewidth}
    \centering
    \includegraphics[width=\textwidth]{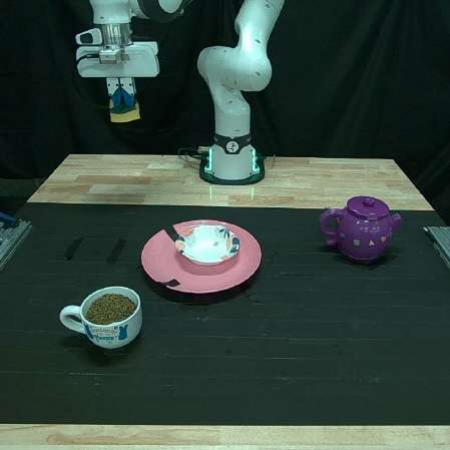}
    \caption{Goal state}
  \end{subfigure}

  \caption{Example task in Franka.}
\end{figure}
\begin{figure}[htbp]
  \centering

  \begin{subfigure}[b]{0.45\linewidth}
    \centering
    \includegraphics[width=\textwidth]{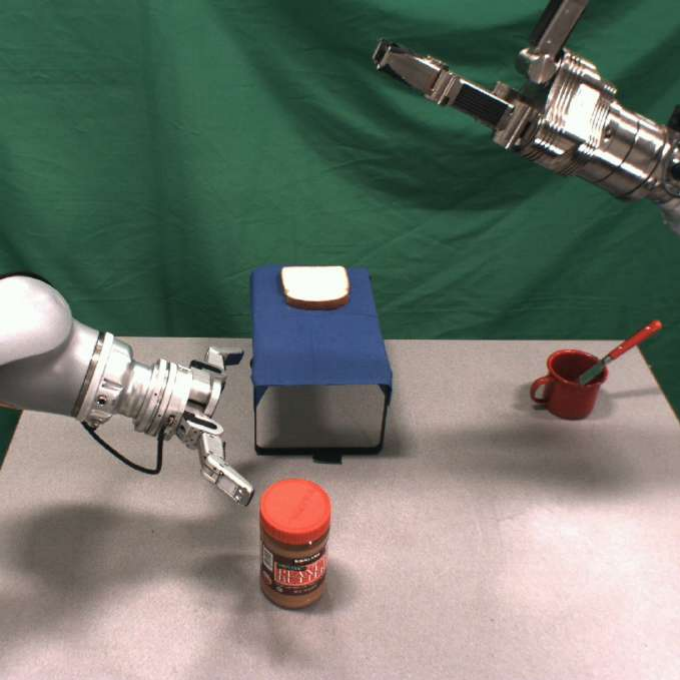}
    \caption{Initial state}
  \end{subfigure}
  \hfill
  \hspace{15pt}
  \begin{subfigure}[b]{0.45\linewidth}
    \centering
    \includegraphics[width=\textwidth]{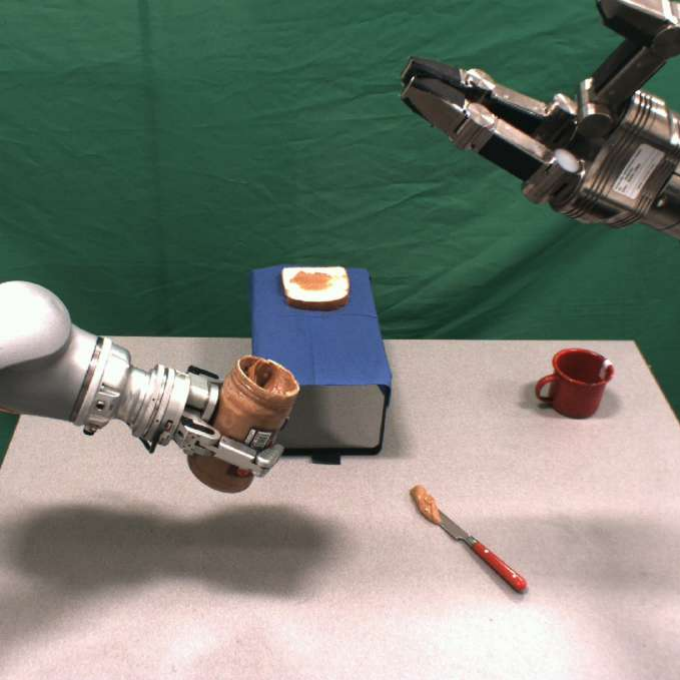}
    \caption{Goal state}
  \end{subfigure}

  \caption{Example task in Bimanual Kuka.}
\end{figure}

\clearpage

%% file: sections/appendix/vlm_study.tex
%

\subsection{Classification Accuracy}
We here analyze the classification accuracy of the vision-language model (VLM) used in both robotic experiments. Since the predicates are generated on the fly by the VLM, we do not have the ground truth values for them, and thus we must verify if the truth values match the images manually.

\paragraph{Per-predicate Classification Evaluation} Since predicates are originally lifted and can be grounded with different combinations of objects, we first define a classification over a low-level state of a grounded predicate as correct if (1) all parameters appear in the scene (if the predicate is not nullary) and (2) the truth value of the predicate match the low-level state specified by the image input. Then, we define a classification of a lifted predicate over a low-level state as correct if all of its grounded instances are classified correctly over that state.

\paragraph{Results and Analysis}
Over all predicates, the classification accuracy is $86.7\%$ for the Franka experiment, and $98.5\%$ for the bimanual Kuka experiment. Compared to the planning performance reported for both experiments, the classification accuracy is generally much higher. One reason for this mismatch is that, due to the rigidity of symbolic planning, even flipping the truth value of a single predicate can cause planning failure.
To investigate further, we identify predicates with poor classification accuracy and analyze how they harm planning performance in the next paragraph.

\paragraph{Per-predicate Accuracy}
The learned symbolic model of the Franka experiment contains $6$ predicates, which have $11$ possible grounded instances. The learned symbolic model of the bimanual Kuka experiment contains $12$ predicates, which have $13$ possible grounded instances. We evaluate per-predicate accuracy for both in Table~\ref{tab:franka_predicate_accuracy} and Table~\ref{tab:dorfl_predicate_accuracy}. From the results of the Franka experiment, we identify the two predicates, $\mathtt{gripper\_empty}$ and $\mathtt{holding}$, that caused all planning failures, and they fail almost simultaneously due to their semantic correlation. With further investigation, we found that the misclassifications were induced by a single object, $\mathtt{Sponge}$, which is possibly due to the color of the object and the background being too similar. In the bimanual Kuka experiment, it is $\mathtt{coated}$ (if the knife has peanut butter on it) that caused most of the planning failure, likely caused by the lighting conditions.  These observations suggest the accuracy of VLM is a limiting factor, and resolving them poses a promising path to improving the performance of \sys{}.

\begin{table}[H]
\centering
\caption{Per-predicate accuracy of Franka.}
\resizebox{\textwidth}{!}{%
\begin{tabular}{lcccccc}
\toprule
 & gripper\_empty & holding & mug\_full & plate\_top\_unoccupied & stacked\_on & plate\_is\_dirty \\
\midrule
Accuracy (\%) & 60.0 & 60.0 & 100.0 & 100.0 & 100.0 & 100.0 \\
\bottomrule
\end{tabular}
}
\label{tab:franka_predicate_accuracy}
\end{table}

\begin{table}[H]
\centering
\caption{Per-predicate accuracy of Bimanual Kuka.}
\small 
\resizebox{\textwidth}{!}{%
\begin{tabular}{lcccccc}
\toprule
 & InLeftGripper & InRightGripper & RightGripperEmpty & LeftGripperEmpty & LidOff & InContainer \\
\midrule
Accuracy (\%) & 100.0 & 100.0 & 100.0 & 100.0 & 100.0 & 98.3 \\
\midrule
 & OpenableOnTable & Closed & Coated & SpreadOn & HeldByRobot & UtensilOnTable \\
\midrule
Accuracy (\%) & 96.7 & 100.0 & 88.3 & 98.3 & 100.0 & 100.0 \\
\bottomrule
\end{tabular}
}
\label{tab:dorfl_predicate_accuracy}
\end{table}

\begin{figure}[H]
  \centering

  \begin{subfigure}[b]{0.3\linewidth}
    \centering
    \includegraphics[width=\textwidth]{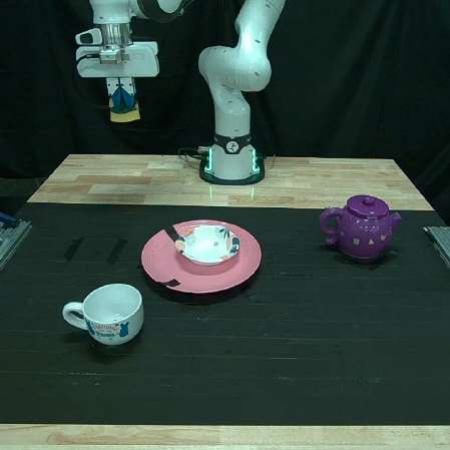}
    \caption{{\Large \textcolor{red}{\textbf{×}}}$\mathtt{holding(Sponge)} = \mathtt{\textbf{F}}$}
  \end{subfigure}
  \hspace{15pt}
  \begin{subfigure}[b]{0.3\linewidth}
    \centering
    \includegraphics[width=\textwidth]{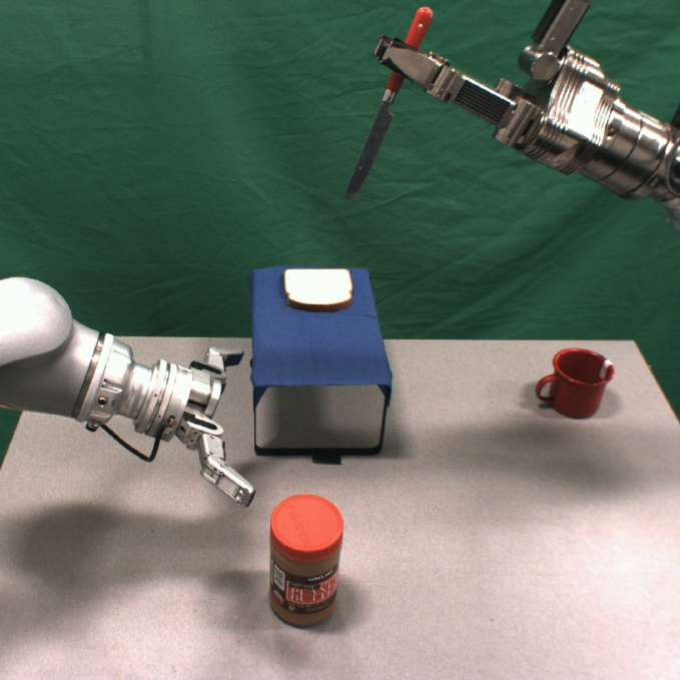}
    \caption{{\Large \textcolor{red}{\textbf{×}}}$\mathtt{coated(Knife)} = \mathtt{\textbf{T}}$}
  \end{subfigure}

\end{figure}

\subsection{Real-world Robustness}
To evaluate the real-world robustness of the VLM, we additionally conduct experiments to investigate factors such as viewpoints, lighting conditions, or domain shifts. For each of them, we collect a held-out set of images by varying these factors. We report per-predicate accuracy, and all numbers are averaged across three individual runs.

\paragraph{Viewpoints}
We collect visual observation data from two viewpoints and sample five configurations for each viewpoint. From the results of Franka experiments, we observe that the classification accuracies of certain predicates are higher from the viewpoint closer to the corresponding objects: at viewpoint \#1, all predicates can be perfectly classified, while predicates involving gripper or mug, such as $\mathtt{gripper\_empty}, \mathtt{holding}$ and $\mathtt{mug\_full}$, are significantly lower from viewpoint \#2, which is farther from the objects. In bimanual Kuka experiments, the result is mostly stable across different viewpoints and generally better than in the Franka environment, which is possibly due to fewer background distractions. Though the accuracy varies across viewpoints, its performance remains reliable as long as the full observability assumption still holds.

\begin{table}[H]
\centering
\caption{Per-predicate accuracy of Franka at Viewpoint \#1.}
\resizebox{\textwidth}{!}{%
\begin{tabular}{lcccccc}
\toprule
 & gripper\_empty & holding & mug\_full & plate\_top\_unoccupied & stacked\_on & plate\_is\_dirty \\
\midrule
Accuracy (\%) & 100.0 & 100.0 & 100.0 & 100.0 & 100.0 & 100.0 \\
\bottomrule
\end{tabular}
}
\label{tab:franka_predicate_accuracy_v1}
\end{table}

\begin{table}[H]
\centering
\caption{Per-predicate accuracy of Franka at Viewpoint \#2.}
\resizebox{\textwidth}{!}{%
\begin{tabular}{lcccccc}
\toprule
 & gripper\_empty & holding & mug\_full & plate\_top\_unoccupied & stacked\_on & plate\_is\_dirty \\
\midrule
Accuracy (\%) & 80.0 & 80.0 & 90.0 & 100.0 & 100.0 & 100.0 \\
\bottomrule
\end{tabular}
}
\label{tab:franka_predicate_accuracy_v2}
\end{table}

\begin{figure}[H]
  \centering
  \begin{subfigure}[b]{0.3\linewidth}
    \centering
    \includegraphics[width=\textwidth]{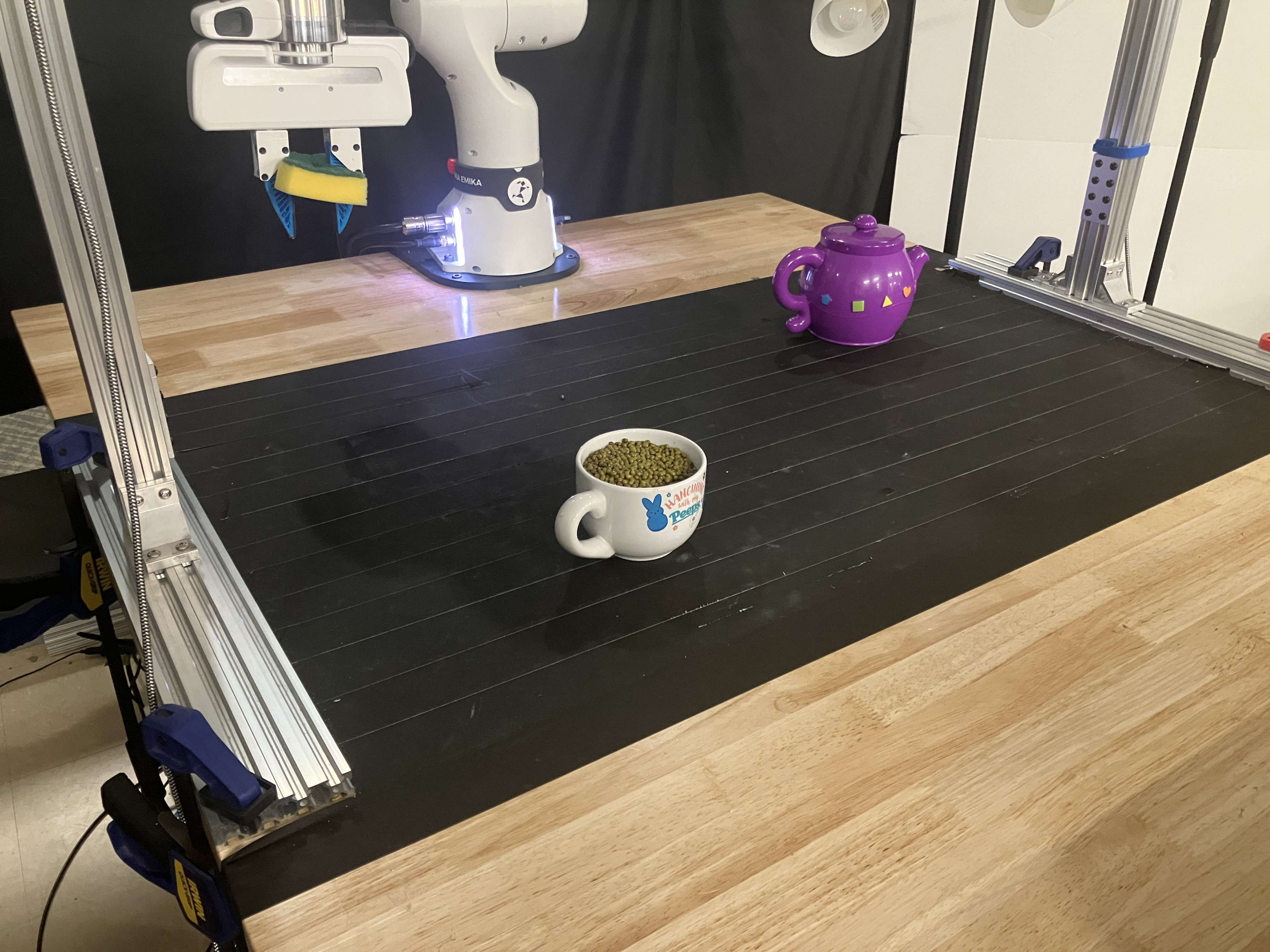}
    \caption{{\large \textcolor{DeepGreen}{\checkmark}}$\mathtt{holding(Sponge)} = \mathtt{\textbf{T}}$}
  \end{subfigure}
  \hspace{15pt}
  \begin{subfigure}[b]{0.3\linewidth}
    \centering
    \includegraphics[width=\textwidth]{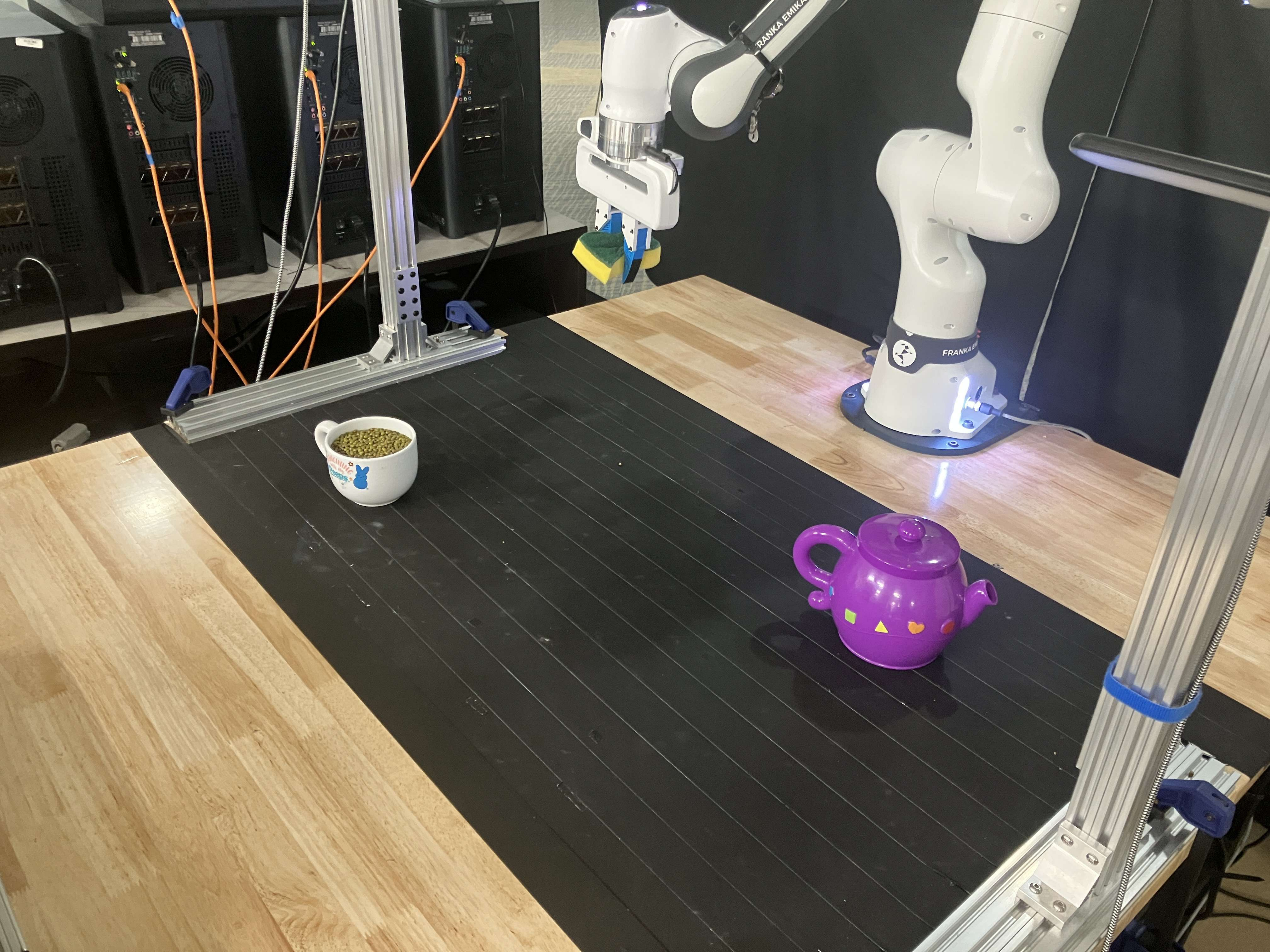}
    \caption{{\Large \textcolor{red}{\textbf{×}}}$\mathtt{holding(Sponge)} = \mathtt{\textbf{F}}$}
  \end{subfigure}
\end{figure}

\begin{table}[H]
\centering
\caption{Per-predicate accuracy of Bimanual Kuka at Viewpoint \#1.}
\small 
\resizebox{\textwidth}{!}{%
\begin{tabular}{lcccccc}
\toprule
 & InLeftGripper & InRightGripper & RightGripperEmpty & LeftGripperEmpty & LidOff & InContainer \\
\midrule
Accuracy (\%) & 100.0 & 100.0 & 100.0 & 100.0 & 100.0 & 100.0 \\
\midrule
 & OpenableOnTable & Closed & Coated & SpreadOn & HeldByRobot & UtensilOnTable \\
\midrule
Accuracy (\%) & 100.0 & 100.0 & 100.0 & 100.0 & 100.0 & 100.0 \\
\bottomrule
\end{tabular}
}
\label{tab:dorfl_predicate_accuracy_v1}
\end{table}

\begin{table}[H]
\centering
\caption{Per-predicate accuracy of Bimanual Kuka at Viewpoint \#2.}
\small 
\resizebox{\textwidth}{!}{%
\begin{tabular}{lcccccc}
\toprule
 & InLeftGripper & InRightGripper & RightGripperEmpty & LeftGripperEmpty & LidOff & InContainer \\
\midrule
Accuracy (\%) & 100.0 & 100.0 & 100.0 & 100.0 & 100.0 & 100.0 \\
\midrule
 & OpenableOnTable & Closed & Coated & SpreadOn & HeldByRobot & UtensilOnTable \\
\midrule
Accuracy (\%) & 100.0 & 100.0 & 100.0 & 100.0 & 100.0 & 93.3 \\
\bottomrule
\end{tabular}
}
\label{tab:dorfl_predicate_accuracy_v2}
\end{table}

\begin{figure}[H]
  \centering

  \begin{subfigure}[b]{0.35\linewidth}
    \centering
    \includegraphics[width=\textwidth]{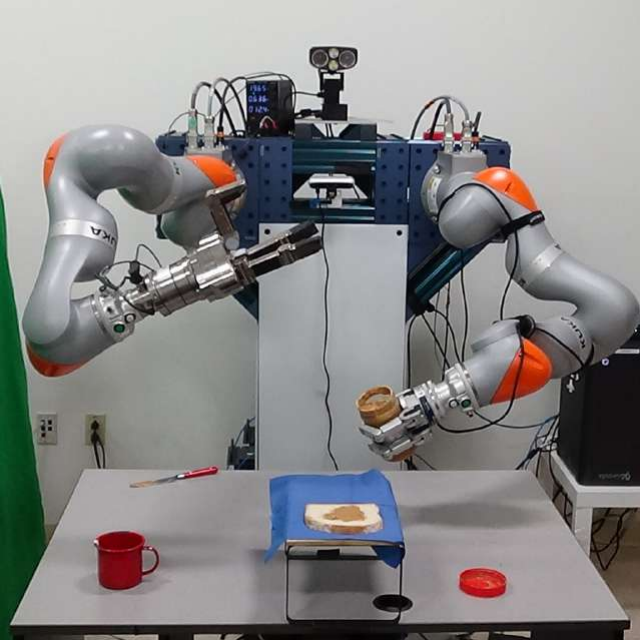}
    \caption{{\large \textcolor{DeepGreen}{\checkmark}}$\mathtt{UtensilOnTable(Knife)} = \mathtt{\textbf{T}}$}
  \end{subfigure}
  \hspace{15pt}
  \begin{subfigure}[b]{0.35\linewidth}
    \centering
    \includegraphics[width=\textwidth]{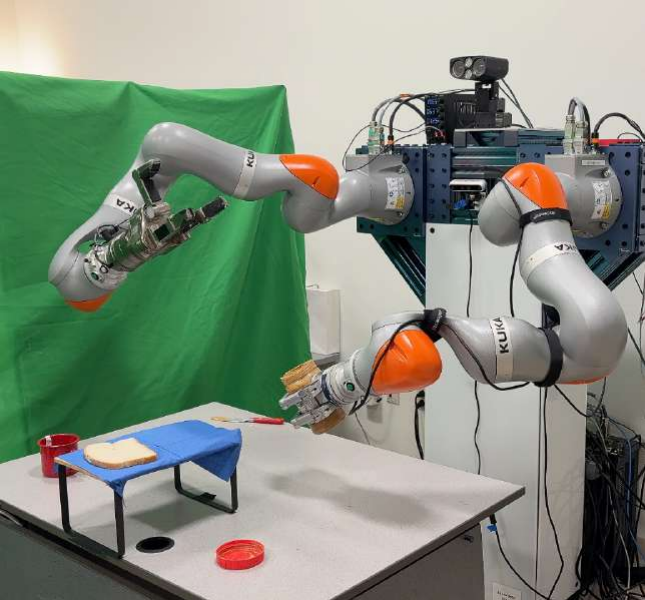}
    \caption{{\Large \textcolor{red}{\textbf{×}}}$\mathtt{UtensilOnTable(Knife)} = \mathtt{\textbf{F}}$}
  \end{subfigure}
\end{figure}

\paragraph{Lighting conditions}
We collect visual observation data under two lighting conditions and sample five configurations for each one. We find that the VLM is generally robust to different lighting conditions, except for several extremely difficult ones, such as under lighting condition \#2 in Bimanual Kuka, where the objects are heavily shadowed.

\begin{table}[H]
\centering
\caption{Per-predicate accuracy of Franka under Lighting Condition \#1.}
\resizebox{\textwidth}{!}{%
\begin{tabular}{lcccccc}
\toprule
 & gripper\_empty & holding & mug\_full & plate\_top\_unoccupied & stacked\_on & plate\_is\_dirty \\
\midrule
Accuracy (\%) & 90.0 & 90.0 & 100.0 & 96.7 & 100.0 & 100.0 \\
\bottomrule
\end{tabular}
}
\label{tab:franka_predicate_accuracy_l1}
\end{table}

\begin{table}[H]
\centering
\caption{Per-predicate accuracy of Franka under Lighting Condition \#2.}
\resizebox{\textwidth}{!}{%
\begin{tabular}{lcccccc}
\toprule
 & gripper\_empty & holding & mug\_full & plate\_top\_unoccupied & stacked\_on & plate\_is\_dirty \\
\midrule
Accuracy (\%) & 90.0 & 90.0 & 100.0 & 98.3 & 100.0 & 100.0 \\
\bottomrule
\end{tabular}}
\label{tab:franka_predicate_accuracy_l2}
\end{table}

\begin{figure}[H]
  \centering
  \begin{subfigure}[b]{0.4\linewidth}
    \centering
    \includegraphics[width=\textwidth]{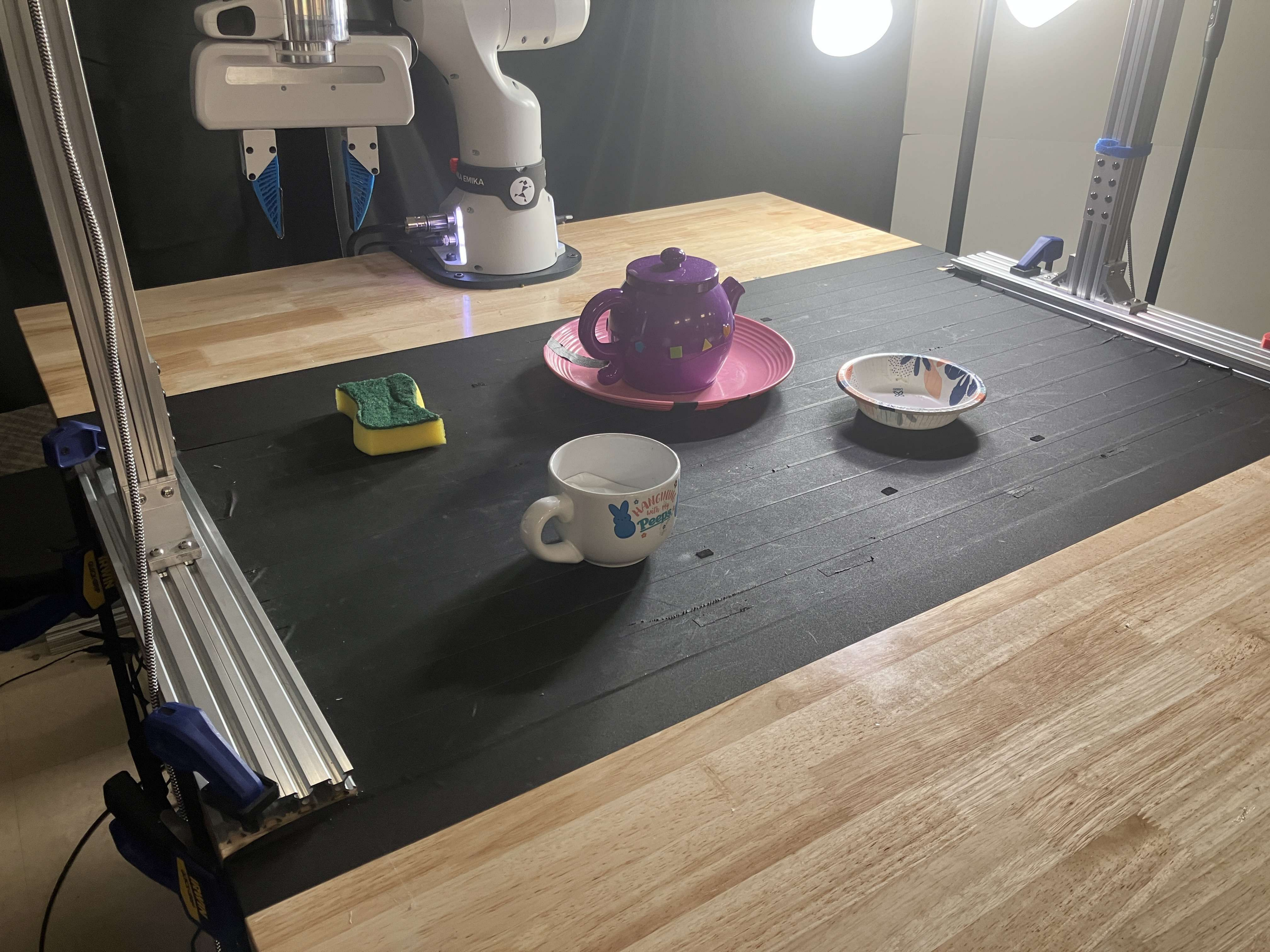}
    \caption{{\large \textcolor{DeepGreen}{\checkmark}}$\mathtt{stacked\_on(Teapot, Plate)} = \mathtt{\textbf{T}}$}
  \end{subfigure}
  \hspace{15pt}
  \begin{subfigure}[b]{0.4\linewidth}
    \centering
    \includegraphics[width=\textwidth]{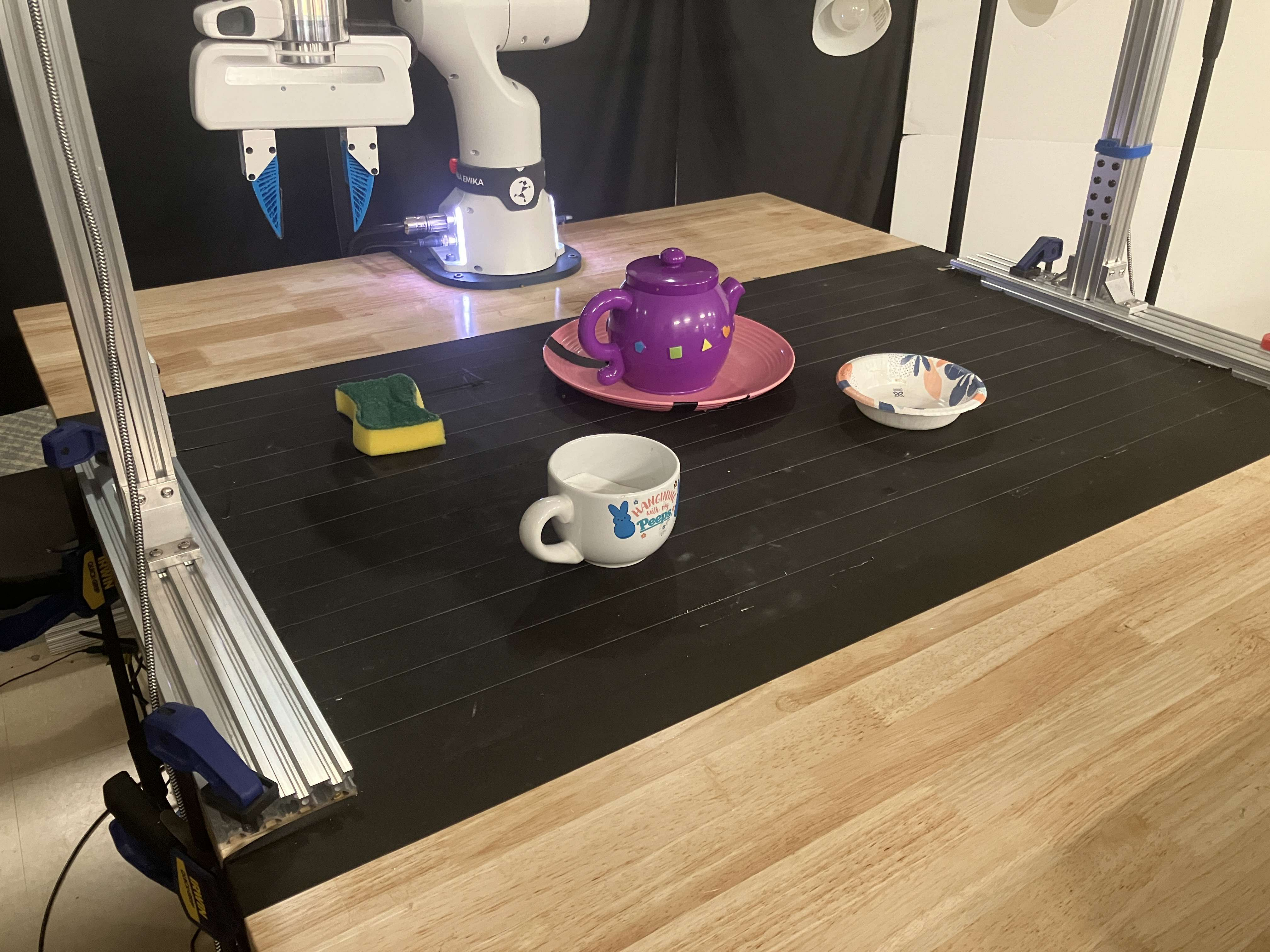}
    \caption{{\large \textcolor{DeepGreen}{\checkmark}}$\mathtt{stacked\_on(Teapot, Plate)} = \mathtt{\textbf{T}}$}
  \end{subfigure}
\end{figure}

\begin{table}[H]
\centering
\caption{Per-predicate accuracy of Bimanual Kuka under Lighting Condition \#1.}
\small 
\resizebox{\textwidth}{!}{%
\begin{tabular}{lcccccc}
\toprule
 & InLeftGripper & InRightGripper & RightGripperEmpty & LeftGripperEmpty & LidOff & InContainer \\
\midrule
Accuracy (\%) & 100.0 & 100.0 & 100.0 & 100.0 & 100.0 & 100.0 \\
\midrule
 & OpenableOnTable & Closed & Coated & SpreadOn & HeldByRobot & UtensilOnTable \\
\midrule
Accuracy (\%) & 100.0 & 100.0 & 100.0 & 100.0 & 100.0 & 100.0 \\
\bottomrule
\end{tabular}
}
\label{tab:dorfl_predicate_accuracy_l1}
\end{table}

\begin{table}[H]
\centering
\caption{Per-predicate accuracy of Bimanual Kuka under Lighting Condition \#2.}
\small 
\resizebox{\textwidth}{!}{%
\begin{tabular}{lcccccc}
\toprule
 & InLeftGripper & InRightGripper & RightGripperEmpty & LeftGripperEmpty & LidOff & InContainer \\
\midrule
Accuracy (\%) & 100.0 & 80.0 & 80.0 & 100.0 & 100.0 & 73.3 \\
\midrule
 & OpenableOnTable & Closed & Coated & SpreadOn & HeldByRobot & UtensilOnTable \\
\midrule
Accuracy (\%) & 100.0 & 100.0 & 100.0 & 86.7 & 100.0 & 100.0 \\
\bottomrule
\end{tabular}
}
\label{tab:dorfl_predicate_accuracy_l2}
\end{table}

\begin{figure}[H]
  \centering
  \begin{subfigure}[b]{0.35\linewidth}
    \centering
    \includegraphics[width=\textwidth]{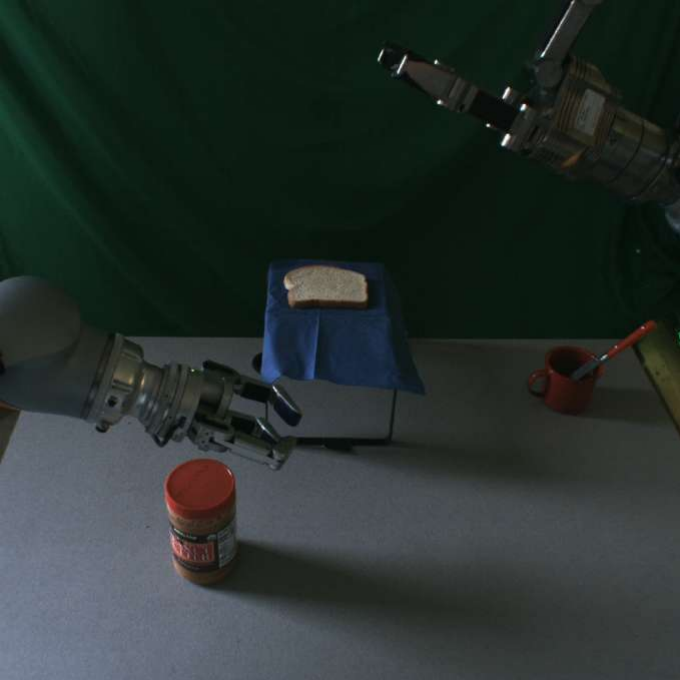}
    \caption{{\large \textcolor{DeepGreen}{\checkmark}}$\mathtt{RightGripperEmpty()} = \mathtt{\textbf{T}}$}
  \end{subfigure}
  \hspace{15pt}
  \begin{subfigure}[b]{0.35\linewidth}
    \centering
    \includegraphics[width=\textwidth]{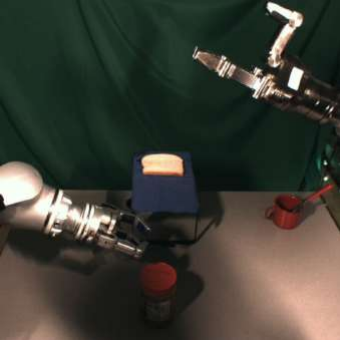}
    \caption{{\Large \textcolor{red}{\textbf{×}}}$\mathtt{RightGripperEmpty()} = \mathtt{\textbf{F}}$}
  \end{subfigure}
\end{figure}

\paragraph{Domain shift} \sys{} relies entirely on semantics to prompt the VLM, abstracting raw states into symbolic states without using visual features. The only requirement for generalizing to novel objects is that the VLM can correctly identify the object referents in the image based on their type information provided in the language prompt. To evaluate this generalization capability, we collected visual observation data (five images per environment) under domain shift by swapping objects with new instances and sampling two configurations. From this observation, we found that the only failure mode introduced by domain shifts occurs when the VLM cannot recognize an object because its visual appearance does not align with the semantics. For example, a plate that is too small might be misclassified as a saucer, leading to incorrect symbolic states.

\begin{table}[H]
\centering
\caption{Per-predicate accuracy of Franka under Domain Shift.}
\resizebox{\textwidth}{!}{%
\begin{tabular}{lcccccc}
\toprule
 & gripper\_empty & holding & mug\_full & plate\_top\_unoccupied & stacked\_on & plate\_is\_dirty \\
\midrule
Accuracy (\%) & 90.0 & 90.0 & 100.0 & 100.0 & 73.3 & 100.0 \\
\bottomrule
\end{tabular}
}
\label{tab:franka_predicate_accuracy_ds}
\end{table}

\begin{figure}[H]
  \centering
  \begin{subfigure}[b]{0.4\linewidth}
    \centering
    \includegraphics[width=\textwidth]{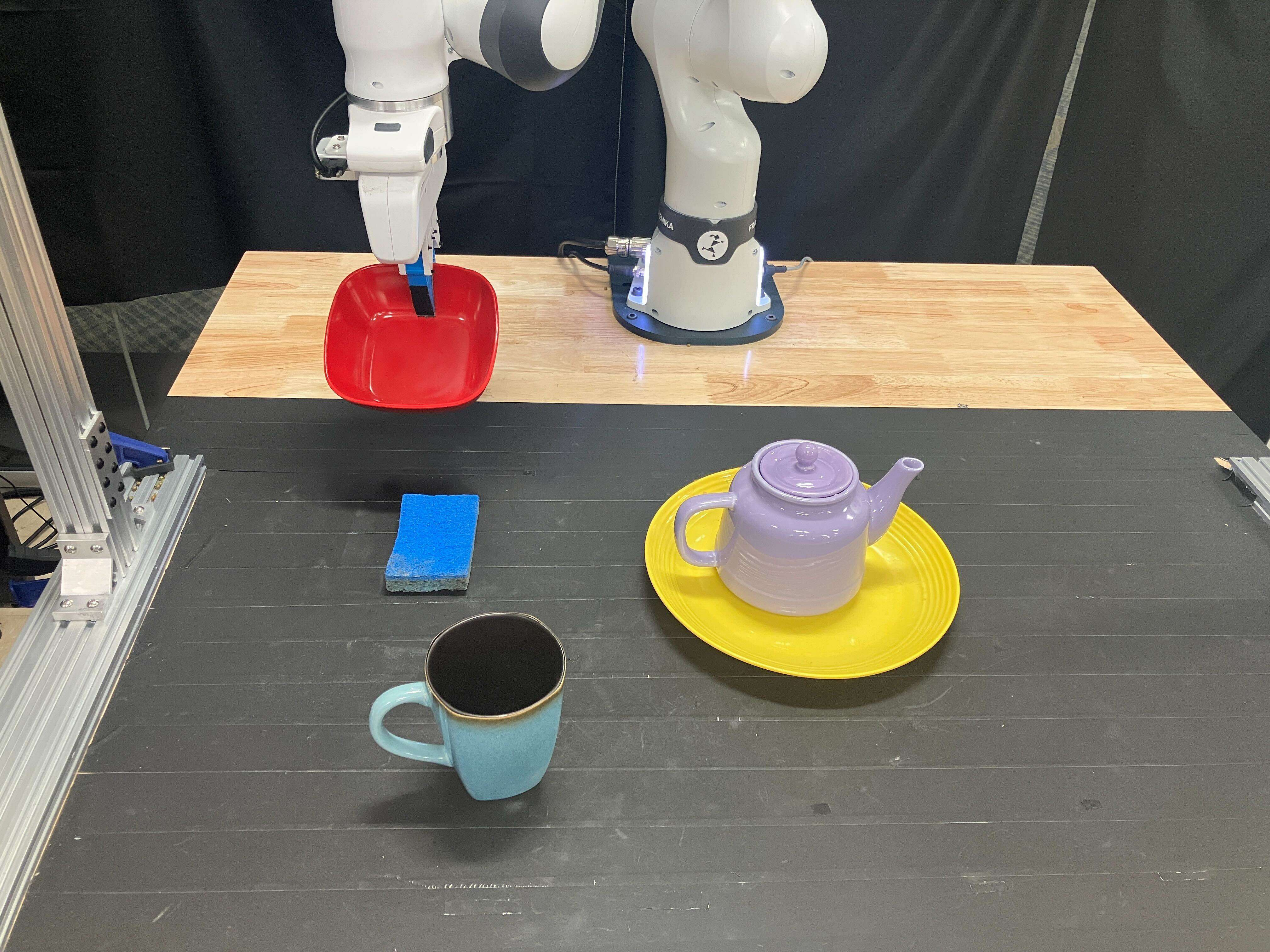}
    \caption{{\large \textcolor{DeepGreen}{\checkmark}}$\mathtt{stacked\_on(Teapot, Plate)} = \mathtt{\textbf{T}}$}
  \end{subfigure}
  \hspace{15pt}
  \begin{subfigure}[b]{0.4\linewidth}
    \centering
    \includegraphics[width=\textwidth]{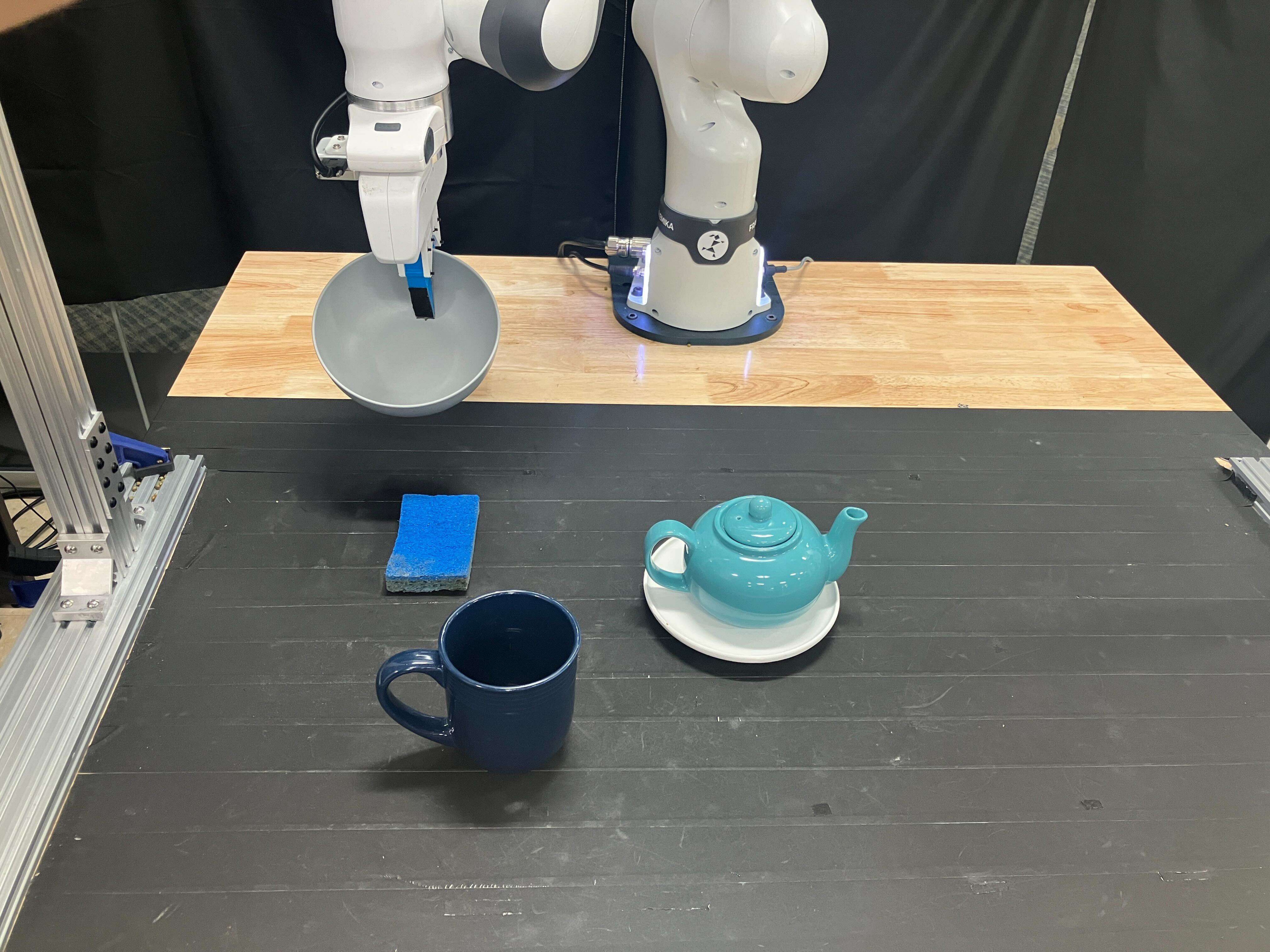}
    \caption{{\Large \textcolor{red}{\textbf{×}}}$\mathtt{stacked\_on(Teapot, Plate)} = \mathtt{\textbf{F}}$}
  \end{subfigure}
\end{figure}

\begin{table}[H]
\centering
\caption{Per-predicate accuracy of Bimanual Kuka under Domain Shift.}
\small 
\resizebox{\textwidth}{!}{%
\begin{tabular}{lcccccc}
\toprule
 & InLeftGripper & InRightGripper & RightGripperEmpty & LeftGripperEmpty & LidOff & InContainer \\
\midrule
Accuracy (\%) & 100.0 & 100.0 & 100.0 & 100.0 & 100.0 & 100.0 \\
\midrule
 & OpenableOnTable & Closed & Coated & SpreadOn & HeldByRobot & UtensilOnTable \\
\midrule
Accuracy (\%) & 100.0 & 100.0 & 100.0 & 100.0 & 100.0 & 100.0 \\
\bottomrule
\end{tabular}
}
\label{tab:dorfl_predicate_accuracy_ds}
\end{table}

\begin{figure}[H]
  \centering
  \begin{subfigure}[b]{0.4\linewidth}
    \centering
    \includegraphics[width=\textwidth]{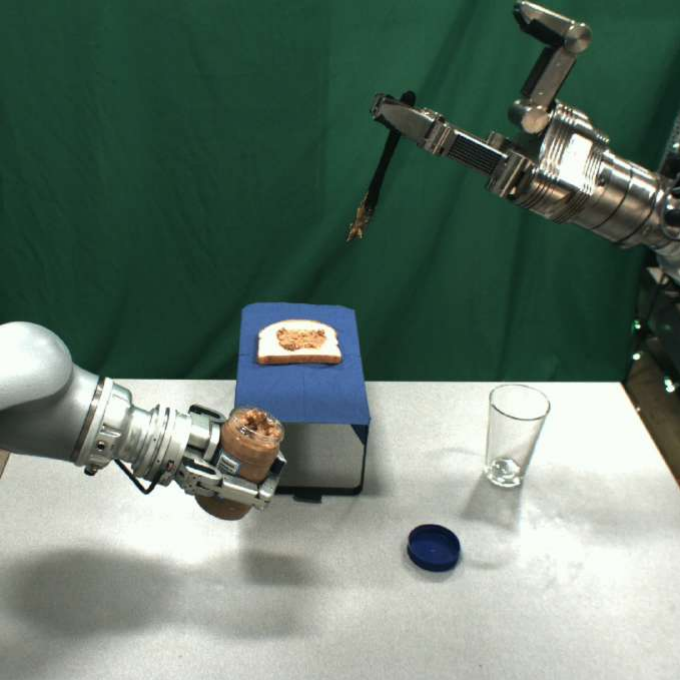}
    \caption{{\large \textcolor{DeepGreen}{\checkmark}}$\mathtt{coated(Knife)} = \mathtt{\textbf{T}}$}
  \end{subfigure}
  \hspace{15pt}
  \begin{subfigure}[b]{0.4\linewidth}
    \centering
    \includegraphics[width=\textwidth]{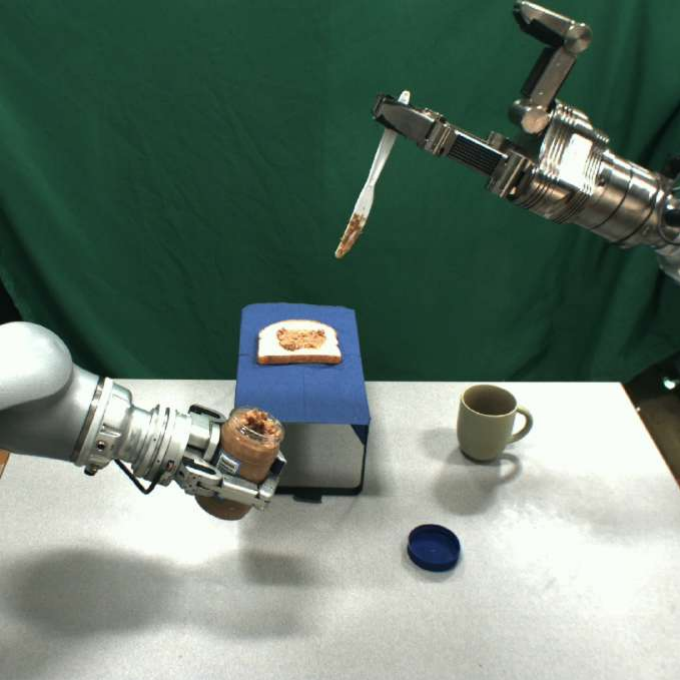}
    \caption{{\large \textcolor{DeepGreen}{\checkmark}}$\mathtt{coated(Knife)} = \mathtt{\textbf{T}}$}
  \end{subfigure}
\end{figure}

\subsection{Other VLMs}
We further examine the possibility of using open-source VLMs as alternatives for \sys{}. We choose Qwen3-VL-235B~\citep{Qwen3-VL} for comparison. To evaluate its capability, we conduct two sets of preliminary experiments: predicate classification and predicate invention.

\paragraph{Predicate Classification} We collected a subset of images (five from Franka and ten from Bimanual Kuka) and evaluated the truth values of each predicate with the two models. From the result, we observed that the two models have different failure patterns, and a prominent one is that Qwen3 can reliably detect if the gripper is holding an object, except for occasional classification errors on the objects being held. In general, we found two models perform on par with each other, and thus we believe they can be used interchangeably for predicate classification.

\begin{table}[H]
\centering
\caption{Per-predicate accuracy of Franka.}
\resizebox{\textwidth}{!}{%
\begin{tabular}{lcccccc}
\toprule
 & gripper\_empty & holding & mug\_full & plate\_top\_unoccupied & stacked\_on & plate\_is\_dirty \\
\midrule
GPT-5 Acc. (\%) & 60.0 & 60.0 & 100.0 & 100.0 & 100.0 & 100.0 \\
Qwen3 Acc. (\%) & 100.0 & 80.0 & 100.0 & 100.0 & 100.0 & 80.0 \\
\bottomrule
\end{tabular}
}
\label{tab:franka_predicate_accuracy_qwen}
\end{table}

\begin{table}[H]
\centering
\caption{Per-predicate accuracy of Bimanual Kuka.}
\small 
\resizebox{\textwidth}{!}{%
\begin{tabular}{lcccccc}
\toprule
 & InLeftGripper & InRightGripper & RightGripperEmpty & LeftGripperEmpty & LidOff & InContainer \\
\midrule
GPT-5 Acc. (\%) & 100.0 & 100.0 & 100.0 & 100.0 & 100.0 & 100.0 \\
Qwen3 Acc. (\%) & 100.0 & 100.0 & 100.0 & 100.0 & 100.0 & 100.0 \\
\midrule
 & OpenableOnTable & Closed & Coated & SpreadOn & HeldByRobot & UtensilOnTable \\
\midrule
GPT-5 Acc. (\%) & 100.0 & 100.0 & 100.0 & 100.0 & 100.0 & 100.0 \\
Qwen3 Acc. (\%) & 100.0 & 100.0 & 90.0 & 80.0 & 100.0 & 100.0 \\
\bottomrule
\end{tabular}
}
\label{tab:dorfl_predicate_accuracy_qwen}
\end{table}

\paragraph{Predicate Invention} We qualitatively compare the performance of both models on inventing predicates by reasoning over contrastive pairs of transitions. For each environment, we curated two contrastive pairs, and each model is prompted by the same input to invent one new predicate. A predicate is considered correct if it is a semantic synonym or antonym of the target predicate. From the result, we can conclude that GPT-5 is much more reliable in reasoning over the transitions for predicate invention, and thus Qwen3 cannot be used as an alternative for this specific task. (We omitted $\mathtt{?\,robot}$ from all predicates' arguments for simplicity.)

\begin{figure}[H]
  \centering
  \captionsetup[subfigure]{justification=centering, font=footnotesize} 
  \captionsetup{justification=centering}
  \begin{subfigure}[H]{0.2\textwidth}
    \centering
    \includegraphics[width=0.95\linewidth]{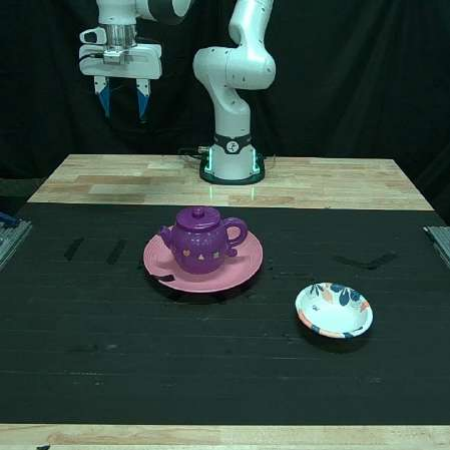}
    \caption{\\{\large \textcolor{DeepGreen}{\checkmark}}$\mathtt{PickUp(Bowl)}$}
  \end{subfigure}%
  \hfill
  \begin{subfigure}[H]{0.2\textwidth}
    \centering
    \includegraphics[width=0.95\linewidth]{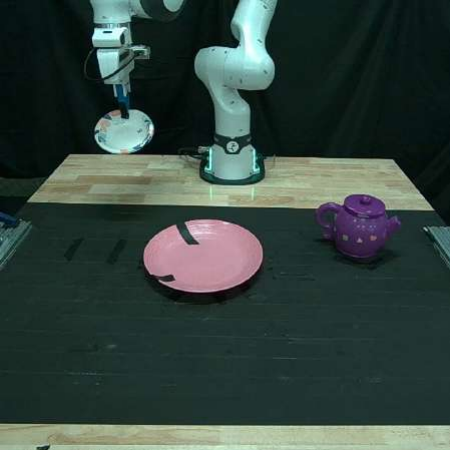}
    \caption{\\{\Large \textcolor{red}{\textbf{×}}}$\mathtt{PickUp(Teapot)}$}
  \end{subfigure}%
  \hfill
  \begin{subfigure}[H]{0.25\textwidth}
    \centering
    \small 
    \begin{tabular}{|p{0.95\linewidth}|}
    \hline
    \textbf{GPT-5} \\ \hline
    {\large \textcolor{DeepGreen}{\checkmark}}
    $\mathtt{gripper\_is\_empty()}$ \\ \hline
    {\large \textcolor{DeepGreen}{\checkmark}}
    $\mathtt{gripper\_is\_empty()}$ \\ \hline
    {\large \textcolor{DeepGreen}{\checkmark}}
    $\mathtt{gripper\_is\_free()}$ \\ \hline
    \end{tabular}
  \end{subfigure}%
  \hfill
  \begin{subfigure}[H]{0.25\textwidth}
    \centering
    \small 
    \begin{tabular}{|p{0.99\linewidth}|}
    \hline
    \textbf{Qwen3} \\ \hline
    {\large \textcolor{DeepGreen}{\checkmark}}
    $\mathtt{gripper\_free()}$ \\ \hline
    {\large \textcolor{DeepGreen}{\checkmark}}
    $\mathtt{gripper\_free()}$ \\ \hline
    {\large \textcolor{red}{\ding{55}}}
    $\mathtt{is\_visible(?pickupable)}$ \\ \hline
    \end{tabular}
  \end{subfigure}
  
  \caption{\textbf{Predicate Invention Case \#1 in Franka.} \textit{Target predicate}: $\mathtt{GripperEmpty()}$
  \\
  \textit{Existing predicates}: $\emptyset$
  }
  
  \label{fig:pred_invent_franka_case_1}
\end{figure}

\begin{figure}[H]
  \centering
  \captionsetup[subfigure]{justification=centering, font=footnotesize} 
  \captionsetup{justification=centering} 
  \begin{subfigure}[H]{0.2\textwidth}
    \centering
    \includegraphics[width=0.95\linewidth]{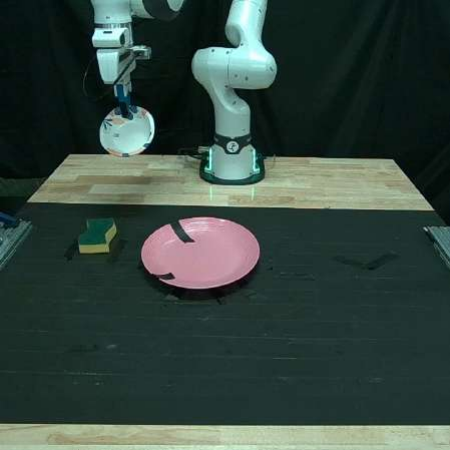}
    \caption{{\large \textcolor{DeepGreen}{\checkmark}}\\$\mathtt{Stack(Bowl, Plate)}$}
  \end{subfigure}%
  \hfill
  \begin{subfigure}[H]{0.2\textwidth}
    \centering
    \includegraphics[width=0.95\linewidth]{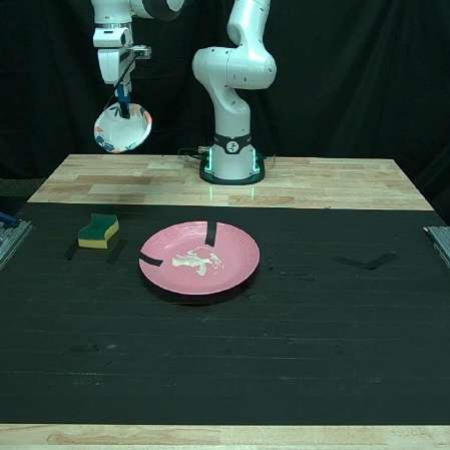}
    \caption{\\{\Large \textcolor{red}{\textbf{×}}}$\mathtt{Stack(Bowl, Plate)}$}
  \end{subfigure}%
  \hfill
  \begin{subfigure}[H]{0.26\textwidth}
    \centering
    \small 
    \begin{tabular}{|p{0.95\linewidth}|}
    \hline
    \textbf{GPT-5} \\ \hline
    {\large \textcolor{DeepGreen}{\checkmark}}
    $\mathtt{plate\_top\_empty(?plate)}$ \\ \hline
    {\large \textcolor{DeepGreen}{\checkmark}}
    $\mathtt{plate\_is\_clean(?plate)}$ \\ \hline
    {\large \textcolor{DeepGreen}{\checkmark}}
    $\mathtt{plate\_is\_clean(?plate)}$ \\ \hline
    \end{tabular}
  \end{subfigure}%
  \hfill
  \begin{subfigure}[H]{0.25\textwidth}
    \centering
    \small 
    \begin{tabular}{|p{0.95\linewidth}|}
    \hline
    \textbf{Qwen3} \\ \hline
    {\large \textcolor{red}{\ding{55}}}
    {\small $\mathtt{stacked\_on}$}\\{\small $\mathtt{(?pickupable, ?plate)}$}\\ \hline
    {\large \textcolor{red}{\ding{55}}}
    {\small $\mathtt{on\_center\_of}$}\\{\small $\mathtt{(?pickupable, ?plate)}$} \\ \hline
    {\large \textcolor{red}{\ding{55}}}
    {\small $\mathtt{is\_fully\_supported}$}\\{\small $\mathtt{(?pickupable, ?plate)}$}\\ \hline
    \end{tabular}
  \end{subfigure}
  
  \caption{\textbf{Predicate Invention Case \#2 in Franka.} \textit{Target predicate}: $\mathtt{PlateIsDirty(?\, plate)}$
  \\
  \textit{Existing predicates}: $\mathtt{GripperEmpty()}, \mathtt{Holding(?\, pickupable)}$
  }
  \label{fig:pred_invent_franka_case_2}
\end{figure}

\begin{figure}[H]
  \centering
  \captionsetup[subfigure]{justification=centering, font=footnotesize} 
  \captionsetup{justification=centering}
  \begin{subfigure}[H]{0.2\textwidth}
    \centering
    \includegraphics[width=0.95\linewidth]{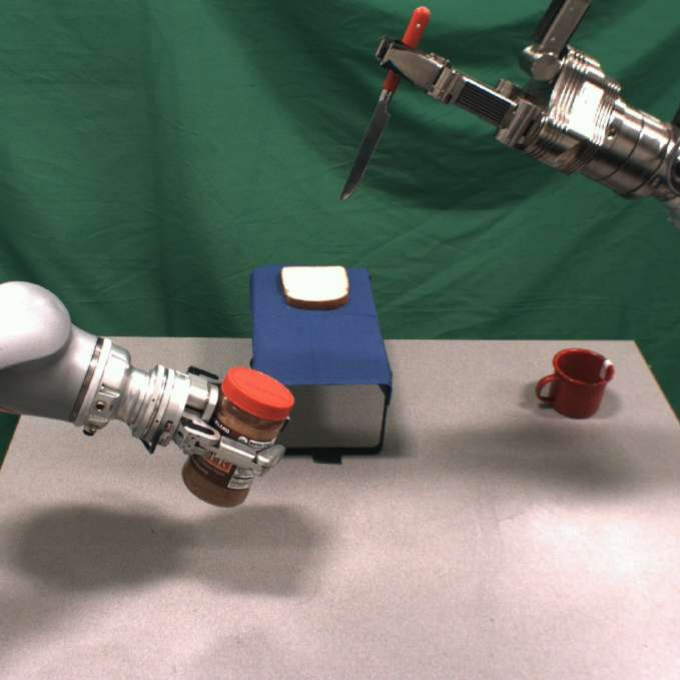}
    \caption{\\{\large \textcolor{DeepGreen}{\checkmark}}$\mathtt{Scoop(Knife, Jar)}$}
  \end{subfigure}%
  \hfill
  \begin{subfigure}[H]{0.2\textwidth}
    \centering
    \includegraphics[width=0.95\linewidth]{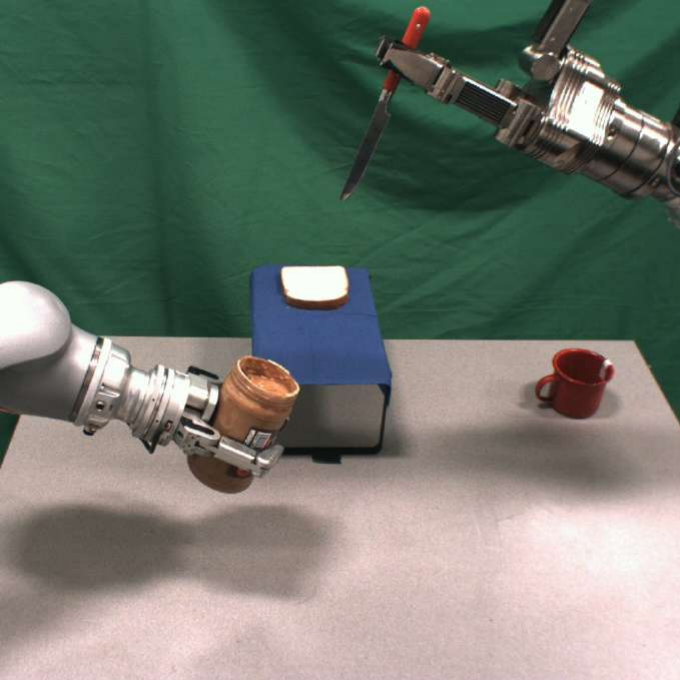}
    \caption{\\{\large \textcolor{red}{\ding{55}}}$\mathtt{Scoop(Knife, Jar)}$}
  \end{subfigure}%
  \hfill
  \begin{subfigure}[H]{0.25\textwidth}
    \centering
    \small 
    \begin{tabular}{|p{0.95\linewidth}|}
    \hline
    \textbf{GPT-5} \\ \hline
    {\large \textcolor{DeepGreen}{\checkmark}} $\mathtt{Open(?openable)}$ \\ \hline
    {\large \textcolor{DeepGreen}{\checkmark}} $\mathtt{Open(?openable)}$ \\ \hline
    {\large \textcolor{DeepGreen}{\checkmark}} $\mathtt{Open(?openable)}$ \\ \hline
    \end{tabular}
  \end{subfigure}%
  \hfill
  \begin{subfigure}[H]{0.25\textwidth}
    \centering
    \small 
    \begin{tabular}{|p{0.95\linewidth}|}
    \hline
    \textbf{Qwen3} \\ \hline
    {\tiny {\large \textcolor{red}{\ding{55}}} $\mathtt{UtensilInOpenable}$}\\{\tiny $\mathtt{ (?utensil, ?openable)}$} \\ \hline
    {\tiny {\large \textcolor{red}{\ding{55}}} $\mathtt{UtensilInOpening}$}\\{\tiny $\mathtt{(?utensil, ?openable)}$} \\ \hline
    {\tiny {\large \textcolor{red}{\ding{55}}} $\mathtt{UtensilInOpenable}$}\\{\tiny $\mathtt{(?utensil, ?openable)}$}\\ \hline
    \end{tabular}
  \end{subfigure}
  
  \caption{\textbf{Predicate Invention Case \#1 in Bi-Kuka.} \textit{Target predicate}: $\mathtt{LidOff(?\,openable)}$
  \\
  \textit{Existing predicates}:$\mathtt{InLeftGripper(?\,openable)}, \mathtt{InRightGripper(?\, utensil)}$
  }
  \label{fig:pred_invent_dorfl_case_1}
\end{figure}

\begin{figure}[H]
  \centering
  \captionsetup[subfigure]{justification=centering, font=footnotesize}
  \captionsetup{justification=centering}
  \begin{subfigure}[H]{0.2\textwidth}
    \centering
    \includegraphics[width=0.95\linewidth]{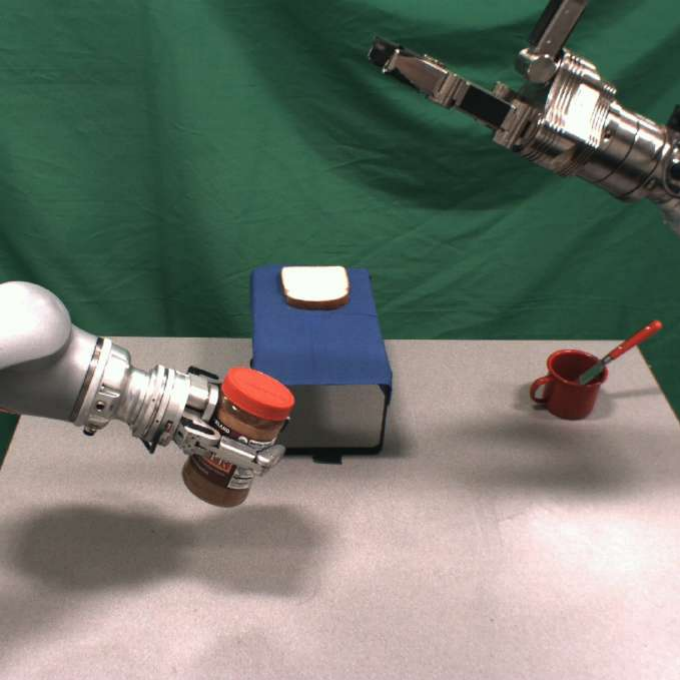}
    \caption{\\
    {\large \textcolor{DeepGreen}{\checkmark}}$\mathtt{Open(Jar)}$}
  \end{subfigure}%
  \hfill
  \begin{subfigure}[H]{0.2\textwidth}
    \centering
    \includegraphics[width=0.95\linewidth]{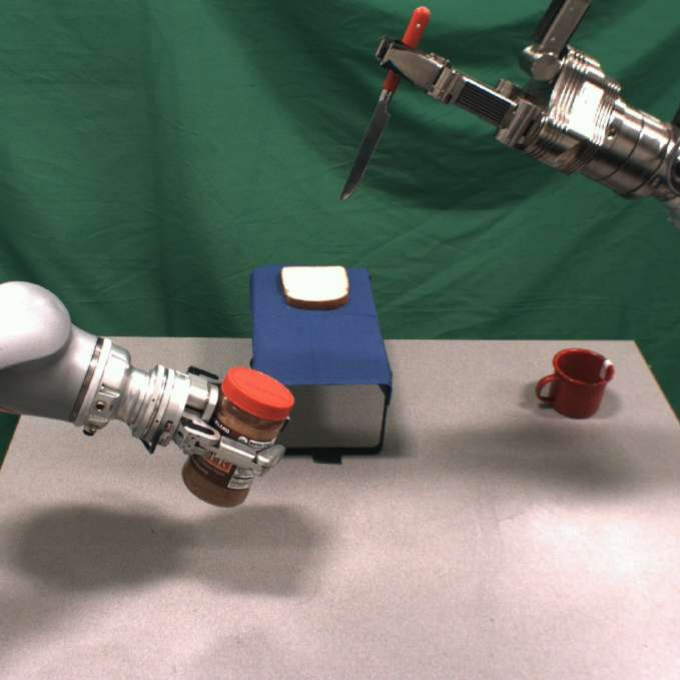}
    \caption{\\{\Large \textcolor{red}{\textbf{×}}}$\mathtt{Open(Jar)}$}
  \end{subfigure}%
  \hfill
  \begin{subfigure}[H]{0.25\textwidth}
    \centering
    \small 
    \begin{tabular}{|p{0.98\linewidth}|}
    \hline
    \textbf{GPT-5} \\ \hline
    {\large \textcolor{DeepGreen}{\checkmark}}$\mathtt{RightHandEmpty()}$ \\ \hline
    {\large \textcolor{DeepGreen}{\checkmark}}$\mathtt{RightHandEmpty()}$ \\ \hline
    {\large \textcolor{red}{\ding{55}}} $\mathtt{LidAttached(? openable)}$ \\ \hline
    \end{tabular}
  \end{subfigure}%
  \hfill
  \begin{subfigure}[H]{0.25\textwidth}
    \centering
    \small 
    \begin{tabular}{|p{0.95\linewidth}|}
    \hline
    \textbf{Qwen3} \\ \hline
    {\large \textcolor{red}{\ding{55}}}
    {\tiny $\mathtt{FullyEnclosedByLeftGripper}$}
    \\{\tiny $\mathtt{(?openable)}$} \\ \hline
    {\large \textcolor{red}{\ding{55}}}
    {\tiny $\mathtt{FullyEnclosedByLeftGripper}$}
    \\{\tiny $\mathtt{(?openable)}$} \\ \hline
    {\large \textcolor{red}{\ding{55}}}
    {\tiny $\mathtt{FullyEnclosedByLeftGripper}$}
    \\{\tiny $\mathtt{(?openable)}$} \\ \hline
    \end{tabular}
  \end{subfigure}
  
  \caption{\textbf{Predicate Invention Case \#2 in Bi-Kuka.} \textit{Target predicate}: $\mathtt{RightGripperEmpty()}$
  \\
  \textit{Existing predicates}:$\mathtt{InLeftGripper(?\,openable)}, \mathtt{LidOff(?\,openable)}$
  }
  \label{fig:pred_invent_dorfl_case_2}
\end{figure}

\clearpage

%% file: sections/appendix/implementation_details.tex
\vspace{-5pt}
\subsection{Planner and Planning Time}
We use K* planner~\citep{katz-lee-socs2023} to generate top $K$ optimal plans, where $K$ in practice is the maximum planning budget. We use an i9-13900F CPU for running all the planning tasks. On average, each planning problem takes $0.0599$ seconds. Specifically, in Robotouille experiments, easy problems take $0.0549$ seconds, hard problems takes $0.0583$ seconds, and impossible problems take $0.0565$ seconds per problem; in Franka experiments, in-domain problems take $0.0529$ seconds, generalization problems take $0.5175$ seconds, and impossible tasks take $0.0516$ seconds per problem; in Bimanual Kuka experiments, all problems take $0.0553$ seconds on average.

\vspace{-5pt}
\subsection{API Call}
For running the experiments, we made approximately 9,300 calls to GPT-5, which cost \$96.59 in total.

\vspace{-5pt}
\subsection{Hyperparameters}
We here report and summarize all hyperparameters of \sys{} used for the experiment to provide better reproducibility. For all experiments, we set the batch size of skill sequence proposal to be $5$ and interaction budget per iteration to be $15$, and we run \sys{} for $5$ iterations for each individual run, except in the Franka generalization experiment, where we use one iteration in each of the three training environments.. For Robotouille experiments, we set the threshold $h$ to be $0.6$. For Franka and bimanual Kuka experiments, we set the threshold $h$ to be $0.5$.

\vspace{-5pt}
\subsection{Experiments}
\label{app:robot-implementation}

\paragraph{Simulated Robotouille}
We conduct experiments in Robotouille~\citep{gonzalez-pumariega2025robotouille}, a simulated kitchen environment with six objects and a robot with five skills:\textit{Pick}, \textit{Place}, \textit{Cut}, \textit{Cook}, and \textit{Stack}.
In the environment, there are several objects: a patty, lettuce, a top bun, and a bottom bun; there is also a cutting board and a stove for cutting the lettuce and cooking the patty. 
We design and categorize 50 skill planning problems for this domain: 20 easy problems, whose optimal solutions have at most 7 steps; 20 hard problems, whose optimal solutions have at most 15 steps; and 10 impossible tasks that cannot be realized in the environment, which a sound planning model must identify.

\paragraph{Single-Arm Manipulation}
We employ a Franka Emika Research 3 robotic arm equipped with a UMI gripper~\citep{UMI}. 
The workspace is observed by a single Intel RealSense D455 exocentric RGB-D camera, oriented to capture both the tabletop scene and the robot. 
The RGB data from this camera are used for learning symbolic models, while the depth information supports object pose estimation. 
Object poses are estimated using FoundationPose~\citep{wen2024foundationpose}, which leverages high-fidelity 3D scanned models of the target objects. 
System-level communication and coordination are implemented in ROS~2 (Humble), which interfaces with motion planning, perception, and control modules. This setup supports five parameterized skills: \emph{Pick}, \emph{Place}, \emph{Stack}, \emph{Pour}, and \emph{Wipe}. 
The first four skills (\emph{Pick}, \emph{Place}, \emph{Stack}, and \emph{Pour}) are executed through motion planning with the MoveIt framework, conditioned on both the end-effector and object poses.
The \emph{Wipe} skill is implemented by replaying a teleoperated trajectory.
The object set $\mathcal{O}$ contains five objects: a mug, a teapot, a plate, a bowl, and a sponge. The robot can pick up and place all objects except the mug and plate, pour ingredients from the teapot into the mug, and use the sponge to wipe the plate if it is dirty. 

\paragraph{Bimanual Manipulation} We use a robot with two horizontally mounted KUKA LBR iiwa 7 R800 manipulators, one with a BarrettHand BH8-282 gripper, and the other with a Schunk Dextrous Hand 2.0 gripper. The robot collects RGB data used for learning symbolic models with a MultiSense S7 camera mounted on a Pan-Tilt unit, while using an Intel RealSense D455 camera for RGB-D data used in pose estimation \citep{wen2024foundationpose} of the objects in the scene.
We use ROS 1 and KUKA FRI to communicate with the robot and utilize the built-in joint impedance control with position target as the low-level controller. At the high level, we create collision models of all objects in the scene and use a task and motion planner to generate motion plans for each skill. In this  setting, the robot is equipped with six skills: \textit{LeftArmPick}, \textit{RightArmPick}, \textit{Open}, \textit{Scoop}, \textit{Spread}, and \textit{Drop}. The \textit{Pick} skills (compatible with knife and peanut butter jar) are implemented using motion planning. The \textit{OpenJar}, \textit{Scoop}, and \textit{Spread} skills are implemented using a combination of motion planning and pre-defined trajectory playback. 
The object set $\mathcal{O}$ consists of three objects: a peanut butter jar, a knife, and a slice of bread. 
The robot can pick up the knife and jar from their initial positions, drop the knife, open the jar, scoop peanut butter with the knife, and spread it on the bread. 
Notably, this environment contains multiple dead ends, which would hinder the data collection process. For example, the knife cannot be picked up again once dropped on tabletop, the jar cannot be released once picked up, and the bread and knife cannot be cleaned once in contact with peanut butter. 

\clearpage

\subsection{Prompts}
\input{sections/appendix/prompts}

%% file: sections/appendix/prompts.tex
In this section, we provide the prompts used for the core components of \sys{} (specifically skill sequence proposal, predicate invention, and predicate evaluation) as well as the ViLA~\citep{hu2023lookleapunveilingpower} baseline. For predicate evaluation (Appendix~\ref{sec:prompt-pred-eval}), we empirically observed that it is more accurate when evaluation is done in batches, where the truth values of multiple predicates are evaluated at once rather than one at a time. In addition, when asking for a fixed and structured output, the accuracy is significantly lower than a free-form output. Therefore, we adopt a two-stage evaluation process: in the first stage, the foundation model generates a response in any format, and in the second stage, it provides a summary of the output from the previous step.

\paragraph{Skill Sequence Proposal} ~\\
\begin{tcolorbox}[
    standard jigsaw,
    title=System Prompt,
    opacityback=0,
]
\PlanCode{
{\color{RedViolet}<AGENT\_DESCRIPTION>} is attempting to learn the preconditions and effects for a finite set of skills by executing exploratory skill sequences and exploring the environment.
}
\end{tcolorbox}

\begin{tcolorbox}[
    standard jigsaw,
    title=Skill Sequence Proposal Prompt,
    opacityback=0,
]
\PlanCode{
Propose a set of skill sequences for a robot to execute.
The robot is attempting to learn the preconditions and effects for a finite set of operators. 
The robot can navigate the environment freely but only has one gripper. 
The robot has access to the following skills with their associated arguments:
\\\\
{\color{RedViolet}[SKILL\_PROMPT]}
\\\\
The list of objects the robot has previously encountered in the environment are: 
\\\\
{\color{RedViolet}[OBJECT\_IN\_SCENE]}
\\
{\color{RedViolet}[ENV\_DESCRIPTION]}
\\
\\
The pairs of consecutive skills (skill1, skill2) that have been least explored are: 
[{\color{RedViolet}[LEAST\_EXPLORED\_SKILLS]}].
Certain skills have similar names and arguments, but different preconditions and effects. 
Using the list of objects and the skill preconditions / effects learned, generate 5 skill sequences and their sequence of skills such that:

(1) the skill sequences should violate their preconditions occasionally.

(2) at least 1 unexplored skill pair is used in each skill sequence.

(3) all skill sequences have at least 15 skills in sequence.

(4) there are no same skills with same arguments consecutively in the sequence.
\\

Output only the sequence of skills to execute, ensuring to follow the naming/syntax/arguments for skills provided. Output 1 skill every new line, following the format below:
\\

Skill Sequence 1: 

GoTo(CounterTop)

PickUp(Apple, CounterTop)
\\ 

Skill Sequence 2:
}
\end{tcolorbox}

\clearpage
\paragraph{Predicate Invention} ~\\

\begin{tcolorbox}[
    standard jigsaw,
    title=Predicate Invention Prompt,
    opacityback=0,
]
\PlanCode{
{\color{RedViolet}[AGENT\_DESCRIPTION]}
\\

The robot has been programmed with the skill {\color{RedViolet}[LIFTED\_SKILL]} two times. In the first execution, the grounded skill {\color{RedViolet}[GROUNDED\_SKILL\_1]} {\color{RedViolet}[SUCCESS\_1]}, and in the second execution, {\color{RedViolet}[GROUNDED\_SKILL\_2]} {\color{RedViolet}[SUCCESS\_2]}. The difference in outcomes suggests that the existing predicate set is insufficient to fully capture the preconditions for successful execution of this skill.
\\
    
Your task is to propose a single new high-level predicate and its semantic meaning based on the visual comparison of the two input images taken before each execution.
\\

Predicates should meet these criteria:
\\
- The predicate must be grounded in visual state only (e.g., "gripper is open," "object is above table," "arm is holding object").
\\
- Describe object state or spatial relations relevant to task success (e.g., gripper open/closed, object on left/right of gripper, object touching/supporting another object, etc.)
\\
- Do not infer properties like affordances (is\_graspable), alignment with grippers, or success likelihood that are vaguely defined and cannot be clearly determined visually. 
\\
- Avoid using concept like grasping zone or robot's reachability to define the predicate since they are not defined by common sense.
\\
- Use at most 2 parameters (e.g., predicate(x), predicate(x, y), predicate()), where robot arm must be included for any robot-environment relation.
\\
- Avoid predicates that assume internal properties like is\_graspable, is\_properly\_aligned, or any accessibility/reachability reasoning that cannot be determined visually.
\\
- The semantic meaning should be a grounded and objective description of the predicate in terms of the physical scene (e.g., "the object is fully enclosed by the robot's gripper"), not about execution success or skill dynamics.
\\
- The parameters of the predicate must be subset of the parameters of the skill.
\\\\
Format your output as follows:
\\
`predicate\_name(parameters)`: semantic\_meaning.
\\

for example:

`CloseTo(arm, location)`: the robot arm is close to the location.
\\

Current predicates: 
{\color{RedViolet}[PRED\_LIST]}
\\

Previously proposed but rejected predicates: 
{\color{RedViolet}[TRIED\_PRED]}
\\

Avoid duplicates or near-duplicates of existing predicates and rejected predicates. Reason over using a paragraph and generate the predicate and the semantic meaning in the given format in a separate line.
\\

One new predicate candidate for improving the representation of the precondition for {\color{RedViolet}[LIFTED\_SKILL]} ( Don't use any parameter other than {\color{RedViolet}[PARAMETERS]}):
}
\end{tcolorbox}

\clearpage
\paragraph{Predicate Evaluation} ~\\
\label{sec:prompt-pred-eval}

\begin{tcolorbox}[
    standard jigsaw,
    title=Predicate Evaluation Prompt: Step 1,
    opacityback=0,
]
\PlanCode{
Given the current observation of the simulated kitchen domain, the object types, and the list of predicates, what are the true grounded predicates? 
\\

{\color{RedViolet}[ENVIRONMENTAL\_DESCRIPTION]}
\\

Objects:

{\color{RedViolet}[OBJECTS]}
\\

Predicates: ~\\

{\color{RedViolet}[PREDICATES]}
}
\end{tcolorbox}

\begin{tcolorbox}[
    standard jigsaw,
    title=Predicate Evaluation Prompt: Step 2,
    opacityback=0,
]
\PlanCode{
Summarize what the true grounded predicates are from this response, and list them in the format of predicate\_name(arg1, arg2, ...) in separate lines with no any formatting. If the response contains typos of object names or redundant indices, you should correct them. Correct object names are: {\color{RedViolet}[OBJECT\_NAMES]}. If the response include redundant predicates that are not in this list, you should filter them. Correct predicates are: {\color{RedViolet}[PRED\_NAMES]}. The response is:
\\

"""

{\color{RedViolet}[RESPONSE]}

""
}
\end{tcolorbox}

\paragraph{ViLa} ~\\

\begin{tcolorbox}[
    standard jigsaw,
    title=ViLA Prompt,
    opacityback=0,
]
\PlanCode{
You are {\color{RedViolet}[AGENT\_DESCRIPTION]}. As a robot, you are able to execute the following skills:

{\color{RedViolet}[SKILLS]}
\\

Here are the objects and their types that are compatible with your skills:

{\color{RedViolet}[OBJECTS]}
\\

You are given two images: The first one captures your current observation, and the second one specifies your goal. Given both images, your job is to generate a plan starting from the *current state* to the goal state. You should first reason about the goal of the task and how the skills can be chained to solve it in the first paragraph. After the reasoning, return the plan from the current state in a new paragraph by listing skills in separate lines with no additional explanation, header, or numbering.

Use "Done" in the skill list to indicate the task is complete, and report if the task is impossible to solve by simply returning "Impossible".
}
\end{tcolorbox}
\clearpage

%% file: sections/appendix/full_related_work.tex
\sys{} draws ideas from different fields of research, such as model learning, abstraction learning, and task and motion planning (TAMP). 
Other methods have used FMs as high-level planners~\citep{hu2023lookleapunveilingpower}.
Concurrent work has also explored how representations can be learned directly from pixels~\citep{athalye2026pix2pred}.
Multiple approaches~\citep{han2024interpret, liang2025visualpredicator} have used foundation models to learn symbolic representations of robot skills, but require extensive feedback or prior knowledge from human experts. 
To the best of our knowledge, our work is the first to use formally characterized planning model properties to guide generative predicate invention.

TAMP has long been used to solve complex robot tasks~\citep{shah2020anytime,garrett2021integrated}. 
However, these approaches require symbolic models of robot skills for task planning. Various approaches have been developed to learn such symbolic models compatible with TAMP solvers from high-dimensional inputs~\citep{Konidaris2018FromST,silver2023predinvent,shah2024reals}.
Additionally, the abstract representations learned through these methods are not directly human-interpretable. In contrast, we explicitly design our approach to work with high-dimensional inputs and generate human-interpretable abstractions using pre-trained FMs. 

\paragraph{Abstraction Learning for Robotic Skills}
There has been a long track of works focusing on building hierarchies that abstract away high-dimensional details with low-dimensional abstractions for planning~\citep{konidaris2009skill,Konidaris2018FromST,shah2024reals}, and those applied to robotics are usually connected to TAMP~\citep{shah2020anytime,garrett2021integrated}. These approaches, however, are incapable of handling high-dimensional sensory-motor signals (such as images) as input. Research on action model learning~\citep{xi2024neuro,juba2021safe} learn symbolic action models for input skills. However, unlike our method, these approaches require symbols to be provided as input. 
Similar to our system's integration of self-play and focus on uncovering skill conditions, \citet{verma2022discovering} focus on assessing capabilities of black-box agents for grid world-like tasks while assuming that the agent is an oracle. 
A tangential research effort on chaining various skills in novel environments involves training extra models~\citep{yokoyama2023ascadaptiveskillcoordination} and STRIPS task planner with action primitives~\citep{gu2022multiskillmobilemanipulationobject, szot2022habitat20traininghome}.

\paragraph{Predicate Learning for Robotic Tasks}
Predicates provide a convenient way to abstract away low-level details of the environment and build efficient and compact representations.
Prior to foundation models, previous attempts to build classifiers for predicates from raw image inputs originated from the neuro-symbolic domain~\citep{johnson2016clevrdiagnosticdatasetcompositional, mao2019neurosymbolicconceptlearnerinterpreting}, and their initial application for robotics took a similar supervised learning approach with labeled demonstrations~\citep{migimatsu2022grounding} or generated tasks~\citep{Lamanna2023PlanningFL}. 
After the emergence of foundation models, recent works guide skill learning with predicates generated by LLM or together with human interaction. \citet{li2024league} invents symbolic skills for reward functions used for RL training but cannot generalize to skills learned through latent objectives, which is more commonly seen in imitation learning. \citet{Li2023EmbodiedAL} and \citet{han2024interpret} leverage human experts to provide feedback to the LLM to help it improve the learned predicates and skills. 

\paragraph{Task Generation for Robotics}
The approach of automatically proposing tasks has been studied for active learning and curriculum learning in grid worlds and games~\citep{wang2019paired, jiang2021prioritizedlevelreplay} to robotic domains~\citep{fang2021adaptiveproceduraltaskgeneration,Fang2022ActiveTR}. \cite{Lamanna2023PlanningFL} generates tasks in PDDL as training sets to learn classifiers for object properties in predicates format, while they assume the action operators are given. With the commonsense reasoning ability of foundation models, recent works have applied the idea of automatic task proposing and self-playing for exploration~\citep{google2024pivot,exploreeqa2024}, data collection~\citep{wang2023robogen,yang2023octopus,ahn2024autort}, boosting skills learning~\citep{Ha2023ScalingUA,wang2023voyageropenendedembodiedagent}, and scene understanding~\citep{jiang2024roboexp}. These works indicate a promising direction for generating robotic data and scaling up. Following the idea, we equipped our system with a task-proposing module for generating skill sequences specific to skills and predicates, which serves the idea of both data collection and exploration.

\paragraph{Embodied Reasoning with Foundation Models} There has been a track of work on leveraging large language models (LLMs) for embodied decision-making~\citep{huang2022languagemodelszeroshotplanners,raman2024cape} and reasoning~\citep{huang2022inner}, while vision-language models (VLMs) are often considered to have limited embodied reasoning ability due to their pre-training corpora that focus primarily on language generation~\citep{valmeekam2023llm}. Common ways of addressing this issue include fine-tuning on datasets from a specific domains~\citep{hong20233dllm,mu2023embodiedgptvisionlanguagepretrainingembodied,chen2024spatialvlm} or knowledge distillation~\citep{sumers2023distillinginternetscalevisionlanguagemodels, yang2023octopus}. Meanwhile, many works manage to leverage preexisting models without further training from direct visual observation~\citep{liu2024moka, google2024pivot} to complete robotic tasks~\citep{jiang2024roboexp}. In these works, the embodied reasoning ability of the foundation models serves as the central part of the systems. However, most benchmarking works evaluate the embodied reasoning ability of the models in a question-answering fashion~\citep{sermanet2023robovqamultimodallonghorizonreasoning, OpenEQA2023,cheng2024egothink,chen2025visionlanguage}, where it remains unclear whether they are capable of solving robotic tasks. 

%% file: sections/appendix/glossary.tex

We summarize the notation used in the main paper. Modifiers apply throughout;
grounded, indexed, and estimated variants are not listed separately unless
they require an additional definition.
\begingroup
\small
\setlength{\tabcolsep}{4pt}
\renewcommand{\arraystretch}{1.08}
\setlength{\LTpre}{5pt}
\setlength{\LTpost}{6pt}
\begin{longtable}{@{}>{\raggedright\arraybackslash}p{0.28\linewidth}>{\raggedright\arraybackslash}p{\dimexpr0.72\linewidth-8pt\relax}@{}}
\toprule
\textbf{Notation} & \textbf{Meaning} \\
\midrule
\endfirsthead
\toprule
\textbf{Notation} & \textbf{Meaning (continued)} \\
\midrule
\endhead
\bottomrule
\endfoot
\multicolumn{2}{@{}l}{\textbf{Modifiers}} \\*
\addlinespace[2pt]
$\underline{x}$ & Grounded instance; underlining denotes its valid groundings. \\
$\boldsymbol{x}$ & Ordered tuple or sequence, including object arguments, parameters, and plans. \\
$x^{\uparrow}$ & Symbolic abstraction of a state or state space. In result tables. \\
$\widehat{x}$ & Hypothetical or predicted quantity, as specified locally. \\
$x^{\ast}$ & The target model or target success ratio. \\
$x^{\prime}$ & Successor state. \\
$x^{+},x^{-}$ & Add and delete components of an effect. \\*
$x_{\omega},x_a,x_p$ & Association with a skill, operator, or predicate. \\
\addlinespace[5pt]
\multicolumn{2}{@{}l}{\textbf{Environment and skills}} \\*
\addlinespace[2pt]
$\mathcal{S},s$ & Low-level state space and a state. \\
$\mathsf{T},\mathsf{t}(\cdot)$ & Object-type set and typing function: $\mathsf{t}(o)$ gives an object\textquotesingle s types; $\mathsf{t}(\theta^i)$ gives a parameter\textquotesingle s required type. \\
$\mathcal{O},o$ & Objects in a scene and an individual object. \\
$\Omega,\omega$ & Library of black-box skill schemas and an individual skill. \\
$\boldsymbol{\theta}$ & Typed parameter tuple of a skill, predicate, or operator. \\
$\pi$ & Low-level policy of a skill. \\
$\mathcal{I}_{\omega}$ & Pre-success set: states from which the skill succeeds, parameterized by object arguments. \\
$\beta_{\omega}$ & Success set: states satisfying the skill\textquotesingle s successful outcome, parameterized by object arguments. \\
$T$ & Deterministic low-level transition function. \\
$s_0$ & The initial state in a skill planning problem.\\
$\mathcal{S}_g$ & Set of goal states in a skill planning problem. \\*
$N$ & Number of scenes in the learning problem. \\
\addlinespace[5pt]
\multicolumn{2}{@{}l}{\textbf{Symbolic model}} \\*
\addlinespace[2pt]
$\mathcal{P},p$ & Predicate vocabulary and a predicate schema. \\
$\ell_p,C_p$ & Predicate\textquotesingle s semantic label and classifier. \\
$\alpha$ & Abstraction function mapping a low-level state to its true grounded predicates. \\
$\Gamma$ & Grounding function mapping a symbolic condition to the low-level states satisfying it. \\
$\mathcal{A},a$ & Set of operator schemas and an individual operator. \\
$\mathrm{Pre},\mathrm{Eff}$ & Operator preconditions and effects; effects have add and delete components. \\
$\mathcal{M}$ & Abstract transition model consisting of predicates and operators: $\langle\mathcal{P},\mathcal{A}\rangle$. \\
$\Phi_{\omega},\Xi_{\omega}$ & Model-estimated pre-success and success sets, respectively; instance-specific sets use a grounded skill subscript. \\
$\Lambda_{\boldsymbol{o}}$ & Lifting relative to skill arguments: retain relevant predicates and replace objects with typed variables. \\*
$\Delta(s,s^{\prime})$ & Abstract change: the pair of added and deleted grounded predicates between two states. \\
\addlinespace[5pt]
\multicolumn{2}{@{}l}{\textbf{Data collection}} \\*
\addlinespace[2pt]
$\tau$ & Transition $\langle s,\underline{\omega},s^{\prime}\rangle$ recording a skill execution. \\
$\mathcal{D}$ & Collected transition dataset; $D$ denotes its size and $d\subseteq\mathcal{D}_{\omega}$ an effect cluster. \\
$b,c$ & Balance and coverage scores for a proposed skill sequence. \\
$r_{\mathrm{suc}}$ & Success ratio; a hat denotes its prediction and a star its target. \\
$Q$ & Count matrix of consecutive skill pairs; $Q[i,j]$ counts pair $(i,j)$. \\
$p_{i,j},H(Q)$ & Normalized frequency of skill pair $(i,j)$ and entropy of the skill-pair distribution. \\
\addlinespace[5pt]
\multicolumn{2}{@{}l}{\textbf{Theoretical analysis}} \\*
\addlinespace[2pt]
$\mathcal{H}$ & Fixed finite hypothesis class of candidate symbolic models. \\
$\ell_{\mathcal{M}}(\tau)$ & Indicator of disagreement between predicted and true initial or terminal set membership. \\
$d_{\mathrm{compl}}$ & Expected error averaged over transition positions within an episode. \\
$d_{\mathrm{compl},K}$ & Average conditional error across $K$ episodes, conditioning each episode on its preceding history. \\
$n,K$ & Transitions per episode and number of collection episodes, respectively. \\
$r,t$ & Episode and within-episode position indices; $\tau_{r,t}$ identifies a transition. \\
$\mathcal{D}_{1:r-1}$ & Collection history preceding episode $r$. \\
$k,m$ & Separation lag and effective sample size $m=\lfloor n/k\rfloor$. \\
$\beta(k)$ & Bound on temporal dependence at lag $k$, distinct from the skill success set $\beta_{\omega}$. \\
$\epsilon$ & Allowed expected-error tolerance. \\*
$\ominus$ & Symmetric difference of sets. \\
\end{longtable}
\endgroup

%% file: main.bbl
\begin{thebibliography}{66}
\providecommand{\natexlab}[1]{#1}
\providecommand{\url}[1]{\texttt{#1}}
\expandafter\ifx\csname urlstyle\endcsname\relax
  \providecommand{\doi}[1]{doi: #1}\else
  \providecommand{\doi}{doi: \begingroup \urlstyle{rm}\Url}\fi

\bibitem[Ahn et~al.(2022)Ahn, Brohan, Brown, Chebotar, Cortes, David, Finn, et~al.]{saycan_corl}
Michael Ahn, Anthony Brohan, Noah Brown, Yevgen Chebotar, Omar Cortes, Byron David, Chelsea Finn, et~al.
\newblock {Do As I Can, Not As I Say: Grounding Language in Robotic Affordances}.
\newblock In \emph{Proceedings of the 6th Conference on Robot Learning}, 2022.

\bibitem[Ahn et~al.(2024)Ahn, Dwibedi, Finn, Arenas, Gopalakrishnan, Hausman, Ichter, et~al.]{ahn2024autort}
Michael Ahn, Debidatta Dwibedi, Chelsea Finn, Montserrat~Gonzalez Arenas, Keerthana Gopalakrishnan, Karol Hausman, Brian Ichter, et~al.
\newblock {Auto{RT}: Embodied Foundation Models for Large Scale Orchestration of Robotic Agents}.
\newblock In \emph{First Workshop on Vision-Language Models for Navigation and Manipulation at ICRA 2024}, 2024.

\bibitem[Arora et~al.(2018)Arora, Fiorino, Pellier, Métivier, and Pesty]{arora2018review}
Ankuj Arora, Humbert Fiorino, Damien Pellier, Marc Métivier, and Sylvie Pesty.
\newblock {A Review of Learning Planning Action Models}.
\newblock \emph{The Knowledge Engineering Review}, 33:\penalty0 e20, 2018.

\bibitem[Athalye et~al.(2026)Athalye, Kumar, Silver, Liang, Wang, Lozano-Pérez, and Kaelbling]{athalye2026pix2pred}
Ashay Athalye, Nishanth Kumar, Tom Silver, Yichao Liang, Jiuguang Wang, Tomás Lozano-Pérez, and Leslie~Pack Kaelbling.
\newblock {From Pixels to Predicates: Learning Symbolic World Models via Pretrained VLMs}.
\newblock \emph{IEEE Robotics and Automation Letters (RA-L)}, 11\penalty0 (4):\penalty0 4002--4009, 2026.

\bibitem[Bai et~al.(2025)Bai, Cai, Chen, Chen, Chen, Cheng, Deng, Ding, Gao, Ge, Ge, Guo, Huang, Huang, Huang, Hui, Jiang, Li, Li, Li, Li, Lin, Lin, Liu, Liu, Liu, Liu, Liu, Liu, Lu, Luo, Lv, Men, Meng, Ren, Ren, Song, Sun, Tang, Tu, Wan, Wang, Wang, Wang, Wang, Xie, Xu, Xu, Xu, Yang, Yang, Yang, Yang, Yu, Zhang, Zhang, Zhang, Zheng, Zhong, Zhou, Zhou, Zhou, Zhu, and Zhu]{Qwen3-VL}
Shuai Bai, Yuxuan Cai, Ruizhe Chen, Keqin Chen, Xionghui Chen, Zesen Cheng, Lianghao Deng, Wei Ding, Chang Gao, Chunjiang Ge, Wenbin Ge, Zhifang Guo, Qidong Huang, Jie Huang, Fei Huang, Binyuan Hui, Shutong Jiang, Zhaohai Li, Mingsheng Li, Mei Li, Kaixin Li, Zicheng Lin, Junyang Lin, Xuejing Liu, Jiawei Liu, Chenglong Liu, Yang Liu, Dayiheng Liu, Shixuan Liu, Dunjie Lu, Ruilin Luo, Chenxu Lv, Rui Men, Lingchen Meng, Xuancheng Ren, Xingzhang Ren, Sibo Song, Yuchong Sun, Jun Tang, Jianhong Tu, Jianqiang Wan, Peng Wang, Pengfei Wang, Qiuyue Wang, Yuxuan Wang, Tianbao Xie, Yiheng Xu, Haiyang Xu, Jin Xu, Zhibo Yang, Mingkun Yang, Jianxin Yang, An~Yang, Bowen Yu, Fei Zhang, Hang Zhang, Xi~Zhang, Bo~Zheng, Humen Zhong, Jingren Zhou, Fan Zhou, Jing Zhou, Yuanzhi Zhu, and Ke~Zhu.
\newblock Qwen3-vl technical report.
\newblock \emph{arXiv preprint arXiv:2511.21631}, 2025.

\bibitem[Chen et~al.(2024)Chen, Xu, Kirmani, Ichter, Sadigh, Guibas, and Xia]{chen2024spatialvlm}
Boyuan Chen, Zhuo Xu, Sean Kirmani, Brian Ichter, Dorsa Sadigh, Leonidas Guibas, and Fei Xia.
\newblock {SpatialVLM: Endowing Vision-Language Models with Spatial Reasoning Capabilities}.
\newblock In \emph{Proceedings of the IEEE/CVF Conference on Computer Vision and Pattern Recognition}, 2024.

\bibitem[Chen et~al.(2025)Chen, Mees, Kumar, and Levine]{chen2025visionlanguage}
William Chen, Oier Mees, Aviral Kumar, and Sergey Levine.
\newblock {Vision-Language Models Provide Promptable Representations for Reinforcement Learning}.
\newblock \emph{Transactions on Machine Learning Research (TMLR)}, 2025.

\bibitem[Cheng et~al.(2024)Cheng, Guo, Wu, Fang, Li, Liu, and Liu]{cheng2024egothink}
Sijie Cheng, Zhicheng Guo, Jingwen Wu, Kechen Fang, Peng Li, Huaping Liu, and Yang Liu.
\newblock {EgoThink: Evaluating First-Person Perspective Thinking Capability of Vision-Language Models}.
\newblock In \emph{Proceedings of the 2024 IEEE/CVF Conference on Computer Vision and Pattern Recognition}, 2024.

\bibitem[Chi et~al.(2024)Chi, Xu, Pan, Cousineau, Burchfiel, Feng, Tedrake, and Song]{UMI}
Cheng Chi, Zhenjia Xu, Chuer Pan, Eric Cousineau, Benjamin Burchfiel, Siyuan Feng, Russ Tedrake, and Shuran Song.
\newblock {Universal Manipulation Interface: In-The-Wild Robot Teaching Without In-The-Wild Robots}.
\newblock In \emph{Proceedings of Robotics: Science and Systems}, 2024.

\bibitem[Dantam et~al.(2018)Dantam, Kingston, Chaudhuri, and Kavraki]{dantam2018incremental}
Neil~T Dantam, Zachary~K Kingston, Swarat Chaudhuri, and Lydia~E Kavraki.
\newblock An incremental constraint-based framework for task and motion planning.
\newblock \emph{The International Journal of Robotics Research}, 37\penalty0 (10):\penalty0 1134--1151, 2018.

\bibitem[Doncieux et~al.(2018)Doncieux, Filliat, D{\'\i}az-Rodr{\'\i}guez, Hospedales, Duro, Coninx, Roijers, Girard, Perrin, and Sigaud]{Doncieux18}
S.~Doncieux, D.~Filliat, N.~D{\'\i}az-Rodr{\'\i}guez, T.~Hospedales, R.~Duro, A.~Coninx, D.M. Roijers, B.~Girard, N.~Perrin, and O.~Sigaud.
\newblock Open-ended learning: a conceptual framework based on representational redescription.
\newblock \emph{Frontiers in Neurorobotics}, 2018.

\bibitem[Driess et~al.(2023)Driess, Xia, Sajjadi, Lynch, Chowdhery, Ichter, Wahid, Tompson, Vuong, Yu, Huang, Chebotar, Sermanet, Duckworth, Levine, Vanhoucke, Hausman, Toussaint, Greff, Zeng, Mordatch, and Florence]{driess2023palme}
Danny Driess, Fei Xia, Mehdi S.~M. Sajjadi, Corey Lynch, Aakanksha Chowdhery, Brian Ichter, Ayzaan Wahid, Jonathan Tompson, Quan Vuong, Tianhe Yu, Wenlong Huang, Yevgen Chebotar, Pierre Sermanet, Daniel Duckworth, Sergey Levine, Vincent Vanhoucke, Karol Hausman, Marc Toussaint, Klaus Greff, Andy Zeng, Igor Mordatch, and Pete Florence.
\newblock {PaLM-E: An Embodied Multimodal Language Model}.
\newblock In \emph{Proceedings of the 40th International Conference on Machine Learning}, 2023.

\bibitem[Fang et~al.(2021)Fang, Zhu, Savarese, and Fei-Fei]{fang2021adaptiveproceduraltaskgeneration}
Kuan Fang, Yuke Zhu, Silvio Savarese, and Li~Fei-Fei.
\newblock {Adaptive Procedural Task Generation for Hard-Exploration Problems}.
\newblock In \emph{Proceedings of the 9th International Conference on Learning Representations}, 2021.

\bibitem[Fang et~al.(2022)Fang, Migimatsu, Mandlekar, Fei-Fei, and Bohg]{Fang2022ActiveTR}
Kuan Fang, Toki Migimatsu, Ajay Mandlekar, Li~Fei-Fei, and Jeannette Bohg.
\newblock {Active Task Randomization: Learning Robust Skills via Unsupervised Generation of Diverse and Feasible Tasks}.
\newblock \emph{Proceedings of the 2023 IEEE/RSJ International Conference on Intelligent Robots and Systems}, 2022.

\bibitem[Fang et~al.(2024)Fang, Liu, Abbeel, and Levine]{liu2024moka}
Kuan Fang, Fangchen Liu, Pieter Abbeel, and Sergey Levine.
\newblock {MOKA: Open-World Robotic Manipulation through Mark-Based Visual Prompting}.
\newblock \emph{Proceedings of Robotics: Science and Systems}, 2024.

\bibitem[Garrett et~al.(2021)Garrett, Chitnis, Holladay, Kim, Silver, Kaelbling, and Lozano-P{\'e}rez]{garrett2021integrated}
Caelan~Reed Garrett, Rohan Chitnis, Rachel Holladay, Beomjoon Kim, Tom Silver, Leslie~Pack Kaelbling, and Tom{\'a}s Lozano-P{\'e}rez.
\newblock {Integrated Task and Motion Planning}.
\newblock \emph{Annual Review of Control, Robotics, and Autonomous Systems}, 4:\penalty0 265--293, 2021.

\bibitem[Ghallab et~al.(2016)Ghallab, Nau, and Traverso]{ghallab_nau_traverso_2016}
Malik Ghallab, Dana Nau, and Paolo Traverso.
\newblock \emph{Automated Planning and Acting}.
\newblock Cambridge University Press, 2016.
\newblock \doi{10.1017/CBO9781139583923}.

\bibitem[Gonzalez-Pumariega et~al.(2025)Gonzalez-Pumariega, Yean, Sunkara, and Choudhury]{gonzalez-pumariega2025robotouille}
Gonzalo Gonzalez-Pumariega, Leong~Su Yean, Neha Sunkara, and Sanjiban Choudhury.
\newblock {Robotouille: An Asynchronous Planning Benchmark for {LLM} Agents}.
\newblock In \emph{Proceedings of the 13th International Conference on Learning Representations}, 2025.

\bibitem[Gu et~al.(2022)Gu, Chaplot, Su, and Malik]{gu2022multiskillmobilemanipulationobject}
Jiayuan Gu, Devendra~Singh Chaplot, Hao Su, and Jitendra Malik.
\newblock {Multi-skill Mobile Manipulation for Object Rearrangement}.
\newblock In \emph{Proceedings of the 11th International Conference on Learning Representations}, 2022.

\bibitem[Ha et~al.(2023)Ha, Florence, and Song]{Ha2023ScalingUA}
Huy Ha, Pete Florence, and Shuran Song.
\newblock {Scaling Up and Distilling Down: Language-Guided Robot Skill Acquisition}.
\newblock In \emph{Proceedings of the 7th Conference on Robot Learning}, 2023.

\bibitem[Han et~al.(2024)Han, Zhu, Zhu, Wu, and Zhu]{han2024interpret}
Muzhi Han, Yifeng Zhu, Song-Chun Zhu, Ying~Nian Wu, and Yuke Zhu.
\newblock {InterPreT: Interactive Predicate Learning from Language Feedback for Generalizable Task Planning}.
\newblock In \emph{Proceedings of Robotics: Science and Systems}, 2024.

\bibitem[Hong et~al.(2023)Hong, Zhen, Chen, Zheng, Du, Chen, and Gan]{hong20233dllm}
Yining Hong, Haoyu Zhen, Peihao Chen, Shuhong Zheng, Yilun Du, Zhenfang Chen, and Chuang Gan.
\newblock {3D-LLM: Injecting the 3D World into Large Language Models}.
\newblock In \emph{Advances in Neural Information Processing Systems (NeurIPS)}, 2023.

\bibitem[Hu et~al.(2023)Hu, Lin, Zhang, Yi, and Gao]{hu2023lookleapunveilingpower}
Yingdong Hu, Fanqi Lin, Tong Zhang, Li~Yi, and Yang Gao.
\newblock {Look Before You Leap: Unveiling the Power of GPT-4V in Robotic Vision-Language Planning}.
\newblock \emph{arXiv preprint arXiv:2311.17842}, 2023.

\bibitem[Huang et~al.(2022)Huang, Abbeel, Pathak, and Mordatch]{huang2022languagemodelszeroshotplanners}
Wenlong Huang, Pieter Abbeel, Deepak Pathak, and Igor Mordatch.
\newblock {Language Models as Zero-Shot Planners: Extracting Actionable Knowledge for Embodied Agents}.
\newblock In \emph{Proceedings of the 39th International Conference on Machine Learning}, 2022.

\bibitem[Huang et~al.(2023)Huang, Xia, Xiao, Chan, Liang, Florence, Zeng, Tompson, Mordatch, Chebotar, Sermanet, Jackson, Brown, Luu, Levine, Hausman, and Ichter]{huang2022inner}
Wenlong Huang, Fei Xia, Ted Xiao, Harris Chan, Jacky Liang, Pete Florence, Andy Zeng, Jonathan Tompson, Igor Mordatch, Yevgen Chebotar, Pierre Sermanet, Tomas Jackson, Noah Brown, Linda Luu, Sergey Levine, Karol Hausman, and Brian Ichter.
\newblock {Inner Monologue: Embodied Reasoning through Planning with Language Models}.
\newblock In \emph{Proceedings of the 6th Conference on Robot Learning}, 2023.

\bibitem[James et~al.(2022)James, Rosman, and Konidaris]{james2022autonomous}
Steven James, Benjamin Rosman, and George Konidaris.
\newblock {Autonomous Learning of Object-Centric Abstractions for High-Level Planning}.
\newblock In \emph{Proceedings of the 10th International Conference on Learning Representation}, 2022.

\bibitem[Jiang et~al.(2025)Jiang, Huang, Wu, Li, Garg, Nayyeri, Wang, and Li]{jiang2024roboexp}
Hanxiao Jiang, Binghao Huang, Ruihai Wu, Zhuoran Li, Shubham Garg, Hooshang Nayyeri, Shenlong Wang, and Yunzhu Li.
\newblock {RoboEXP: Action-Conditioned Scene Graph via Interactive Exploration for Robotic Manipulation}.
\newblock In \emph{Proceedings of the 8th Conference on Robot Learning}, 2025.

\bibitem[Jiang et~al.(2021)Jiang, Grefenstette, and Rockt{\"a}schel]{jiang2021prioritizedlevelreplay}
Minqi Jiang, Edward Grefenstette, and Tim Rockt{\"a}schel.
\newblock {Prioritized Level Replay}.
\newblock In \emph{Proceedings of the 38th International Conference on Machine Learning (ICML)}, 2021.

\bibitem[Johnson et~al.(2017)Johnson, Hariharan, Van Der~Maaten, Fei-Fei, Lawrence~Zitnick, and Girshick]{johnson2016clevrdiagnosticdatasetcompositional}
Justin Johnson, Bharath Hariharan, Laurens Van Der~Maaten, Li~Fei-Fei, C~Lawrence~Zitnick, and Ross Girshick.
\newblock {CLEVR: A Diagnostic Dataset for Compositional Language and Elementary Visual Reasoning}.
\newblock In \emph{Proceedings of the IEEE Conference on Computer Vision and Pattern Recognition}, 2017.

\bibitem[Juba et~al.(2021)Juba, Le, and Stern]{juba2021safe}
Brendan Juba, Hai~S. Le, and Roni Stern.
\newblock {Safe Learning of Lifted Action Models}.
\newblock In \emph{{Proceedings of the 18th International Conference on Principles of Knowledge Representation and Reasoning (KR)}}, 11 2021.

\bibitem[Katz \& Lee(2023)Katz and Lee]{katz-lee-socs2023}
Michael Katz and Junkyu Lee.
\newblock K* and partial order reduction for top-quality planning.
\newblock In \emph{Proceedings of the 16th Annual Symposium on Combinatorial Search}. {AAAI} Press, 2023.

\bibitem[Konidaris(2019)]{KONIDARIS20191}
George Konidaris.
\newblock {On the Necessity of Abstraction}.
\newblock \emph{Current Opinion in Behavioral Sciences}, 29:\penalty0 1--7, 2019.
\newblock ISSN 2352-1546.

\bibitem[Konidaris \& Barto(2009)Konidaris and Barto]{konidaris2009skill}
George Konidaris and Andrew Barto.
\newblock {Skill Discovery in Continuous Reinforcement Learning Domains using Skill Chaining}.
\newblock In \emph{Advances in Neural Information Processing Systems (NIPS)}, volume~22, 2009.

\bibitem[Konidaris et~al.(2018)Konidaris, Kaelbling, and Lozano-Pérez]{Konidaris2018FromST}
George Konidaris, Leslie~Pack Kaelbling, and Tomas Lozano-Pérez.
\newblock {From Skills to Symbols: Learning Symbolic Representations for Abstract High-Level Planning}.
\newblock \emph{Journal of Artificial Intelligence Research}, 61:\penalty0 215--289, 2018.

\bibitem[Lamanna et~al.(2023)Lamanna, Serafini, Faridghasemnia, Saffiotti, Saetti, Gerevini, and Traverso]{Lamanna2023PlanningFL}
Leonardo Lamanna, Luciano Serafini, Mohamadreza Faridghasemnia, Alessandro Saffiotti, Alessandro Saetti, Alfonso Gerevini, and Paolo Traverso.
\newblock {Planning for Learning Object Properties}.
\newblock \emph{Proceedings of the AAAI Conference on Artificial Intelligence}, 2023.

\bibitem[Li \& Silver(2023)Li and Silver]{Li2023EmbodiedAL}
Amber Li and Tom Silver.
\newblock {Embodied Active Learning of Relational State Abstractions for Bilevel Planning}.
\newblock In \emph{Proceedings of The 2nd Conference on Lifelong Learning Agents}, 2023.

\bibitem[Li et~al.(2024)Li, Yu, Cheng, and Xu]{li2024league}
Zhaoyi Li, Kelin Yu, Shuo Cheng, and Danfei Xu.
\newblock {LEAGUE++: Empowering Continual Robot Learning via Guided Skill Acquisition with Large Language Models}.
\newblock In \emph{ICLR 2024 Workshop on Large Language Model (LLM) Agents}, 2024.

\bibitem[Liang et~al.(2025)Liang, Kumar, Tang, Weller, Tenenbaum, Silver, Henriques, and Ellis]{liang2025visualpredicator}
Yichao Liang, Nishanth Kumar, Hao Tang, Adrian Weller, Joshua~B. Tenenbaum, Tom Silver, Joao~F. Henriques, and Kevin Ellis.
\newblock {VisualPredicator: Learning Abstract World Models with Neuro-Symbolic Predicates for Robot Planning}.
\newblock In \emph{Proceedings of the 13th International Conference on Learning Representations}, 2025.

\bibitem[Liang et~al.(2026)Liang, Nguyen, Yang, Li, Tenenbaum, Rasmussen, Weller, Tavares, Silver, and Ellis]{liang2025exopredicatorlearningabstractmodels}
Yichao Liang, Dat Nguyen, Cambridge Yang, Tianyang Li, Joshua~B Tenenbaum, Carl~Edward Rasmussen, Adrian Weller, Zenna Tavares, Tom Silver, and Kevin Ellis.
\newblock {ExoPredicator: Learning Abstract Models of Dynamic Worlds for Robot Planning}.
\newblock In \emph{Proceedings of the 14th International Conference on Learning Representations}, volume 2026, pp.\  49174--49214, 2026.

\bibitem[Majumdar et~al.(2024)Majumdar, Ajay, Zhang, Putta, Yenamandra, Henaff, Silwal, Mcvay, Maksymets, Arnaud, Yadav, Li, Newman, Sharma, Berges, Zhang, Agrawal, Bisk, Batra, Kalakrishnan, Meier, Paxton, Sax, and Rajeswaran]{OpenEQA2023}
Arjun Majumdar, Anurag Ajay, Xiaohan Zhang, Pranav Putta, Sriram Yenamandra, Mikael Henaff, Sneha Silwal, Paul Mcvay, Oleksandr Maksymets, Sergio Arnaud, Karmesh Yadav, Qiyang Li, Ben Newman, Mohit Sharma, Vincent Berges, Shiqi Zhang, Pulkit Agrawal, Yonatan Bisk, Dhruv Batra, Mrinal Kalakrishnan, Franziska Meier, Chris Paxton, Alexander Sax, and Aravind Rajeswaran.
\newblock {OpenEQA: Embodied Question Answering in the Era of Foundation Models}.
\newblock In \emph{Proceedings of the 2024 IEEE/CVF Conference on Computer Vision and Pattern Recognition (CVPR)}, 2024.

\bibitem[Mao et~al.(2019)Mao, Gan, Kohli, Tenenbaum, and Wu]{mao2019neurosymbolicconceptlearnerinterpreting}
Jiayuan Mao, Chuang Gan, Pushmeet Kohli, Joshua~B Tenenbaum, and Jiajun Wu.
\newblock {The Neuro-Symbolic Concept Learner: Interpreting Scenes, Words, and Sentences From Natural Supervision}.
\newblock In \emph{Proceedings of the 7th International Conference on Learning Representations}, 2019.

\bibitem[Migimatsu \& Bohg(2022)Migimatsu and Bohg]{migimatsu2022grounding}
Toki Migimatsu and Jeannette Bohg.
\newblock {Grounding Predicates through Actions}.
\newblock In \emph{Proceedings of the 2022 International Conference on Robotics and Automation}, 2022.

\bibitem[Mu et~al.(2024)Mu, Zhang, Hu, Wang, Ding, Jin, Wang, Dai, Qiao, and Luo]{mu2023embodiedgptvisionlanguagepretrainingembodied}
Yao Mu, Qinglong Zhang, Mengkang Hu, Wenhai Wang, Mingyu Ding, Jun Jin, Bin Wang, Jifeng Dai, Yu~Qiao, and Ping Luo.
\newblock {EmbodiedGPT: Vision-Language Pre-Training via Embodied Chain of Thought}.
\newblock In \emph{Advances in Neural Information Processing Systems}, 2024.

\bibitem[Nasiriany et~al.(2024)Nasiriany, Xia, Yu, Xiao, Liang, Dasgupta, Xie, Driess, Wahid, Xu, Vuong, Zhang, Lee, Lee, Xu, Kirmani, Zhu, Zeng, Hausman, Heess, Finn, Levine, and Ichter]{google2024pivot}
Soroush Nasiriany, Fei Xia, Wenhao Yu, Ted Xiao, Jacky Liang, Ishita Dasgupta, Annie Xie, Danny Driess, Ayzaan Wahid, Zhuo Xu, Quan Vuong, Tingnan Zhang, Tsang-Wei~Edward Lee, Kuang-Huei Lee, Peng Xu, Sean Kirmani, Yuke Zhu, Andy Zeng, Karol Hausman, Nicolas Heess, Chelsea Finn, Sergey Levine, and Brian Ichter.
\newblock {PIVOT: Iterative Visual Prompting Elicits Actionable Knowledge for VLMs}.
\newblock In \emph{Proceedings of the 41st International Conference on Machine Learning}, 2024.

\bibitem[OpenAI(2025)]{openai_gpt5}
OpenAI.
\newblock {Introducing GPT-5}, 2025.
\newblock URL \url{https://openai.com/index/introducing-gpt-5/}.

\bibitem[Raman et~al.(2024)Raman, Cohen, Idrees, Rosen, Mooney, Tellex, and Paulius]{raman2024cape}
Shreyas~Sundara Raman, Vanya Cohen, Ifrah Idrees, Eric Rosen, Ray Mooney, Stefanie Tellex, and David Paulius.
\newblock {CAPE: Corrective Actions from Precondition Errors using Large Language Models}.
\newblock In \emph{Proceedings of the 2024 IEEE International Conference on Robotics and Automation}, 2024.

\bibitem[Rana et~al.(2023)Rana, Haviland, Garg, Abou-Chakra, Reid, and Suenderhauf]{krishan2023sayplan}
Krishan Rana, Jesse Haviland, Sourav Garg, Jad Abou-Chakra, Ian Reid, and Niko Suenderhauf.
\newblock {SayPlan: Grounding Large Language Models using 3D Scene Graphs for Scalable Robot Task Planning}.
\newblock In \emph{Proceedings of the 7th Conference on Robot Learning}, 2023.

\bibitem[Ren et~al.(2024)Ren, Clark, Dixit, Itkina, Majumdar, and Sadigh]{exploreeqa2024}
Allen~Z. Ren, Jaden Clark, Anushri Dixit, Masha Itkina, Anirudha Majumdar, and Dorsa Sadigh.
\newblock {Explore until Confident: Efficient Exploration for Embodied Question Answering}.
\newblock In \emph{Proceedings of Robotics: Science and Systems}, 2024.

\bibitem[Sermanet et~al.(2024)Sermanet, Ding, Zhao, Xia, Dwibedi, Gopalakrishnan, Chan, Dulac-Arnold, Maddineni, Joshi, et~al.]{sermanet2023robovqamultimodallonghorizonreasoning}
Pierre Sermanet, Tianli Ding, Jeffrey Zhao, Fei Xia, Debidatta Dwibedi, Keerthana Gopalakrishnan, Christine Chan, Gabriel Dulac-Arnold, Sharath Maddineni, Nikhil~J Joshi, et~al.
\newblock {RoboVQA: Multimodal Long-Horizon Reasoning for Robotics}.
\newblock In \emph{Proceedings of the 2024 IEEE International Conference on Robotics and Automation}. IEEE, 2024.

\bibitem[Shah et~al.(2020)Shah, Vasudevan, Kumar, Kamojjhala, and Srivastava]{shah2020anytime}
Naman Shah, Deepak~Kala Vasudevan, Kislay Kumar, Pranav Kamojjhala, and Siddharth Srivastava.
\newblock {Anytime Integrated Task and Motion Policies for Stochastic Environments}.
\newblock In \emph{Proceedings of the 2020 IEEE International Conference on Robotics and Automation}. IEEE, 2020.

\bibitem[Shah et~al.(2025)Shah, Nagpal, and Srivastava]{shah2024reals}
Naman Shah, Jayesh Nagpal, and Siddharth Srivastava.
\newblock {From Real World to Logic and Back: Learning Generalizable Relational Concepts For Long Horizon Robot Planning}.
\newblock In \emph{Proceedings of the 9th Conference on Robot Learning}, 2025.

\bibitem[Silver et~al.(2021)Silver, Chitnis, Tenenbaum, Kaelbling, and Lozano-Pérez]{silver2021learningsymbolicoperatorstask}
Tom Silver, Rohan Chitnis, Joshua Tenenbaum, Leslie~Pack Kaelbling, and Tomás Lozano-Pérez.
\newblock {Learning Symbolic Operators for Task and Motion Planning}.
\newblock In \emph{Proceedings of the 2021 IEEE/RSJ International Conference on Intelligent Robots and Systems}, pp.\  3182--3189, 2021.

\bibitem[Silver et~al.(2023)Silver, Chitnis, Kumar, McClinton, Lozano-P{\'e}rez, Kaelbling, and Tenenbaum]{silver2023predinvent}
Tom Silver, Rohan Chitnis, Nishanth Kumar, Willie McClinton, Tom{\'a}s Lozano-P{\'e}rez, Leslie Kaelbling, and Joshua~B Tenenbaum.
\newblock {Predicate Invention for Bilevel Planning}.
\newblock In \emph{Proceedings of the AAAI Conference on Artificial Intelligence}, pp.\  12120--12129, 2023.

\bibitem[Sumers et~al.(2023)Sumers, Marino, Ahuja, Fergus, and Dasgupta]{sumers2023distillinginternetscalevisionlanguagemodels}
Theodore Sumers, Kenneth Marino, Arun Ahuja, Rob Fergus, and Ishita Dasgupta.
\newblock {Distilling Internet-Scale Vision-Language Models into Embodied Agents}.
\newblock In \emph{Proceedings of the Fortieth International Conference on Machine Learning}, 2023.

\bibitem[Sur{\'\i}s et~al.(2023)Sur{\'\i}s, Menon, and Vondrick]{surís2023vipergptvisualinferencepython}
D{\'\i}dac Sur{\'\i}s, Sachit Menon, and Carl Vondrick.
\newblock {ViperGPT: Visual Inference via Python Execution for Reasoning}.
\newblock In \emph{Proceedings of the IEEE/CVF International Conference on Computer Vision}, 2023.

\bibitem[Szot et~al.(2021)Szot, Clegg, Undersander, Wijmans, Zhao, Turner, Maestre, Mukadam, Chaplot, Maksymets, Gokaslan, Vondru\v{s}, Dharur, Meier, Galuba, Chang, Kira, Koltun, Malik, Savva, and Batra]{szot2022habitat20traininghome}
Andrew Szot, Alexander Clegg, Eric Undersander, Erik Wijmans, Yili Zhao, John Turner, Noah Maestre, Mustafa Mukadam, Devendra~Singh Chaplot, Oleksandr Maksymets, Aaron Gokaslan, Vladim\'{\i}r Vondru\v{s}, Sameer Dharur, Franziska Meier, Wojciech Galuba, Angel Chang, Zsolt Kira, Vladlen Koltun, Jitendra Malik, Manolis Savva, and Dhruv Batra.
\newblock {Habitat 2.0: Training Home Assistants to Rearrange their Habitat}.
\newblock In \emph{Advances in Neural Information Processing Systems}, 2021.

\bibitem[Valmeekam et~al.(2023)Valmeekam, Marquez, Sreedharan, and Kambhampati]{valmeekam2023llm}
Karthik Valmeekam, Matthew Marquez, Sarath Sreedharan, and Subbarao Kambhampati.
\newblock {On the Planning Abilities of Large Language Models - A Critical Investigation}.
\newblock In \emph{Advances in Neural Information Processing Systems}, 2023.

\bibitem[Verma et~al.(2022)Verma, Marpally, and Srivastava]{verma2022discovering}
Pulkit Verma, Shashank~Rao Marpally, and Siddharth Srivastava.
\newblock {Discovering User-Interpretable Capabilities of Black-Box Planning Agents}.
\newblock In \emph{Proceedings of the 19th International Conference on Principles of Knowledge Representation and Reasoning}, 2022.

\bibitem[Wang et~al.(2024{\natexlab{a}})Wang, Xie, Jiang, Mandlekar, Xiao, Zhu, Fan, and Anandkumar]{wang2023voyageropenendedembodiedagent}
Guanzhi Wang, Yuqi Xie, Yunfan Jiang, Ajay Mandlekar, Chaowei Xiao, Yuke Zhu, Linxi Fan, and Anima Anandkumar.
\newblock {Voyager: An Open-Ended Embodied Agent with Large Language Models}.
\newblock \emph{Transactions on Machine Learning Research}, 2024{\natexlab{a}}.

\bibitem[Wang et~al.(2019)Wang, Lehman, Clune, and Stanley]{wang2019paired}
Rui Wang, Joel Lehman, Jeff Clune, and Kenneth~O Stanley.
\newblock {Paired Open-Ended Trailblazer (POET): Endlessly Generating Increasingly Complex and Diverse Learning Environments and Their Solutions}.
\newblock \emph{arXiv preprint arXiv:1901.01753}, 2019.

\bibitem[Wang et~al.(2024{\natexlab{b}})Wang, Xian, Chen, Wang, Wang, Fragkiadaki, Erickson, Held, and Gan]{wang2023robogen}
Yufei Wang, Zhou Xian, Feng Chen, Tsun-Hsuan Wang, Yian Wang, Katerina Fragkiadaki, Zackory Erickson, David Held, and Chuang Gan.
\newblock {RoboGen: Towards Unleashing Infinite Data for Automated Robot Learning via Generative Simulation}.
\newblock In \emph{Proceedings of the 41st International Conference on Machine Learning}, 2024{\natexlab{b}}.

\bibitem[Wen et~al.(2024)Wen, Yang, Kautz, and Birchfield]{wen2024foundationpose}
Bowen Wen, Wei Yang, Jan Kautz, and Stan Birchfield.
\newblock {FoundationPose: Unified 6D Pose Estimation and Tracking of Novel Objects}.
\newblock In \emph{Proceedings of the 2024 IEEE/CVF Conference on Computer Vision and Pattern Recognition}, 2024.

\bibitem[Xi et~al.(2024)Xi, Gould, and Thiébaux]{xi2024neuro}
Kai Xi, Stephen Gould, and Sylvie Thiébaux.
\newblock {Neuro-Symbolic Learning of Lifted Action Models from Visual Traces}.
\newblock \emph{Proceedings of the International Conference on Automated Planning and Scheduling (ICAPS)}, 34\penalty0 (1), May 2024.

\bibitem[Yang et~al.(2024)Yang, Dong, Liu, Li, Wang, Tan, Jiang, Kang, Zhang, Zhou, et~al.]{yang2023octopus}
Jingkang Yang, Yuhao Dong, Shuai Liu, Bo~Li, Ziyue Wang, Haoran Tan, Chencheng Jiang, Jiamu Kang, Yuanhan Zhang, Kaiyang Zhou, et~al.
\newblock {Octopus: Embodied Vision-Language Programmer from Environmental Feedback}.
\newblock In \emph{Proceedings of the 2024 European Conference on Computer Vision}, 2024.

\bibitem[Yokoyama et~al.(2024)Yokoyama, Clegg, Truong, Undersander, Yang, Arnaud, Ha, Batra, and Rai]{yokoyama2023ascadaptiveskillcoordination}
Naoki Yokoyama, Alex Clegg, Joanne Truong, Eric Undersander, Tsung-Yen Yang, Sergio Arnaud, Sehoon Ha, Dhruv Batra, and Akshara Rai.
\newblock {ASC: Adaptive Skill Coordination for Robotic Mobile Manipulation}.
\newblock \emph{IEEE Robotics and Automation Letters}, 9\penalty0 (1):\penalty0 779--786, 2024.

\bibitem[Yu(1994)]{yu1994rates}
Bin Yu.
\newblock Rates of convergence for empirical processes of stationary mixing sequences.
\newblock \emph{The Annals of Probability}, pp.\  94--116, 1994.

\end{thebibliography}
